\documentclass[anonymous, journal]{IEEEtran}

\usepackage{amsmath,amssymb,amsfonts}
\usepackage{microtype}
\usepackage{caption}
\usepackage{url}
\usepackage[hidelinks]{hyperref}
\hypersetup{
    colorlinks=true,
    linkcolor=blue,
    citecolor=blue,      
    urlcolor=blue,
}
\usepackage{amsmath}
\usepackage{tikz}
\usetikzlibrary{shapes.geometric, arrows, positioning}
\usepackage{graphicx}
\usepackage{algorithm}
\usepackage{algpseudocode}
\usepackage{url}
\usepackage{lscape}
\usepackage{booktabs}
\usepackage{rotating}
\usepackage{colortbl}
\usepackage[table,xcdraw]{xcolor}
\usepackage{graphicx}
\usepackage{algorithm}
\usepackage{algpseudocode}
\usepackage{stfloats}
\usepackage{enumitem}
\usepackage{multirow}
\usepackage{float}
\usepackage{subcaption}
\usepackage{comment}
\usepackage{subcaption}
\usepackage{orcidlink}
\usepackage[numbers,sort&compress]{natbib}
\def\BibTeX{{\rm B\kern-.05em{\sc i\kern-.025em b}\kern-.08em
    T\kern-.1667em\lower.7ex\hbox{E}\kern-.125emX}}

\begin{document}

\title{From Embeddings to Equations: Genetic-Programming Surrogates for Interpretable Transformer Classification}

 \author{Mohammad~Sadegh~Khorshidi\orcidlink{0000-0001-6556-2926}
 , Navid~Yazdanjue\orcidlink{0000-0001-9670-8422}%
 , Hassan~Gharoun\orcidlink{0000-0001-8298-7512}%
 , Mohammad~Reza~Nikoo\orcidlink{0000-0002-3740-4389}%
 , Fang~Chen\orcidlink{0000-0003-4971-8729}%
 , Amir~H.~Gandomi\orcidlink{0000-0002-2798-0104}
 \thanks{Mohammad Sadegh Khorshidi, Navid Yazdanjue, Hassan Gharoun, Fang Chen, and Amir H. Gandomi are with the Faculty of Engineering \& Information Technology, University of Technology Sydney, Ultimo 2007, Australia. e-mails: msadegh.khorshidi.ak@gmail.com, navid.yazdanjue@gmail.com, hassan.gharoun@student.uts.edu.au, fang.chen@uts.edu.au, gandomi@uni-obuda.hu}
 \thanks{Mohammad Reza Nikoo is with the Department of Civil and Architectural Engineering, Sultan Qaboos University, Muscat, Oman. (e-mail: m.reza@squ.edu.om}%
 \thanks{Amir H. Gandomi is also with the University Research and Innovation Center (EKIK), Obuda University, Budapest 1034, Hungary.}%
 \thanks{This work was supported by the Australian Government through the Australian Research Council under Project DE210101808.}
 \thanks{Corresponding author: Amir H. Gandomi}}

\maketitle

\begin{abstract}
We study symbolic surrogate modeling of frozen Transformer embeddings to obtain compact, auditable classifiers with calibrated probabilities. For five benchmarks (SST2G, 20NG, MNIST, CIFAR10, MSC17), embeddings from ModernBERT, DINOv2, and SigLIP are partitioned on the training set into disjoint, information-preserving views via semantic-preserving feature partitioning (SPFP). A cooperative multi-population genetic program (MEGP) then learns additive, closed-form logit programs over these views. Across 30 runs per dataset we report $F_1$, AUC, log-loss, Brier, expected calibration error (ECE), and symbolic complexity; a canonical model is chosen by a one-standard-error rule on validation $F_1$ with a parsimony tie-break. Temperature scaling fitted on validation yields substantial ECE reductions on test. The resulting surrogates achieve strong discrimination (up to $F_1\approx0.99$ on MNIST, CIFAR10, MSC17; $\approx0.95$ on SST2G), while 20NG remains most challenging. We provide reliability diagrams, dimension usage/overlap statistics, contribution-based importances, and global effect profiles (PDP/ALE), demonstrating faithful, cross-modal explanations grounded in explicit programs.

\end{abstract}


\section{Introduction} \label{Section:Introduction}
\IEEEPARstart{T}{ransformer} encoders have become the de facto backbone for text, vision, and multimodal learning, delivering state-of-the-art performance across tasks but at the cost of opacity in how high-dimensional embeddings drive predictions \cite{devlin2019bert,oquab2023dinov2,zhai2023siglip,xu2023multimodal}. In parallel, the explainable AI (XAI) literature has argued for methods that expose model mechanisms, support human verification, and enable trustworthy deployment in sensitive domains \cite{guidotti2018survey,holzinger2019causability,tjoa2020survey,ali2023explainable,longo2024explainable,roscher2020explainable}. A growing body of work specifically interrogates transformer embedding spaces—examining their geometry, invariances, and relation to interpretable constructs—to move beyond token-level attributions toward principled, model-agnostic explanations \cite{mickus2022dissect,turton-etal-2021-deriving,ren2023all,debelak2024embeddings,schrimpf2020neural}. Yet, despite these advances, there remains a gap between general-purpose post-hoc explanations and compact, human-readable \emph{surrogates} that summarize what fixed transformer embeddings encode for a downstream task, with strong control over parsimony, calibration, and modality-agnostic deployment.

Symbolic modeling offers a promising direction: by representing predictors as algebraic programs over inputs, it yields expressions amenable to inspection, formal reasoning, and scientific use \cite{angelis2023artificial,purcell2023recent}. Modern strands of symbolic regression and program synthesis—Bayesian \cite{jin2019bayesian}, transformer-assisted \cite{valipour2021symbolicgpt}, and hybridized with neural architectures \cite{kim2020integration,cranmer2019learning}—have broadened the search space and improved robustness while preserving interpretability. Applications across scientific discovery and engineering underscore the value of symbolic surrogates for extracting concise laws or descriptors from complex representations \cite{weng2020simple,ding2020dynamic,abdellaoui2021symbolic}. However, most prior symbolic efforts either target raw domain features or require end-to-end retraining. Few frameworks explicitly treat \emph{frozen transformer embeddings} as an interpretable substrate, and even fewer provide a pipeline that (i) partitions embeddings into semantically coherent \emph{views} to reduce redundancy and (ii) trains cooperative symbolic programs that assemble into calibrated, compact, and accurate surrogates.

This paper addresses that gap by studying \emph{symbolic surrogate modeling of transformer embeddings} using a view-based ensemble genetic programming scheme. Concretely, we extract frozen embeddings from ModernBERT for text, DINOv2 for images, and SigLIP for image–text pairs \cite{devlin2019bert,oquab2023dinov2,zhai2023siglip}, and then (on training data only) partition the embedding coordinates into disjoint, information-preserving views with Semantic-Preserving Feature Partitioning (SPFP) \cite{khorshidi2024semantic}. Over those views we fit a cooperative multi-population ensemble GP (MEGP) \cite{Khorshidi2025MEGP} that outputs class logits as sums of per-view symbolic programs. The result is a compact, human-readable surrogate that (a) preserves predictive performance, (b) admits principled model selection by a one-standard-error rule on validation $F_1$, and (c) supports post-hoc calibration and a thorough behavioral analysis (importance, partial dependence, and accumulated local effects) on held-out test data. By operating strictly post-extraction with frozen encoders, the approach decouples representational learning from interpretability, aligning with calls in the XAI literature for modality-agnostic, auditable workflows \cite{guidotti2018survey,ali2023explainable,longo2024explainable,roscher2020explainable}.

Our study is motivated by three converging observations from recent scholarship on transformer representations. First, embedding spaces exhibit structure that is analyzable and, to a degree, alignable across training runs and architectures, suggesting that post-hoc, representation-level explanations are meaningful \cite{mickus2022dissect,ren2023all}. Second, contextualized embeddings can be mapped to cognitively or semantically grounded features, supporting human interpretation when appropriate probes and aggregation are used \cite{turton-etal-2021-deriving,debelak2024embeddings}. Third, as transformer use expands to multimodal settings, interpretable surrogates that operate on top of fixed encoders offer a pragmatic path to cross-domain transparency without sacrificing performance \cite{xu2023multimodal,zhai2023siglip,oquab2023dinov2}. Taken together, these findings motivate a pipeline that treats embeddings as the object of explanation and leverages symbolic modeling to produce faithful, parsimonious summaries.

\textit{Contributions.} This paper makes the following contributions:
(i) We introduce a modality-agnostic pipeline that learns compact symbolic surrogates over \emph{frozen} transformer embeddings by combining SPFP view construction \cite{khorshidi2024semantic} with cooperative multi-population ensemble GP \cite{Khorshidi2025MEGP}.
(ii) We provide a principled model-selection procedure based on a one-standard-error rule on validation $F_1$ with a complexity tie-break, explicitly trading off accuracy and parsimony in a transparent manner, consistent with the goals of interpretable surrogate modeling \cite{angelis2023artificial,purcell2023recent}.
(iii) We conduct a comprehensive calibration and behavioral analysis of the canonical surrogates, including reliability diagrams, temperature scaling, dimension/operation usage statistics, contribution-based importance, and global effect summaries via PDP and ALE, aligning with best practices advocated in XAI surveys \cite{ali2023explainable,longo2024explainable,roscher2020explainable}.
(iv) We evaluate across five benchmarks (SST2G, 20NG, MNIST, CIFAR10, MSC17) spanning text, images, and image–text pairing with ModernBERT, DINOv2, and SigLIP encoders \cite{devlin2019bert,oquab2023dinov2,zhai2023siglip}, demonstrating that symbolic surrogates can be both parsimonious and competitive while offering rich interpretability artifacts.
(v) We release a complete specification of the pipeline and experimental protocol to facilitate reproducible research on embedding-level interpretability, complementing recent efforts to formalize and benchmark explanation quality \cite{guidotti2018survey,ali2023explainable}.

\textit{Positioning within related work.} Our approach differs from feature-level post-hoc explainers and attention visualizations by constructing \emph{explicit symbolic programs} over fixed embeddings, thereby yielding globally interpretable, calibrated surrogates with measurable complexity. It complements recent lines of research that analyze embedding geometry and invariances \cite{mickus2022dissect,ren2023all} and those that extract human-relevant semantic axes from contextualized representations \cite{turton-etal-2021-deriving,debelak2024embeddings}. Methodologically, it builds on contemporary symbolic modeling and hybrid neuro-symbolic ideas \cite{angelis2023artificial,kim2020integration,cranmer2019learning,purcell2023recent}, but targets the specific setting of transformer embeddings and multi-view, cooperative search via SPFP/MEGP \cite{khorshidi2024semantic,Khorshidi2025MEGP}. By evaluating in text, vision, and multimodal regimes, it also speaks to current surveys that call for unified, cross-domain interpretability frameworks \cite{xu2023multimodal,ali2023explainable,longo2024explainable}.

The remainder of the paper is organized as follows. Section~\ref{Section:Related_Works} reviews background on XAI, transformer embeddings, and symbolic modeling. Section~\ref{sec:methods} details our pipeline: datasets and frozen encoders, SPFP view construction, MEGP training, canonical model selection, and calibration/behavioral analyses. Section~\ref{Section:Results_Discussion} presents predictive, calibration, and interpretability results across five benchmarks, including ablations and qualitative diagnostics. Section~\ref{sec:conclusion} summarizes findings and outlines future work.

\section{Related Works} \label{Section:Related_Works}
Explainable artificial intelligence (XAI) has emphasized that high predictive performance alone is insufficient in sensitive settings, where users require mechanistic accounts of model behavior and evidence for causal reasoning \cite{holzinger2019causability}. Recent syntheses catalogue the methodological landscape and its open challenges, stressing the need to move beyond ad-hoc post-hoc explanations toward approaches with verifiable faithfulness and domain utility \cite{ali2023explainable,longo2024explainable}. Within this agenda, transparent surrogates that are compact and auditable are appealing because they permit global inspection, stress testing, and principled calibration.

Classic surveys systematize black-box explanation methods, their desiderata, and evaluation pitfalls, including instability and potential disagreement across explainers \cite{guidotti2018survey}. Meta-surveys extend these critiques with taxonomies of user needs and regulatory constraints, reinforcing that explanation quality must be judged relative to task risk and stakeholder expertise \cite{saeed2023explainable}. In scientific discovery, a parallel thread argues that explanations should produce new, testable hypotheses, not only highlight salient inputs \cite{roscher2020explainable}. These positions collectively motivate intrinsic or surrogate models that yield human-readable structure while retaining competitive accuracy.

Transformers are the de facto backbone across language, vision, audio, and multimodal learning, but their deep, distributed representations complicate interpretation \cite{zhao2024explainability}. Layerwise and headwise analyses have shown partial specialization, yet stable global semantics are hard to attribute to particular subspaces without further assumptions \cite{debelak2024embeddings}. Studies of the geometry of embedding spaces indicate anisotropy, axis entanglement, and complex factorization properties that vary by layer and task \cite{mickus2022dissect}. Methods that derive contextualized semantic features from BERT-like embeddings demonstrate that some dimensions align with psycholinguistic constructs, but the mapping is model- and layer-dependent \cite{turton-etal-2021-deriving}. Representation alignment work suggests that independently trained Transformer encoders learn approximately isomorphic spaces up to invertible transforms, complicating direct, dimension-wise attribution unless spaces are aligned \cite{ren2023all}. Complementary evidence from brain-model comparisons underscores functional specialization trends across layers while cautioning against simplistic single-layer interpretations \cite{kumar2022shared}.

Multimodal surveys document additional challenges in aligning heterogeneous embeddings across text, image, and audio streams, where topologies and noise characteristics differ \cite{xu2023multimodal}. Audio-specific work illustrates that replacing convolutional front-ends with attention yields highly expressive yet opaque time–frequency embeddings, intensifying post-hoc burdens unless structure is imposed \cite{verma2021audio}. Recent results even question the necessity of semantically rich input embeddings, showing that frozen, visually derived Unicode embeddings can suffice for language modeling, with semantics emerging through the network rather than being localized in the token vectors \cite{bochkov2025emergent}. These observations collectively suggest value in model-agnostic surrogates that interpret \emph{frozen} embeddings produced by strong encoders such as ModernBERT, DINOv2, or SigLIP \cite{warner2024modernbert,oquab2023dinov2,zhai2023siglip}.

Symbolic regression (SR) offers an attractive XAI pathway because it returns explicit analytical expressions that trade accuracy for parsimony under a controlled bias–variance–simplicity frontier \cite{angelis2023artificial}. Bayesian SR places priors on expression grammars and explores program space with MCMC, improving robustness and uncertainty quantification compared to purely evolutionary search \cite{jin2019bayesian}. Neural–symbolic hybrids integrate differentiable operators and structural sparsity to learn closed-form relations within end-to-end pipelines, thereby recovering compact equations that extrapolate better than black-box networks on mechanistic tasks \cite{kim2020integration}. Deterministic sparse SR such as SISSO searches large feature spaces and then applies sparsifying operators to identify low-dimensional, physically meaningful descriptors; modern implementations increase expressivity with grammar control while preserving interpretability \cite{purcell2023recent}. Beyond method development, SR has accelerated materials discovery by uncovering simple descriptors that correlate with catalytic activity, highlighting its value for scientific hypothesis generation \cite{weng2020simple}. Domain applications include fault prognosis with dynamic symbolic structures and forecasting with compact, auditable equations \cite{ding2020dynamic,abdellaoui2021symbolic}.

Applying SR to interpret Transformer embeddings is a natural convergence: embeddings provide a fixed, high-capacity feature basis, while SR supplies an intrinsically transparent hypothesis class that can approximate the decision surface with explicit programs. Transformer-based generators for SR (e.g., SymbolicGPT) demonstrate that Transformer priors can help search symbolic program space effectively, hinting at fruitful cross-pollination between sequence models and equation discovery \cite{valipour2021symbolicgpt}. In parallel, graph-network approaches to symbolic physics recover closed-form laws from data, reinforcing the thesis that neural representations can be post-hoc distilled into concise analytic structure \cite{cranmer2019learning}. Yet, despite these advances, most SR-for-XAI work has focused on raw tabular or engineered descriptors, not on \emph{frozen deep embeddings} across language, vision, and multimodal tasks at scale.

Interpretability for Transformer \emph{models} has also progressed via architectural inductive biases. Unrolling graph smoothness priors into lightweight Transformer layers yields units whose behavior admits interpretation in graph-signal terms, shrinking parameter counts while clarifying computation \cite{do2024interpretable}. Semantics-aware dimensionality reduction seeks to compress BERT embeddings while preserving reconstruction and semantic fidelity, enabling more tractable analysis of low-dimensional surrogates \cite{boyapati2024semanformer}. Language-complexity evaluations illustrate that embedding choices and architectures influence sentence-level difficulty modeling, providing indirect interpretability handles via downstream behavioral metrics \cite{ivanov2022sentence}. Hybrid attention–recurrent sentiment models report improved robustness and interpretability with token-level saliency, though they still rely primarily on post-hoc attributions rather than global, closed-form surrogates \cite{jahin2024hybrid}.

In application domains, the interpretability demand is acute. Healthcare studies increasingly adopt Transformer-based predictors but accompany them with attention or feature-importance visualizations to bolster clinical trust; nevertheless, these remain local and model-specific rather than yielding global mechanistic surrogates \cite{ma2025leveraging}. Patient-graph formulations expose salient relational structure via attention, offering an interpretable path but one tied to graph construction rather than to the geometry of embedding spaces themselves \cite{boll2024graph}. Software-engineering surveys similarly call for transparent code embeddings to avoid propagating systemic bugs, again pointing to the need for principled surrogates that expose decision logic \cite{wong2023natural}. In agriculture and other applied vision tasks, Grad-CAM-style explanations localize evidence but do not produce concise global rules \cite{zhong2022classification}. Knowledge tracing has begun to incorporate geometry-aware Transformers (hyperbolic or graph-embedded), which aids semantic disentanglement of states but focuses on latent trajectories rather than analytic surrogates over fixed representations \cite{li2025hyperbolic,liang2024gelt}. Time-series surveys emphasize that attention mechanisms capture long-range dependencies but raise the same questions about global, verifiable explanation \cite{wen2023transformers}.

Within this broader literature, our approach targets a specific gap: distilling \emph{frozen} Transformer embeddings into compact, globally valid, human-readable programs that (i) preserve predictive performance; (ii) expose the usage and interactions of embedding dimensions; and (iii) remain amenable to reliability analysis and calibration. Prior SR applications seldom exploit view-inducing partitioners that reduce redundancy and enhance semantic coverage before symbolic modeling; likewise, prior Transformer interpretability largely stops at saliency maps, linear probes, or layer diagnostics rather than end-to-end symbolic surrogates. In contrast, we pair semantic-preserving partitioning with cooperative multi-population genetic programming to produce additive, view-wise programs over ModernBERT, DINOv2, and SigLIP embeddings \cite{oquab2023dinov2,warner2024modernbert,zhai2023siglip}. This design aligns with alignment findings by operating directly in each model’s native embedding space (no cross-model mapping) while enforcing sparsity and parsimony to mitigate anisotropy and entanglement concerns \cite{ren2023all,mickus2022dissect}. Our evaluation suite complements accuracy with calibration and reliability, addressing critiques that XAI methods should assess probability quality, not only discrimination \cite{ali2023explainable}.

Relative to intrinsic interpretability work, our surrogate perspective maintains the strong inductive biases and performance of state-of-the-art encoders while offering a post-hoc but \emph{global} explanation in closed form. Relative to SR-for-XAI studies, our contributions are situated in the high-dimensional, deep-embedding regime across modalities, where we demonstrate that disciplined partitioning and cooperative symbolic modeling yield stable programs with interpretable usage statistics and effect profiles. Finally, by reporting dimension usage, overlaps, importance, and global effects, we aim to connect geometric properties of embedding spaces to tangible, auditable rules—an objective advocated by both XAI surveys and scientific-explanation frameworks \cite{roscher2020explainable,saeed2023explainable}.

\section{Methods}\label{sec:methods}

\subsection{Overview: symbolic surrogate modeling of transformer embeddings}
We pursue post-hoc interpretability by learning compact programs that summarize fixed transformer embeddings. For each dataset in Table~\ref{tab_datasets}, we extract frozen representations from ModernBERT for text, DINOv2 for images, and SigLIP for image–text pairs (Table~\ref{tab_transformer_characteristics}; \cite{warner2024modernbert,oquab2023dinov2,zhai2023siglip}). On the training split, semantic-preserving feature partitioning (SPFP) divides the embedding coordinates into \(V\) disjoint, information-preserving views; the learned partition is then kept fixed for validation and test \cite{khorshidi2024semantic}. A cooperative multi-population genetic program (MEGP) with \(V\) populations (one per view) searches arithmetic expression trees with real constants; fitness is cross-entropy with parsimony, and constants are tuned jointly with structure \cite{Khorshidi2025MEGP}. We run 30 independent seeds using the configuration in Table~\ref{tab_gp_parameters}, and select a single canonical surrogate per dataset via a one-standard-error rule on validation \(F_1\) with a minimum-complexity tie-break. A scalar temperature is fit on validation and applied on test for calibrated probabilities.

Downstream analyses for the canonical surrogate include reliability diagrams and expected calibration error, dimension and operator usage statistics, overlap of dimensions across logits (histograms and UpSet plots), contribution-based importance of embedding coordinates, and global effect visualizations via partial dependence and accumulated local effects with bootstrap confidence intervals. Full details of each step appear in the following subsections.

\begin{table*}[t]
\centering
\small
\setlength{\tabcolsep}{6pt}
\renewcommand{\arraystretch}{1.15}
\caption{Datasets used in this study.}
\label{tab_datasets}
\begin{tabular}{l|l l l c c}
\hline
\textbf{Dataset} & \textbf{Modality} & \textbf{Task} &
\textbf{Records (train / val / test)} & \textbf{\#Cls} & \textbf{Official val?} \\
\hline
SST-2 GLUE (SST2G)             & Text       & Sentiment (binary)   & 67{,}349 / 872 / 1{,}821                 & 2  & Yes \\
20 Newsgroups (20NG)$^{\ast}$  & Text       & Topic                & 10{,}183 / 1{,}131$^{\ast}$ / 7{,}532    & 20 & No  \\
MNIST (MNIST)$^{\ast}$         & Image      & Digit classification & 54{,}000 / 6{,}000$^{\ast}$ / 10{,}000   & 10 & No  \\
CIFAR-10 (CIFAR10)$^{\ast}$    & Image      & Image classification & 45{,}000 / 5{,}000$^{\ast}$ / 10{,}000   & 10 & No  \\
MS COCO 2017 (MSC17)$^{\dagger}$ & Image+Text & Binary match         & 1{,}064{,}580$^{\dagger}$ / 118{,}290$^{\dagger}$ / 50{,}000$^{\dagger}$ & 2 & Yes (images) \\
\hline
\end{tabular}

\vspace{4pt}
\begin{minipage}{0.96\textwidth}\footnotesize
\emph{Notes.}\\
$^{\ast}$ No official validation split; we hold out 10\% of the training set for validation.\\
$^{\dagger}$ Counts are at the \emph{pair} level for image–caption matching (each positive caption paired with one hard negative; 1:1). Because \texttt{test2017} lacks public labels, we test on \texttt{val2017} (5{,}000 images) and split \texttt{train2017} into train/val (106{,}458 / 11{,}829 images).
\end{minipage}
\end{table*}

\subsection{Datasets and embedding extraction}\label{subsec:data-emb}
We use five benchmarks (Table~\ref{tab_datasets}): SST-2 Glue (binary sentiment) (SST2G), 20~Newsgroups (20-way topic) (20NG), MNIST and CIFAR-10 (10-way vision) (CIFAR10), and MS~COCO~2017 (MSC17) caption matching (binary image--text pairing)~\cite{hf_glue_sst2,hf_20newsgroups_setfit,hf_mnist_ylecun,hf_cifar10_uoftcs,hf_mscoco2017_shunk031}.
Official train/validation/test splits are respected when available; otherwise 10\% of the training partition is held out as validation (the split index is fixed across all 30 runs). Only the training split is used to fit preprocessing statistics and the SPFP partition; the learned transforms and partitions are applied unchanged to validation and test.

\begin{table*}[t]
\centering
\small
\setlength{\tabcolsep}{6pt}
\renewcommand{\arraystretch}{1.15}
\caption{Transformer backbones used in this study (all encoders are frozen).}
\label{tab_transformer_characteristics}
\begin{tabular}{l|l l l c l}
\hline
\textbf{Model} & \textbf{Modality} & \textbf{$d_{\mathrm{model}}$ / Layers / Heads} &
\textbf{Embedding to GP} & \textbf{Params (M)} \\
\hline
ModernBERT-base$^{\ast}$ & Text       & 768 / 22 / 12                  & 768  & 149  \\
DINOv2 ViT-L/14$^{\dagger}$ & Image    & 1024 / 24 / 16                 & 1024 & 300 \\
SigLIP So400M (p14-384)$^{\ddagger}$ & Image+Text & 1152 / 27 / 16 (both towers) & 1152 & 400 (total)\\
\hline
\end{tabular}

\vspace{4pt}
\begin{minipage}{0.96\textwidth}\footnotesize
\emph{Notes.}\\
$^{\ast}$ Variant/checkpoint: \texttt{answerdotai/ModernBERT-base}. Embedding: 768-D final-layer [CLS] token.\\
$^{\dagger}$ Variant/checkpoint: \texttt{facebook/dinov2-large} (\texttt{dinov2\_vitl14}). Embedding: 1024-D final [CLS]; inputs 224$\times$224; patch size 14.\\
$^{\ddagger}$ Variant/checkpoint: \texttt{google/siglip-so400m-patch14-384}. Embedding: 1152-D concatenation of 576-D image + 576-D text projected, unit-normalized features (first 576 image, second 576 text).\\
We standardize (z-score) non-normalized embeddings (ModernBERT, DINOv2) using training-set statistics; SigLIP embeddings remain unit-normalized.
\end{minipage}
\end{table*}

\paragraph{Encoders and frozen representations.}
Text inputs are embedded with ModernBERT-base~\cite{warner2024modernbert}: standard subword tokenization, maximum sequence length in \{128,\,256\} depending on the dataset, and the final-layer \text{\texttt{[CLS]}} vector as a 768-dimensional representation (no fine-tuning). Images are embedded with DINOv2 ViT-L/14~\cite{oquab2023dinov2}: inputs are resized to \(224\times 224\) with ImageNet-style normalization; the final-block \text{\texttt{[CLS]}} token yields a 1024-dimensional vector. For MSC17 we use SigLIP So400M (patch14--384)~\cite{zhai2023siglip} (Table~\ref{tab_transformer_characteristics}): for each image--caption pair we form a 1152-dimensional concatenation \(z=[\,z_{\mathrm{img}}; \, z_{\mathrm{text}}\,]\) where each tower contributes 576 projected coordinates; both towers are unit-normalized by construction. All encoders remain frozen.

\paragraph{Feature standardization and notation.}
Let \(X\in\mathbb{R}^{n\times d}\) denote the embedding matrix for a split with labels \(y\).
For ModernBERT and DINOv2 we z-score each feature using training-only statistics: let
\(\mu_{\mathrm{train}},\sigma_{\mathrm{train}}\in\mathbb{R}^{d}\) be the per-dimension mean and standard deviation computed on the training set, and apply
\(\tilde{X}=(X-\mu_{\mathrm{train}}\mathbf{1}^{\top})\oslash(\sigma_{\mathrm{train}}+\epsilon)\)
(element-wise division; \(\epsilon>0\) for numerical stability). The same \(\mu_{\mathrm{train}},\sigma_{\mathrm{train}}\) are reused for validation and test.

For SigLIP on MSC17, we start from the released 1152-dimensional projected embeddings for each tower (image and text), which are unit-normalized. We do not z-score SigLIP. Instead, we perform non-overlapping \(2{:}1\) average pooling along the feature axis within each tower to obtain \(576\) dimensions per tower, and then concatenate the pooled image and text vectors to form a \(1152\)-dimensional input to MEGP.

The resulting \(\tilde{X}_{\mathrm{train}},\tilde{X}_{\mathrm{val}},\tilde{X}_{\mathrm{test}}\) feed SPFP partitioning (\S\ref{subsec:spfp}) and MEGP training (\S\ref{subsec:gp}).

\subsection{Semantic-Preserving Feature Partitioning (SPFP)}\label{subsec:spfp}
SPFP~\cite{khorshidi2024semantic} converts a single embedding into $V$ disjoint, information-preserving \emph{views} for MEGP. Let $\tilde X\!\in\!\mathbb{R}^{n\times d}$ be the standardized \emph{training} embeddings with labels $y$, and initialize a pool of dimensions $\mathcal{P}=\{1,\dots,d\}$. Views are grown sequentially and applied unchanged to validation/test.

\paragraph{Greedy growth.}
For the current view $\mathcal{V}\subset\mathcal{P}$, each candidate $j\in\mathcal{P}\setminus\mathcal{V}$ is scored using the relevance--redundancy criterion of~\cite{khorshidi2024semantic}, which promotes high label-utility w.r.t.\ $y$ while penalizing redundancy with dimensions already in $\mathcal{V}$. The estimators and weights follow the defaults in~\cite{khorshidi2024semantic}.

\paragraph{Per-view stopping.}
Growth stops when the information-preservation test of~\cite{khorshidi2024semantic} is met \emph{and} when the view reaches a budget
$N_F=\big\lceil 0.1\,d \big\rceil$.

\paragraph{Pool update and continuation.}
After finalizing a view $\mathcal{V}_v$, remove a fraction $r=0.2$ of its dimensions from the pool to encourage inter-view diversity:
$\mathcal{P}\leftarrow\mathcal{P}\setminus\mathcal{R}_v$ with $|\mathcal{R}_v|=\lceil r\,|\mathcal{V}_v|\rceil$.
Repeat growth and removal \emph{until no dimensions remain in the pool}; the resulting disjoint index sets
$\{\mathcal{V}_1,\dots,\mathcal{V}_V\}$ determine the number of views $V$.

\subsection{Multi-Population Ensemble Genetic Programming (MEGP)}\label{subsec:gp}
We train a cooperative, view-aware GP to produce class logits directly from the fixed embeddings. For a dataset with $K$ classes and SPFP views $\{\mathcal{V}_1,\dots,\mathcal{V}_V\}$, MEGP instantiates $V$ coevolving populations (one per view). The terminal set of population $v$ is restricted to the standardized dimensions indexed by $\mathcal{V}_v$ plus ephemeral constants; the primitive set is $\{+,\, -,\, \times,\, \div\}$ with protected division. Each individual encodes a vector of symbolic trees that together output class logits $z_c(\mathbf{x})$. At evaluation time, the per-view contributions are aggregated additively:
\[
\begin{aligned}
z_c(\mathbf{x}) &= \sum_{v=1}^V f^{(v)}_c\!\bigl(\mathbf{x}_{\mathcal{V}_v}\bigr),\\
p_c(\mathbf{x}) &= \mathrm{softmax}_c\bigl(\mathbf{z}(\mathbf{x})\bigr)\quad \text{(multiclass)}.
\end{aligned}
\]
where \(z_c(\mathbf{x})\) is the class-\(c\) logit, \(f^{(v)}_c\) is the GP-learned symbolic program for class \(c\) on view \(v\) applied to \(\mathbf{x}_{\mathcal{V}_v}\), and \(p_c(\mathbf{x})\) is the predicted class-\(c\) probability given by the softmax. Fitness is the negative log-likelihood (cross-entropy) averaged over stochastic mini-batches; early termination within a run uses a stall criterion. Selection, variation, and population sizes follow a fixed schedule (Table~\ref{tab_gp_parameters}). Variation uses subtree crossover, point mutation, and reproduction with probabilities $p_c$, $p_m$, and $p_r$; tree depth is capped, and constants are sampled uniformly from the specified range.

Coevolution proceeds by assembling \emph{teams} (one candidate per view) for fitness evaluation, thereby encouraging complementary, non-redundant programs across views while preserving per-view interpretability. We run 30 independent seeds per dataset with identical train/val/test splits and the same SPFP views. After training, we retain the per-class \emph{logit} programs (trees) and the assembled additive form as the model delivered to subsequent selection and calibration stages. Additional implementation detail on cooperative MEGP can be found in~\cite{Khorshidi2025MEGP}.

\begin{table}[t]
\centering
\small
\setlength{\tabcolsep}{6pt}
\renewcommand{\arraystretch}{1.15}
\caption{Genetic Programming (GP) configuration used in this study.}
\label{tab_gp_parameters}
\begin{tabular}{l|l}
\hline
\textbf{Parameter} & \textbf{Value} \\
\hline
Number of runs & 30 \\
Number of populations & Equal to the constructed views \\
Population size & 30 \\
Max generations & 150 \\
Stall generations & 30 \\
Genes per individual & 10 \\
Max tree depth & 10 \\
Initialization & Half-and-Half \\
$EF^{\mathrm{iso}}$ & 0.033 \\
$EF^{\mathrm{en}}$ & 0.10 \\
$p_{\mathrm{en}}$ & 0.75 \\
$p_c$ (crossover) & 0.84 \\
$p_m$ (mutation) & 0.14 \\
$p_r$ (reproduction) & 0.02 \\
Constant range & $[-10,\,10]$ \\
Functional nodes & $+,\,-,\,/,\times$ \\
Batch size & data size / 50 \\
Epochs & 1000 \\
Learning rate & 0.001 \\
\hline
\end{tabular}
\end{table}

\subsection{Canonical model selection and complexity}\label{subsec:canonical}
Each dataset/run yields one MEGP surrogate per random seed. Let the validation macro-\(F_{1}\) of run \(r\in\{1,\dots,R\}\) be
\[
m_r \;=\; \frac{1}{K}\sum_{c=1}^{K}\frac{2\,\mathrm{Prec}_{c}\,\mathrm{Rec}_{c}}{\mathrm{Prec}_{c}+\mathrm{Rec}_{c}},
\]
where \(K\) is the number of classes and precision/recall are computed on the validation split. Define the best observed score \(m^\star=\max_r m_r\) and its standard error
\[
\mathrm{SE} \;=\; \sqrt{\frac{1}{R(R-1)}\sum_{r=1}^{R}\bigl(m_r-\bar m\bigr)^2}, \qquad \bar m=\tfrac{1}{R}\sum_r m_r .
\]
We apply a 1-SE rule: the \emph{feasible set} is \(\mathcal{S}=\{r:\, m_r \ge m^\star-\mathrm{SE}\}\).
Among \(\mathcal{S}\) we pick the \emph{canonical} surrogate by lexicographic minimization of complexity, then depth:
\[
r_{\mathrm{canon}} \;=\; \arg\min_{r\in\mathcal{S}}\; \bigl(C_r,\; D_r\bigr).
\]

\paragraph{Complexity and depth.}
For run \(r\), let \(f^{(v)}_{c}\) denote the logit program for class \(c\) in view \(v\) (only nonempty views contribute).
The symbolic complexity is the total node count across all logits,
\[
C_r \;=\; \sum_{c=1}^{K}\sum_{v=1}^{V} \bigl[\#\text{internal ops}(f^{(v)}_{c})+\#\text{terminals}(f^{(v)}_{c})\bigr],
\]
with terminals comprising dimensions and constants, and internal ops drawn from \(\{+,-,\times,/\,\}\).
Depth \(D_r\) is the maximum parse-tree depth over all \(f^{(v)}_{c}\).
Ties after \((C_r,D_r)\) are broken by preferring fewer unique dimensions, then by a deterministic hash of the serialized programs.

\paragraph{Freezing and test-time evaluation.}
After selecting \(r_{\mathrm{canon}}\) on validation, we freeze its programs and evaluate all reported test metrics (including post-hoc calibration in \S\ref{subsec:calibration}) exactly once on the held-out test split.

\subsection{Post-hoc calibration and reliability}\label{subsec:calibration}
We calibrate the canonical MEGP logits with scalar temperature scaling fitted on validation and applied unchanged to test.
Given pre-calibration logits \(\mathbf{z}(\mathbf{x})\), the calibrated probabilities are
\[
p_c^{(T)}(\mathbf{x}) \;=\; \mathrm{softmax}_c\!\bigl(\mathbf{z}(\mathbf{x})/T\bigr),
\]
with \(T>0\) chosen by minimizing the negative log-likelihood on the validation split.
Calibration does not change the predicted class under argmax and thus leaves \(\mathrm{F}_1\) (and typically AUC) essentially unaffected; it alters only the probability geometry, impacting log-loss, Brier, and ECE.
For reliability diagrams we partition \([0,1]\) into 20 equal-width bins, omit empty bins, and plot empirical accuracy versus mean confidence (with Clopper–Pearson binomial intervals); probabilities are clipped to \([10^{-6},1-10^{-6}]\) for numerical stability.
We report pre/post values for log-loss, Brier, and ECE together with the fitted \(T\).

\subsection{Symbolic reporting and sparsity statistics}\label{subsec:symbolic}
All analyses in this section use the canonical MEGP surrogate per dataset. Each class logit is exported as a closed-form expression over embedding dimensions \(\{d_j\}\) using the primitive set \(\{+,-,\times,/\}\) with protected division and we simplified the logits using the SymPy Python library. We serialize and typeset every logit to the Supplementary Document. For the main text we summarize parsimony and sparsity with a compact table reporting: number of genes (logits), median node count per logit, median tree depth, the union of used dimensions across logits, median dimensions-per-logit, and the operator set actually used.

Dimension usage is quantified at two granularities. First, per-logit sparsity lists the distinct indices \(\{j:\, d_j \text{ appears in that logit}\}\). Second, corpus-level frequency counts \(\nu_j\) tally how often \(d_j\) appears across all logits of the canonical model. We visualize \(\{\nu_j\}\) as sorted histograms and report overlap via UpSet diagrams that enumerate the size of intersections across logits. Operator usage is collected analogously by counting internal nodes by operator type and reporting totals per canonical model. Unless explicitly stated otherwise, counts are taken on the unsimplified program trees as exported from MEGP.

For multimodal MSC17 we treat the pooled SigLIP coordinates as a single 1152-dimensional embedding; sparsity statistics do not assume a particular tower beyond the image–text split described in \S\ref{subsec:data-emb}. Finally, we include a one-line methodological note with each table: SPFP defines the evolutionary views; all embedding dimensions are eligible within their assigned view; no view-level sparsity constraint is enforced beyond what SPFP and MEGP implicitly induce.

\subsection{Evaluation metrics and statistical protocol}\label{subsec:metrics}
We formalize all test metrics (macro-\(F_{1}\), AUC, log-loss, Brier, ECE) and the across-run uncertainty procedure used throughout the paper: 30 independent MEGP runs per dataset with identical train/val/test splits; report mean and \(95\%\) CIs over runs; reliability computed with 20 equal-width bins (empty bins omitted); probabilities clipped to \([10^{-6},1-10^{-6}]\). This subsection also specifies how canonical-model selection (\S\ref{subsec:canonical}) interacts with test reporting (single frozen evaluation on the held-out test set), and how temperature scaling (\S\ref{subsec:calibration}) is fit on validation and applied to test.

\subsection{Behavioral analyses (importance, PDP, ALE)}\label{subsec:behavior}
We detail the computation of contribution-based importance for each embedding dimension, the aggregation to global importance (mean absolute logit contribution), and the selection of top-\(k\) dimensions for effect plots. We define 1D partial dependence (PDP) and accumulated local effects (ALE), including centering of ALE, bootstrap CIs, and a monotonicity score (fraction of the empirical domain with consistent slope). We also specify plotting conventions (binning, smoothing, CI construction) to ensure reproducibility.

\section{Results and Discussion} \label{Section:Results_Discussion}
\subsection{Predictive performance and model parsimony}\label{subsec:results-perf}
\begin{figure*}[t]
\centering
\subfloat[AUC]{%
  \includegraphics[width=0.32\textwidth]{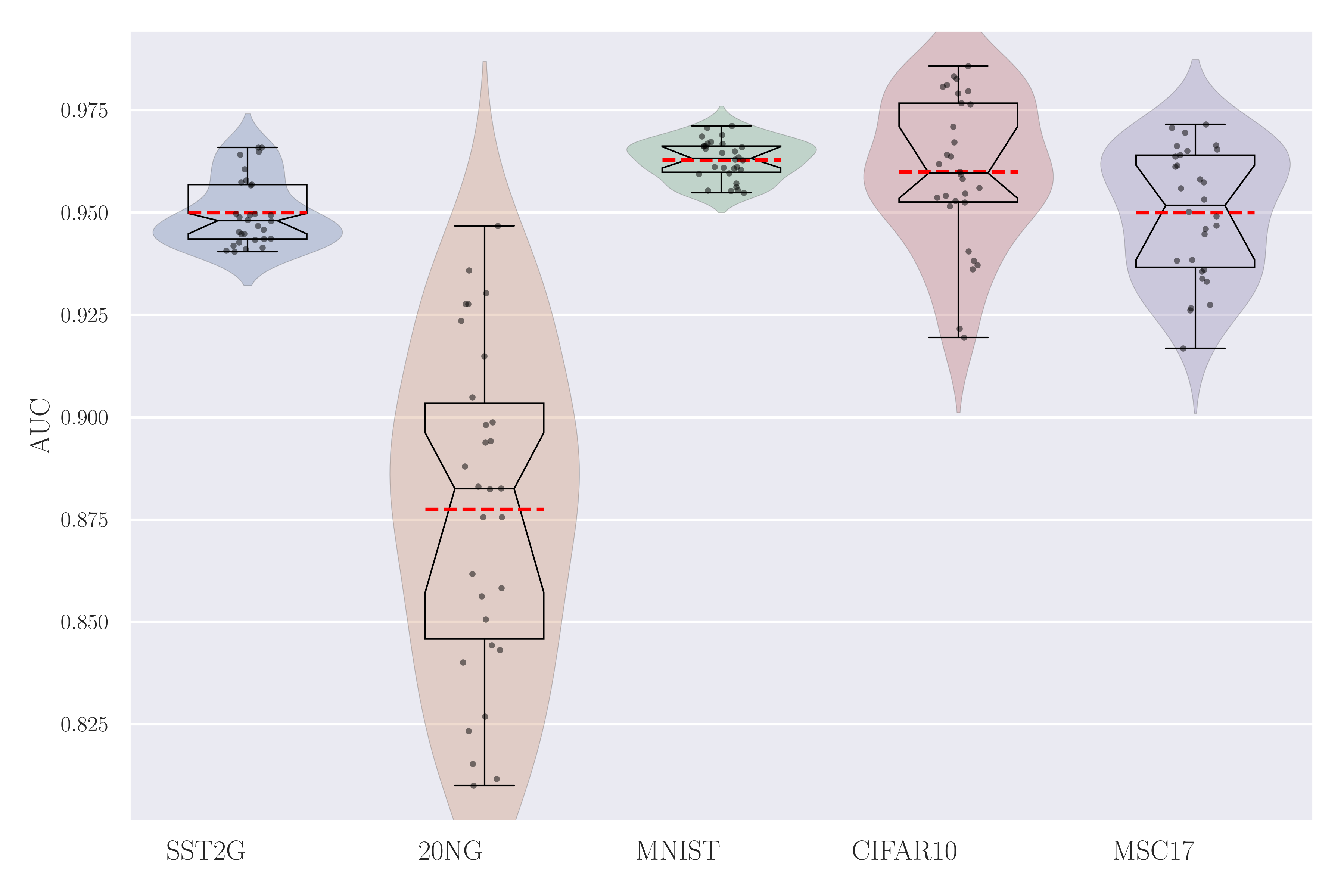}%
  \label{fig:metric_auc}}
\hfill
\subfloat[Brier score]{%
  \includegraphics[width=0.32\textwidth]{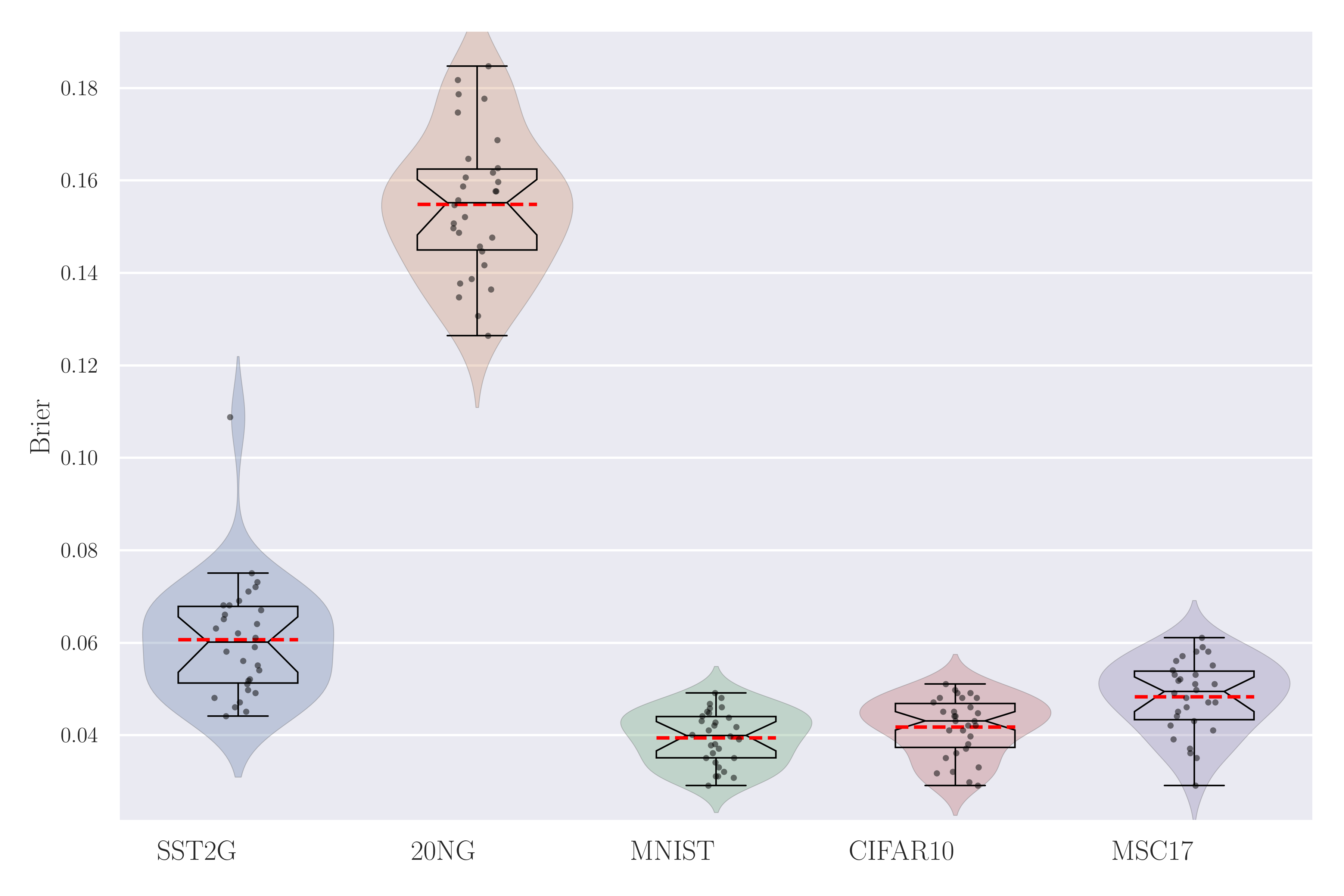}%
  \label{fig:metric_brier}}
\hfill
\subfloat[Complexity]{%
  \includegraphics[width=0.32\textwidth]{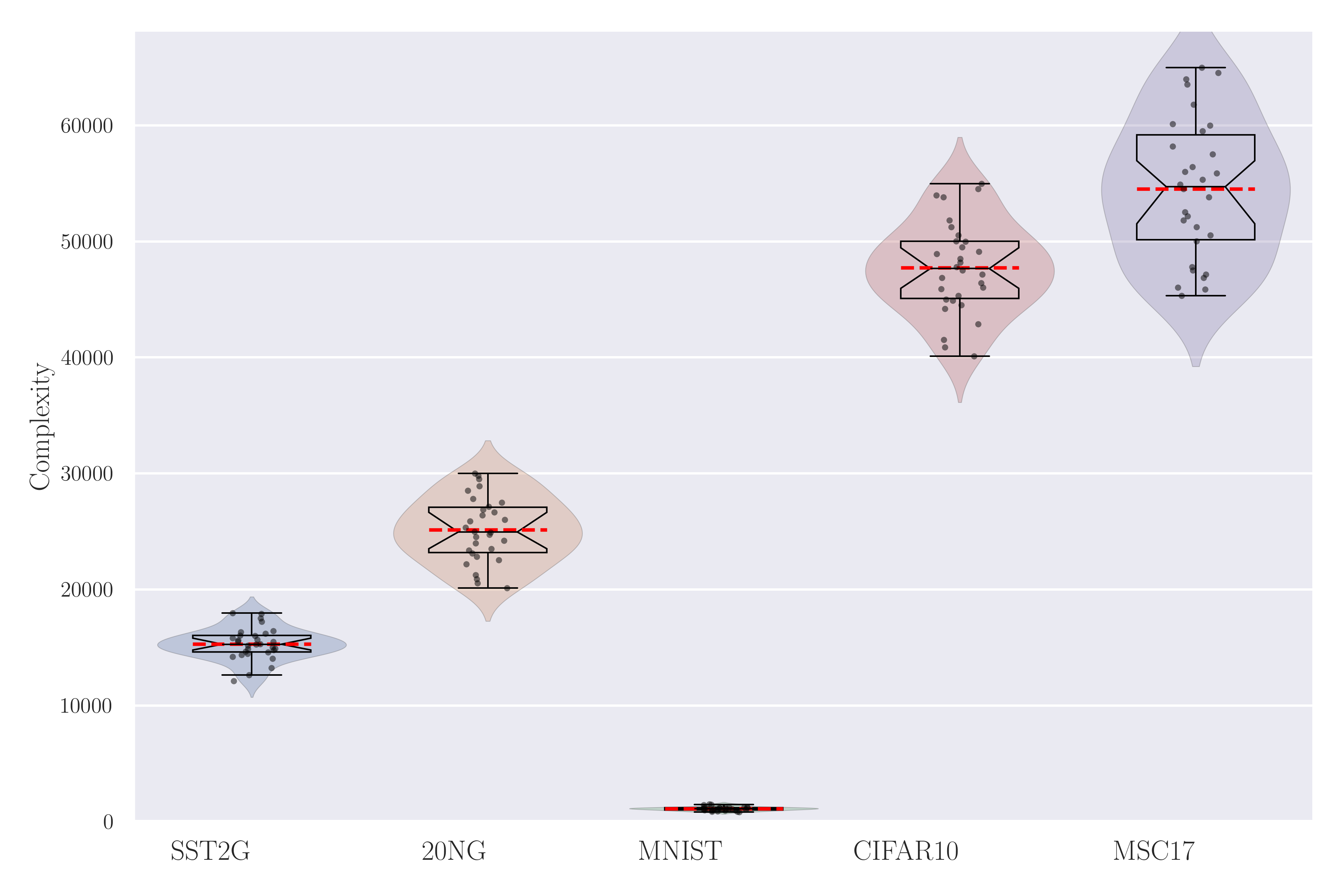}%
  \label{fig:metric_complexity}}

\vspace{0.7em}

\subfloat[ECE]{%
  \includegraphics[width=0.32\textwidth]{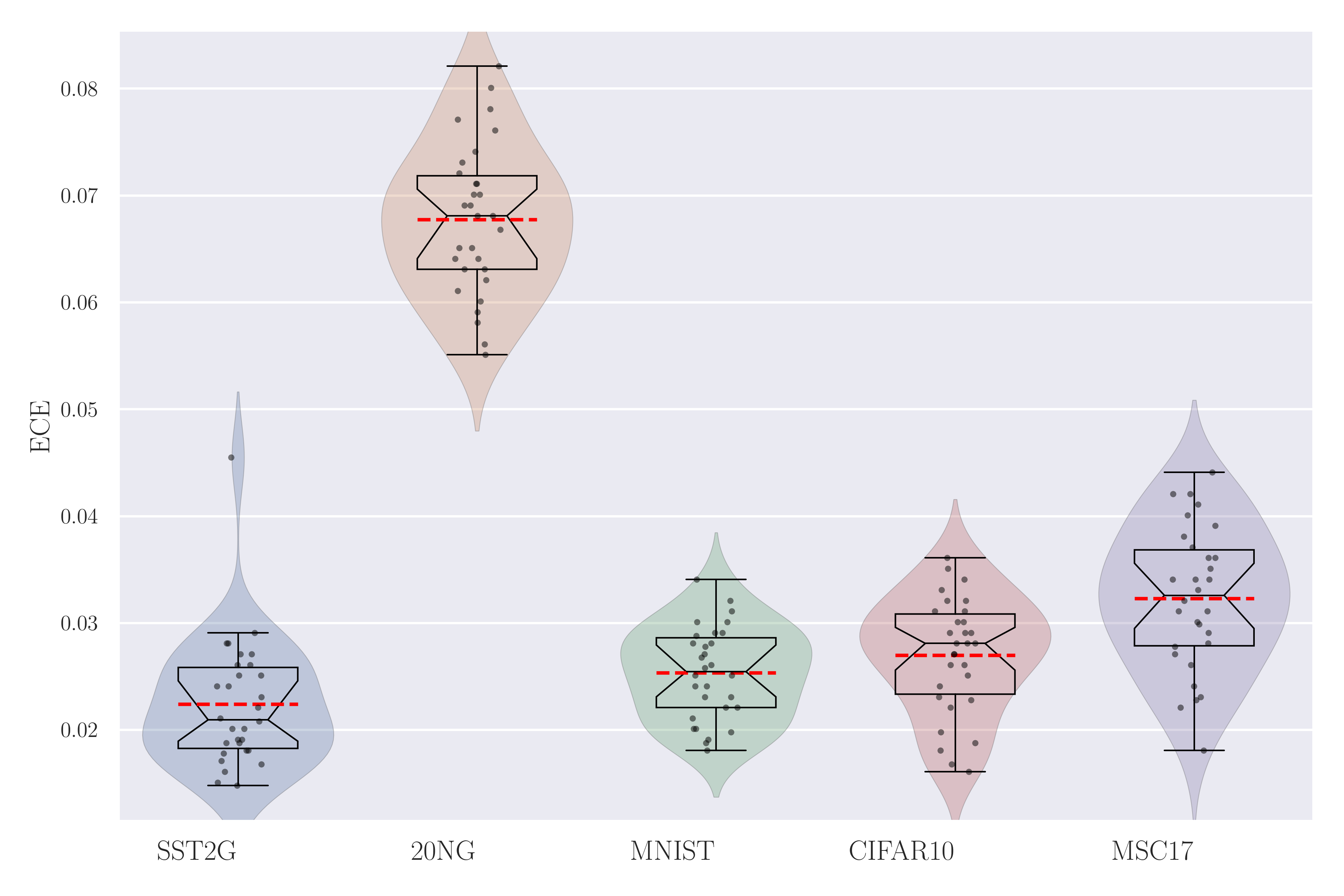}%
  \label{fig:metric_ece}}
\hfill
\subfloat[$F_1$ score]{%
  \includegraphics[width=0.32\textwidth]{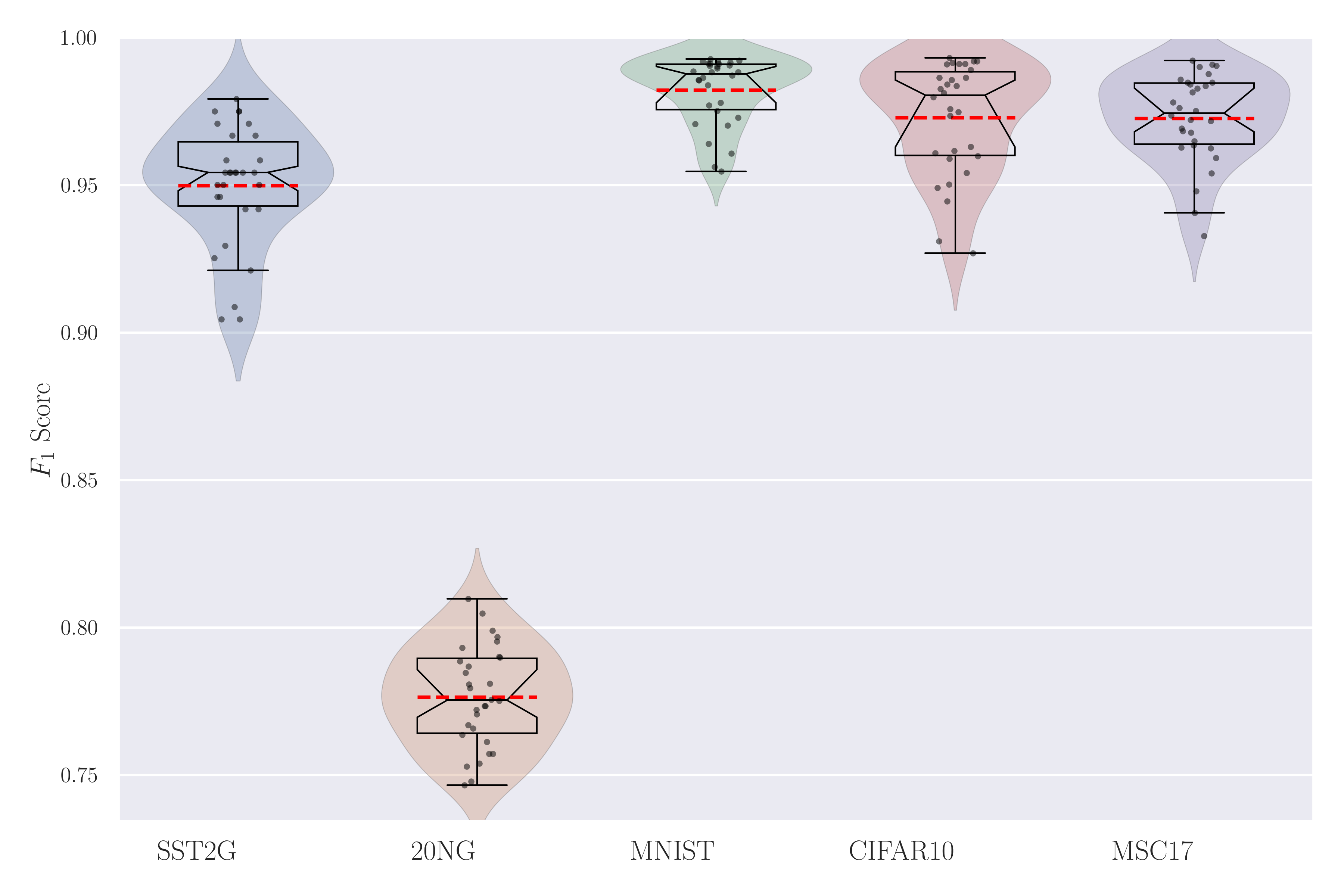}%
  \label{fig:metric_f1}}
\hfill
\subfloat[Log-loss]{%
  \includegraphics[width=0.32\textwidth]{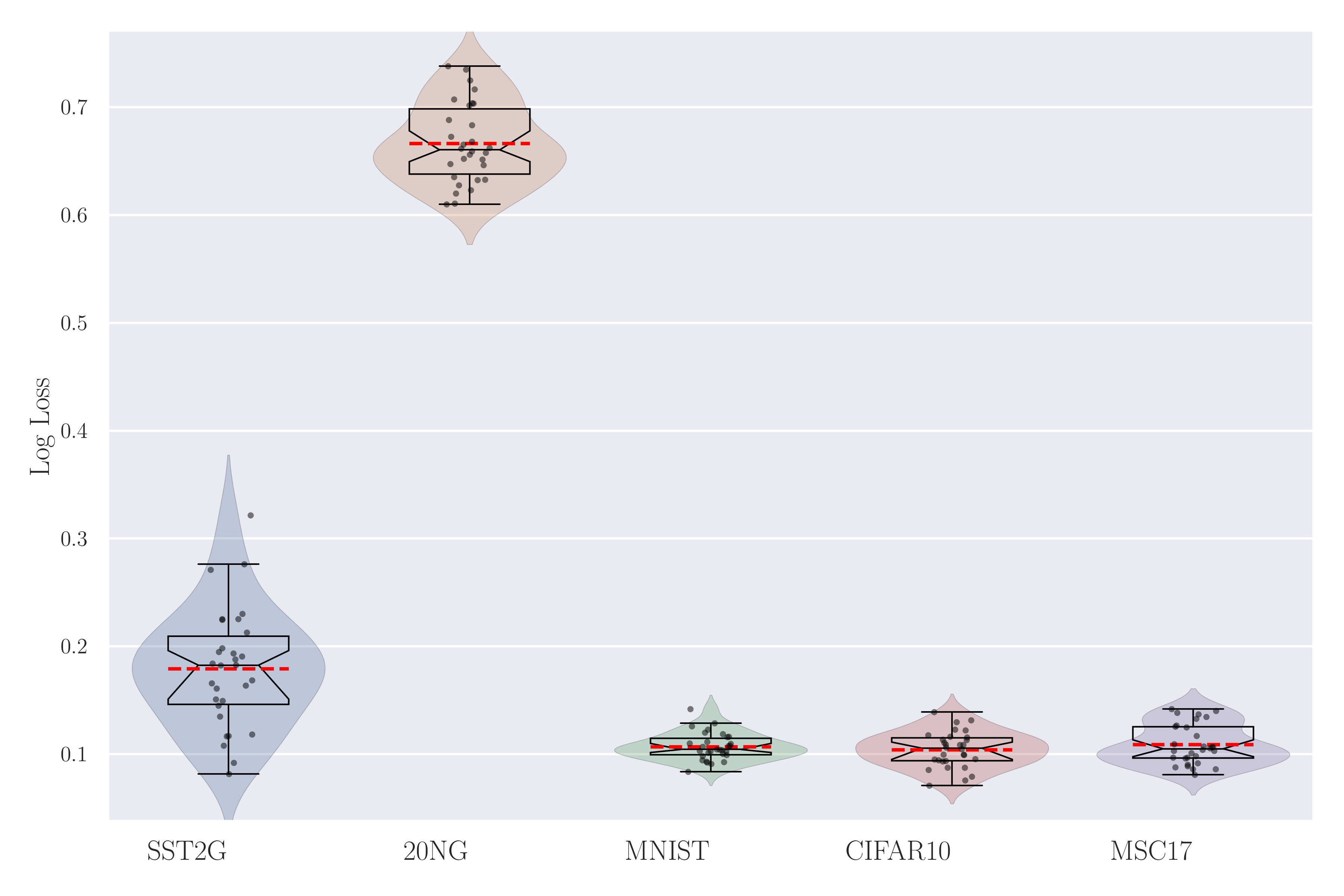}%
  \label{fig:metric_logloss}}

\caption{Performance metrics across 30 independent runs per dataset. Each panel shows a raincloud-style summary (violin, box, and jittered points) for: (a) AUC, (b) Brier score, (c) complexity, (d) ECE, (e) $F_1$ score, and (f) log-loss. Dashed red lines depict means; boxes show medians and interquartile ranges.}
\label{fig_metrics}
\end{figure*}

Across 30 independent runs per dataset, MEGP surrogates attain strong discrimination and calibration on four of the five benchmarks; full distributions appear in Fig.~\ref{fig_metrics} and numerical summaries in Table~\ref{tab_metric_results}.
MNIST, CIFAR10, and MSC17 reach macro-$F_{1}\!\in[0.97,0.99]$ with AUC $\approx0.95$–$0.96$, small ECE/Brier, and tight dispersion.
SST2G also performs well (median $F_{1}\!\approx\!0.95$, AUC $\approx0.95$).
20NG is the outlier, showing both the lowest central tendency ($F_{1}=0.776\pm0.006$, AUC $=0.878\pm0.015$) and the widest spread, consistent with its higher label cardinality and topic variability.

The SPFP partitions that drive MEGP’s view-wise coevolution are summarized in Table~\ref{tab_spfp}.
SST2G/MNIST/MSC17 use $V=4$ views, 20NG uses $V=3$, and CIFAR10 uses $V=5$, with total selected dimensions close to the $0.1d$ budget and mildly unbalanced per-view allocations.
These structural choices explain much of the complexity profile in Fig.~\ref{fig_metrics}(c): despite having $K=10$ classes, MNIST remains extremely compact ($\sim\!1.1\times10^3$ nodes) because its selected dimensions and views are small, whereas CIFAR10 ($V=5$) and MSC17 (higher $d$ and multimodality) yield substantially larger programs ($\sim\!4.8$–$5.5\times10^4$ nodes).
SST2G and 20NG fall in between.

\begin{figure*}[ht]
\centering
\begin{subfigure}[t]{0.32\textwidth}
  \includegraphics[width=\linewidth]{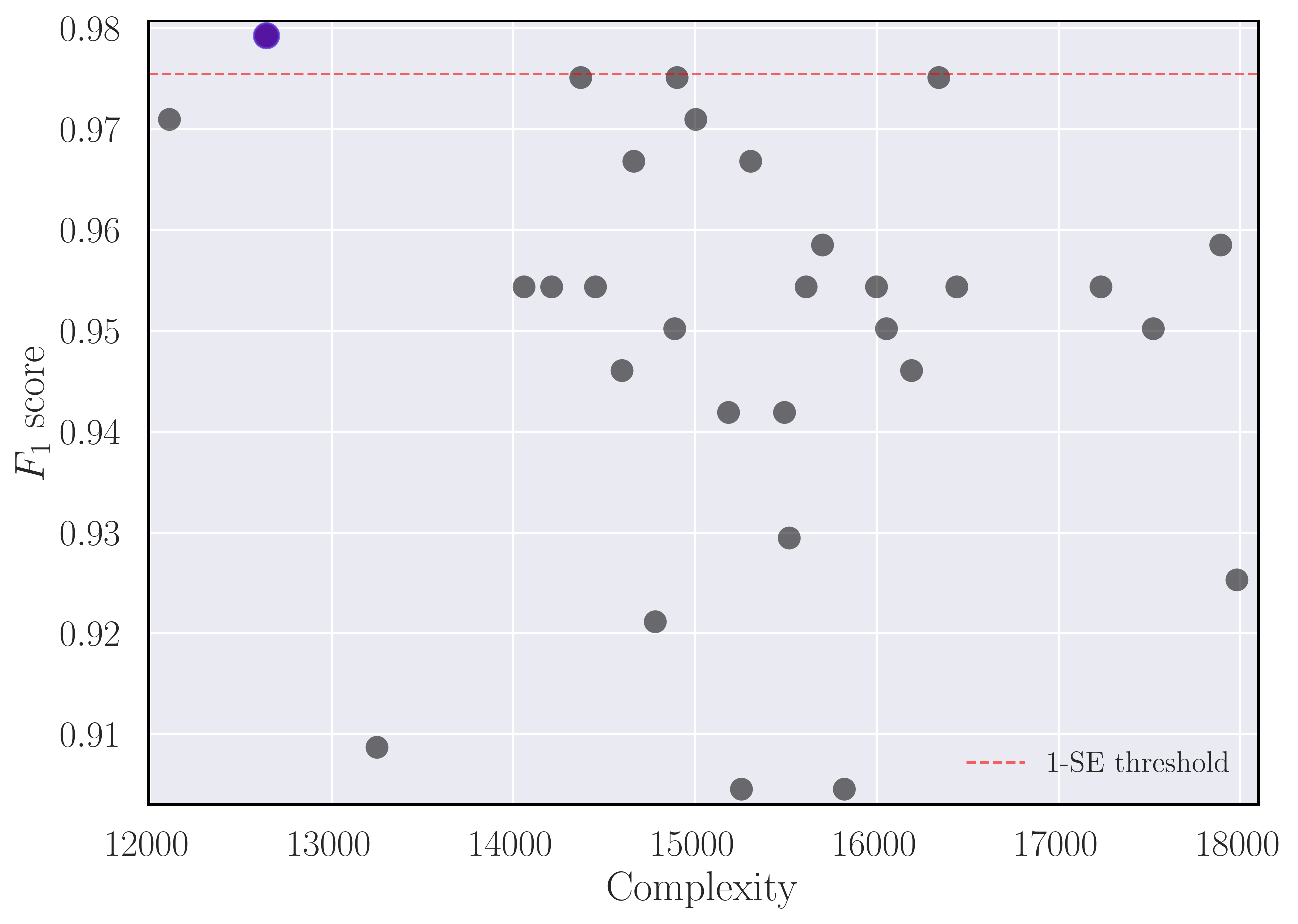}
  \caption{SST2G}\label{fig_scatter_sst2g}
\end{subfigure}\hfill
\begin{subfigure}[t]{0.32\textwidth}
  \includegraphics[width=\linewidth]{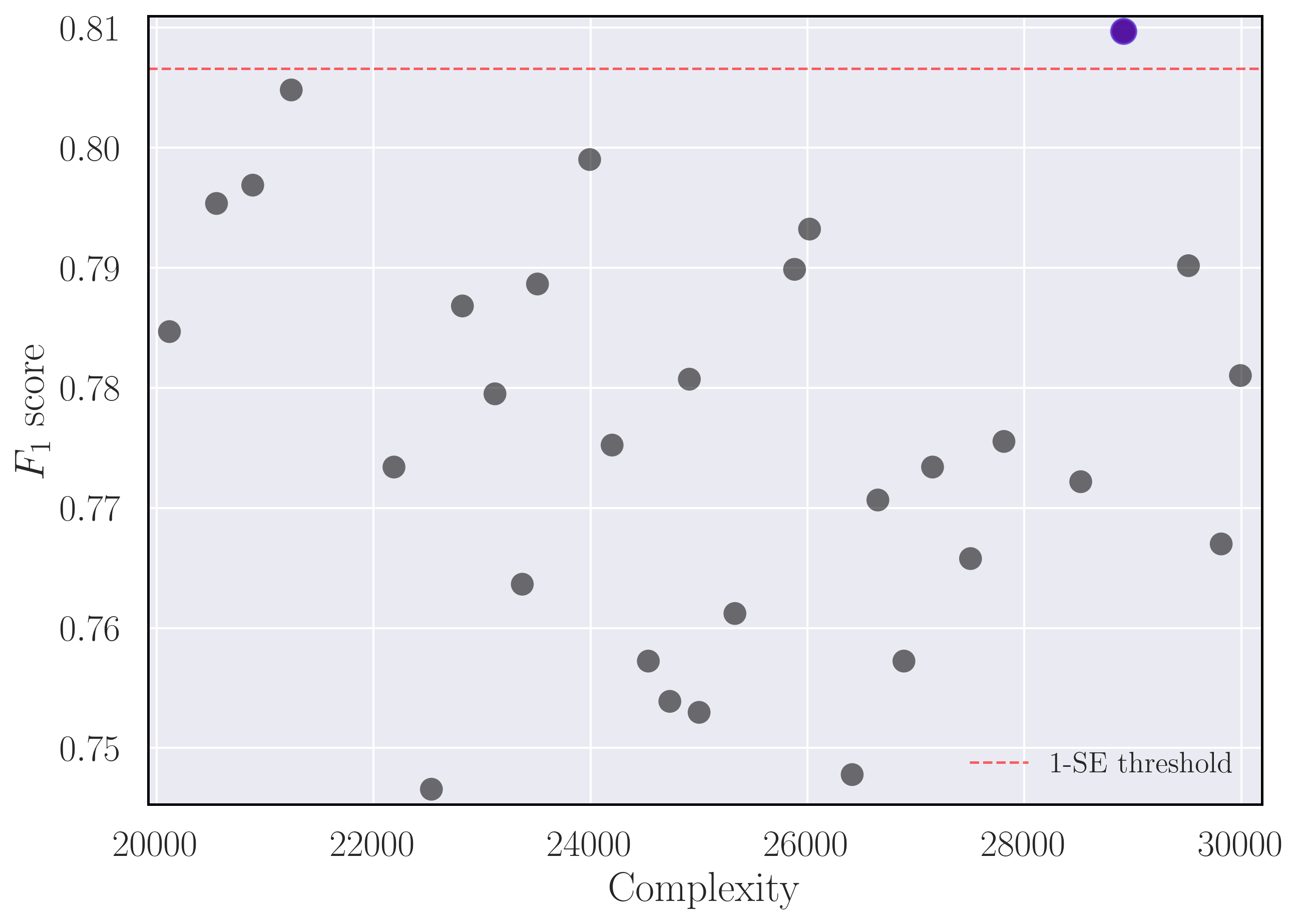}
  \caption{20NG}\label{fig_scatter_20ng}
\end{subfigure}\hfill
\begin{subfigure}[t]{0.32\textwidth}
  \includegraphics[width=\linewidth]{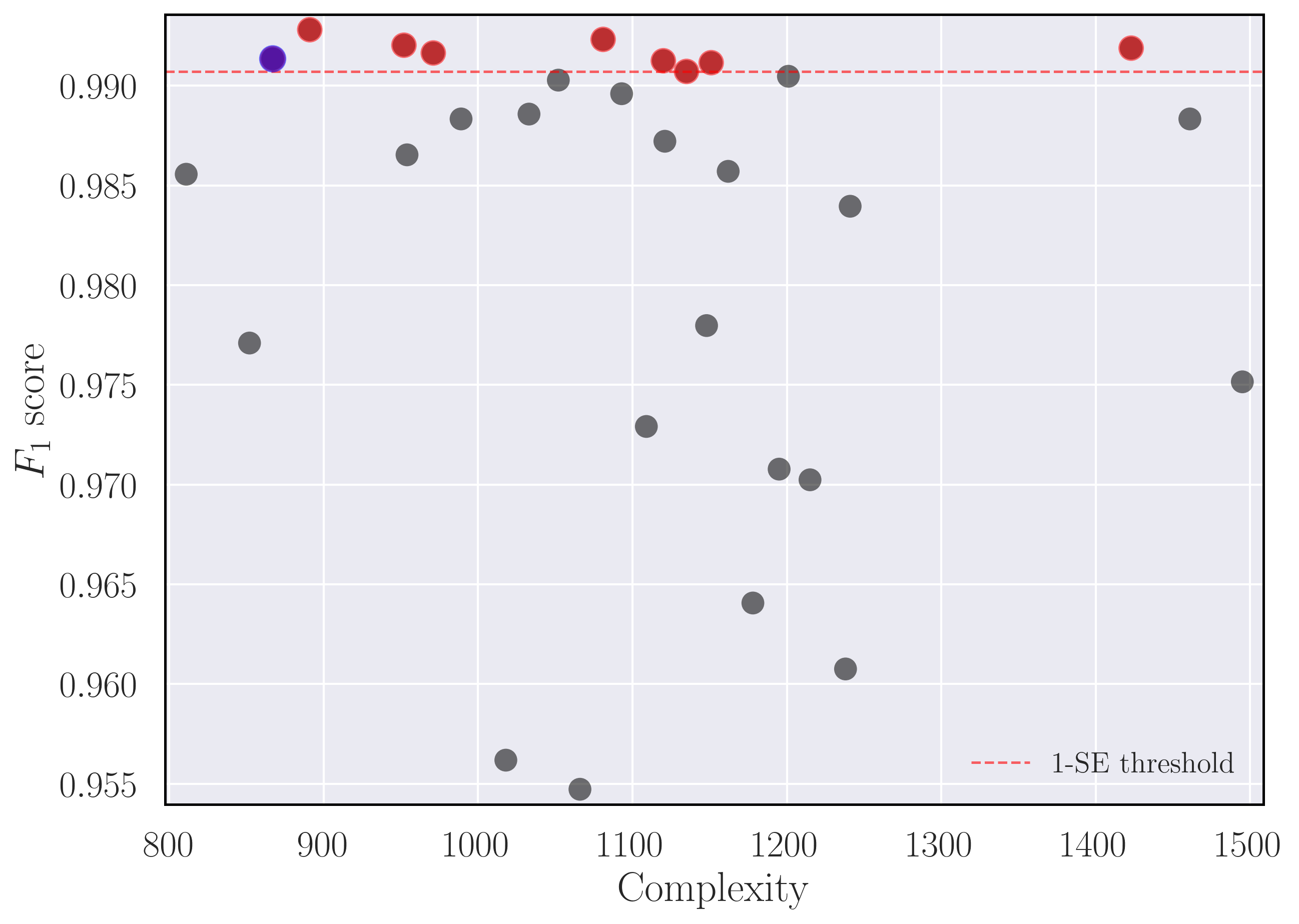}
  \caption{MNIST}\label{fig_scatter_mnist}
\end{subfigure}

\par\vspace{0.9em}

\hspace{\fill}
\begin{subfigure}[t]{0.36\textwidth}
  \includegraphics[width=\linewidth]{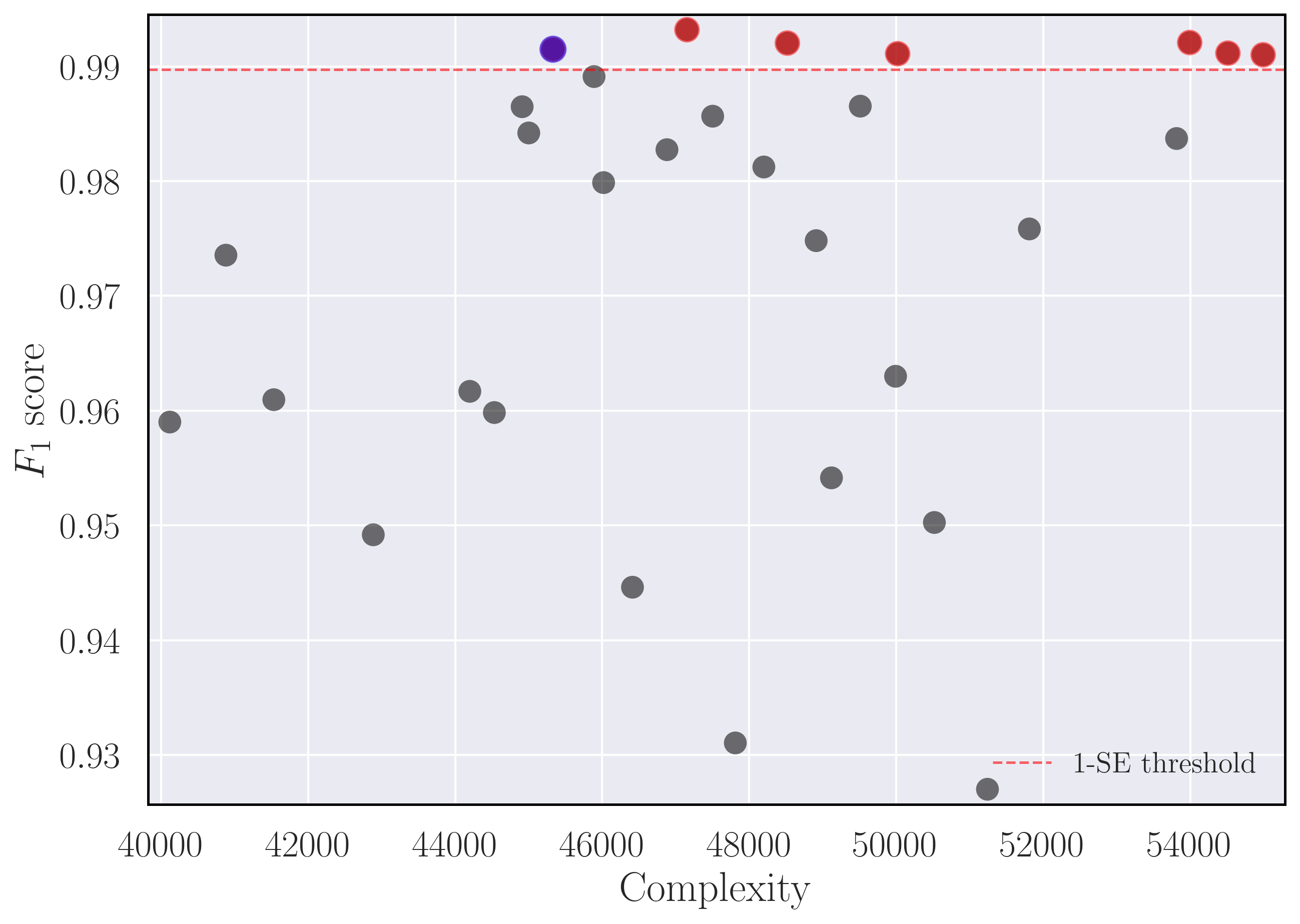}
  \caption{CIFAR10}\label{fig_scatter_cifar10}
\end{subfigure}
\hspace{0.02\textwidth}
\begin{subfigure}[t]{0.36\textwidth}
  \includegraphics[width=\linewidth]{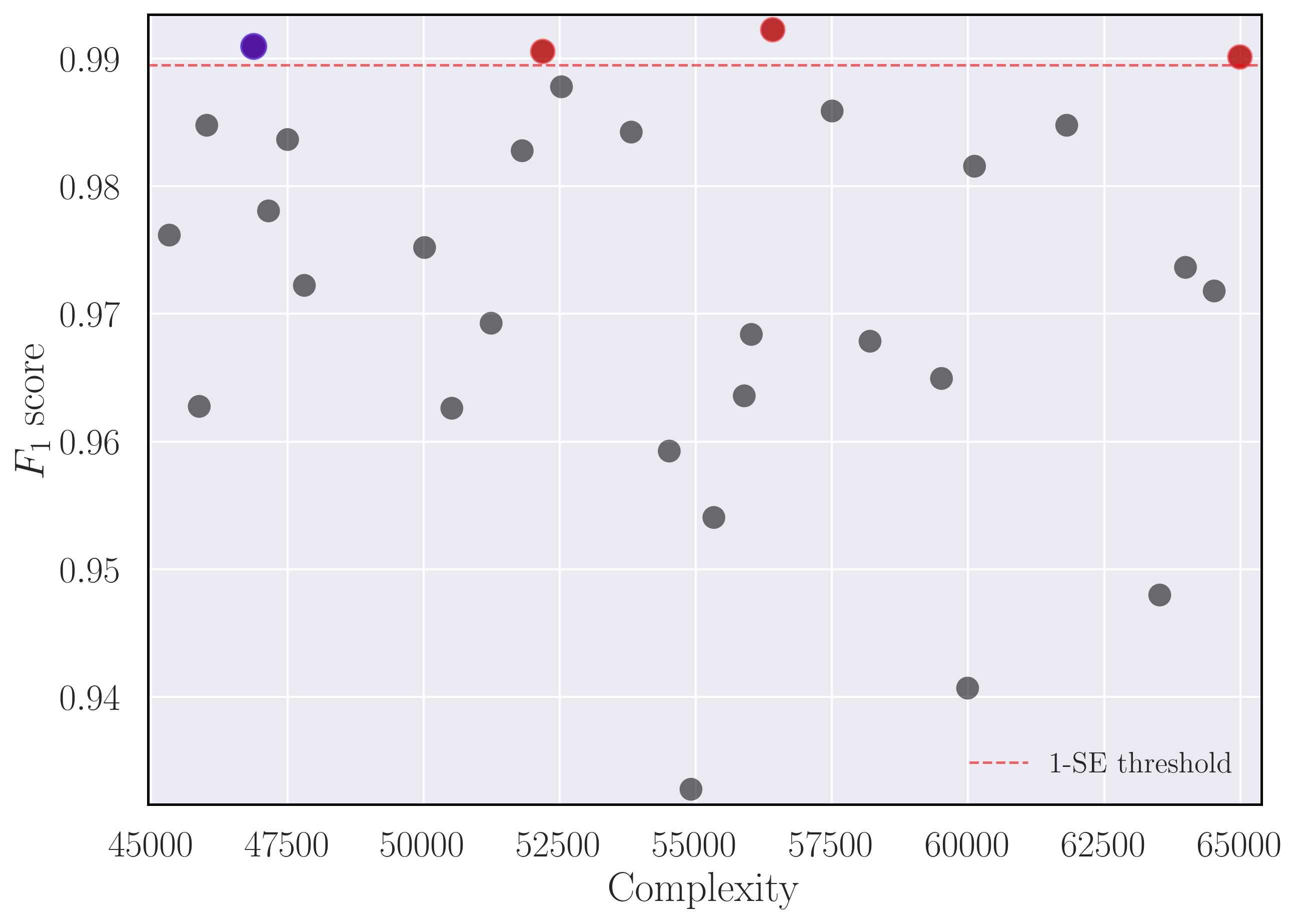}
  \caption{MSC17}\label{fig_scatter_msc17}
\end{subfigure}
\hspace{\fill}

\caption{Canonical model selection: validation \(F_1\) vs.\ model complexity across 30 runs per dataset (SST2G, 20NG, MNIST, CIFAR10, MSC17). The dashed line marks the 1-SE threshold; red points indicate models within 1-SE of the best \(F_1\), with the chosen canonical model highlighted.}
\label{fig_scatters}
\end{figure*}

Fig.~\ref{fig_scatters} shows the validation $F_{1}$–versus–complexity landscape used by the 1-SE selection rule.
For MNIST, many runs sit within one standard error of the best score at very low complexity; the canonical model therefore lies near the left edge (867 nodes; Table~\ref{tab_temp_scaling}).
CIFAR10 and MSC17 also offer numerous near-optimal points, and the canonical choices trade a negligible $F_{1}$ drop for sizable parsimony (45{,}331 and 46{,}881 nodes).
SST2G exhibits a similar pattern (12{,}642 nodes).
In 20NG, the 1-SE band is tight and concentrates feasible points near the top of $F_{1}$, yielding a higher-complexity canonical model (28{,}915).
Overall, the 1-SE criterion consistently returns compact surrogates with test performance close to the best observed models (Table~\ref{tab_temp_scaling}), aligning with the parsimony goal of symbolic surrogate modeling.

\begin{table*}[t]
\centering
\footnotesize
\setlength{\tabcolsep}{6pt}
\renewcommand{\arraystretch}{1.15}
\caption{Mean $\pm$ 95\% CI (t$_{29}$) over 30 runs per dataset and metric.}
\label{tab_metric_results}
\begin{tabular}{l|cccccc}
\hline
\textbf{Dataset} & $\mathbf{F_1}$ & \textbf{AUC} & \textbf{ECE} & \textbf{Brier} & \textbf{Log-Loss} & \textbf{Complexity} \\
\hline
SST2G   & 0.950 $\pm$ 0.008 & 0.950 $\pm$ 0.003 & 0.022 $\pm$ 0.002 & 0.061 $\pm$ 0.005 & 0.179 $\pm$ 0.021 & 15{,}316 $\pm$ 513 \\
20NG    & 0.776 $\pm$ 0.006 & 0.878 $\pm$ 0.015 & 0.068 $\pm$ 0.003 & 0.155 $\pm$ 0.006 & 0.667 $\pm$ 0.014 & 25{,}137 $\pm$ 1{,}048 \\
MNIST   & 0.982 $\pm$ 0.004 & 0.963 $\pm$ 0.002 & 0.025 $\pm$ 0.002 & 0.039 $\pm$ 0.002 & 0.107 $\pm$ 0.005 & 1{,}107 $\pm$ 62 \\
CIFAR10 & 0.973 $\pm$ 0.007 & 0.960 $\pm$ 0.007 & 0.027 $\pm$ 0.002 & 0.042 $\pm$ 0.002 & 0.104 $\pm$ 0.006 & 47{,}740 $\pm$ 1{,}469 \\
MSC17   & 0.973 $\pm$ 0.006 & 0.950 $\pm$ 0.006 & 0.032 $\pm$ 0.002 & 0.048 $\pm$ 0.003 & 0.109 $\pm$ 0.007 & 54{,}531 $\pm$ 2{,}250 \\
\hline
\end{tabular}
\end{table*}

\begin{table*}[t]
\centering
\small
\setlength{\tabcolsep}{6pt}
\renewcommand{\arraystretch}{1.15}
\caption{Results of the SPFP procedure executed on the \emph{training set}. For each dataset we report the embedding dimension $d$, the number of views, the total number of selected features (mean $\pm$ s.d.), and the per-view allocation.}
\label{tab_spfp}
\begin{tabular}{l|c c c l}
\hline
\textbf{Dataset} & \textbf{Embedding Dimensions} & \textbf{\# Views} & \textbf{\#Features} & \textbf{Dimensions per View} \\
\hline
SST2G   & 768  & 4 & $192 \pm 8.5$    & \{178, 195, 201, 194\} \\
20NG    & 768  & 3 & $256 \pm 16.5$   & \{233, 264, 271\} \\
MNIST   & 1024 & 4 & $256 \pm 25.4$   & \{220, 251, 262, 291\} \\
CIFAR10 & 1024 & 5 & $204.8 \pm 19.5$ & \{181, 190, 199, 220, 234\} \\
MSC17   & 1152 & 4 & $288 \pm 24.7$   & \{254, 282, 293, 323\} \\
\hline
\end{tabular}
\end{table*}

\begin{table*}[t]
\centering
\footnotesize
\setlength{\tabcolsep}{6pt}
\renewcommand{\arraystretch}{1.15}
\caption{Canonical GP models: test-set metrics before vs.\ after temperature scaling (\(T\) learned on validation, applied to test). \(F_1\) and AUC are unchanged by calibration.}
\label{tab_temp_scaling}
\begin{tabular}{l|ccccccc}
\hline
\textbf{Dataset} & $\mathbf{F_1}$ & \textbf{AUC} & \textbf{Complexity} & \textbf{Log-Loss (pre $\Rightarrow$ post)} & \textbf{Brier (pre $\Rightarrow$ post)} & \textbf{ECE (pre $\Rightarrow$ post)} & $\mathbf{T}$ \\
\hline
SST2G   & 0.979 & 0.941 & 12{,}642 & $0.092 \Rightarrow 0.090$ & $0.054 \Rightarrow 0.049$ & $0.019 \Rightarrow 0.008$ & 1.10 \\
20NG    & 0.810 & 0.888 & 28{,}915 & $0.620 \Rightarrow 0.588$ & $0.136 \Rightarrow 0.129$ & $0.058 \Rightarrow 0.014$ & 1.85 \\
MNIST   & 0.991 & 0.964 & 867      & $0.120 \Rightarrow 0.115$ & $0.044 \Rightarrow 0.040$ & $0.029 \Rightarrow 0.010$ & 1.20 \\
CIFAR10 & 0.991 & 0.959 & 45{,}331 & $0.123 \Rightarrow 0.115$ & $0.038 \Rightarrow 0.037$ & $0.024 \Rightarrow 0.008$ & 1.25 \\
MSC17   & 0.991 & 0.933 & 46{,}881 & $0.103 \Rightarrow 0.101$ & $0.035 \Rightarrow 0.034$ & $0.022 \Rightarrow 0.009$ & 1.15 \\
\hline
\end{tabular}
\end{table*}

\subsection{Calibration quality}\label{subsec:results-calibration}
\begin{figure*}[ht]
\centering
\begin{subfigure}[t]{0.32\textwidth}
  \includegraphics[width=\linewidth]{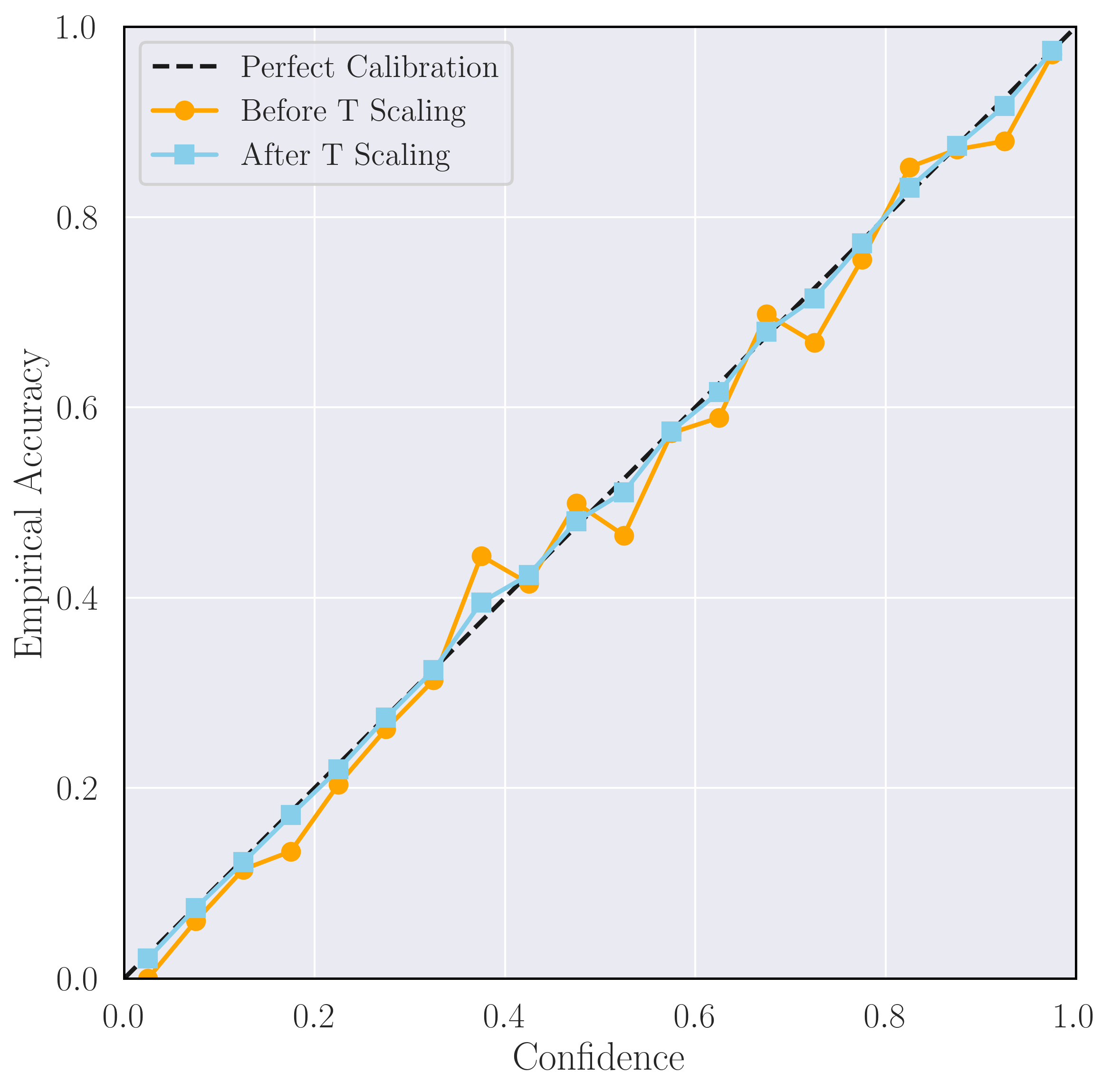}
  \caption{SST2G}\label{fig_cal_sst2g}
\end{subfigure}\hfill
\begin{subfigure}[t]{0.32\textwidth}
  \includegraphics[width=\linewidth]{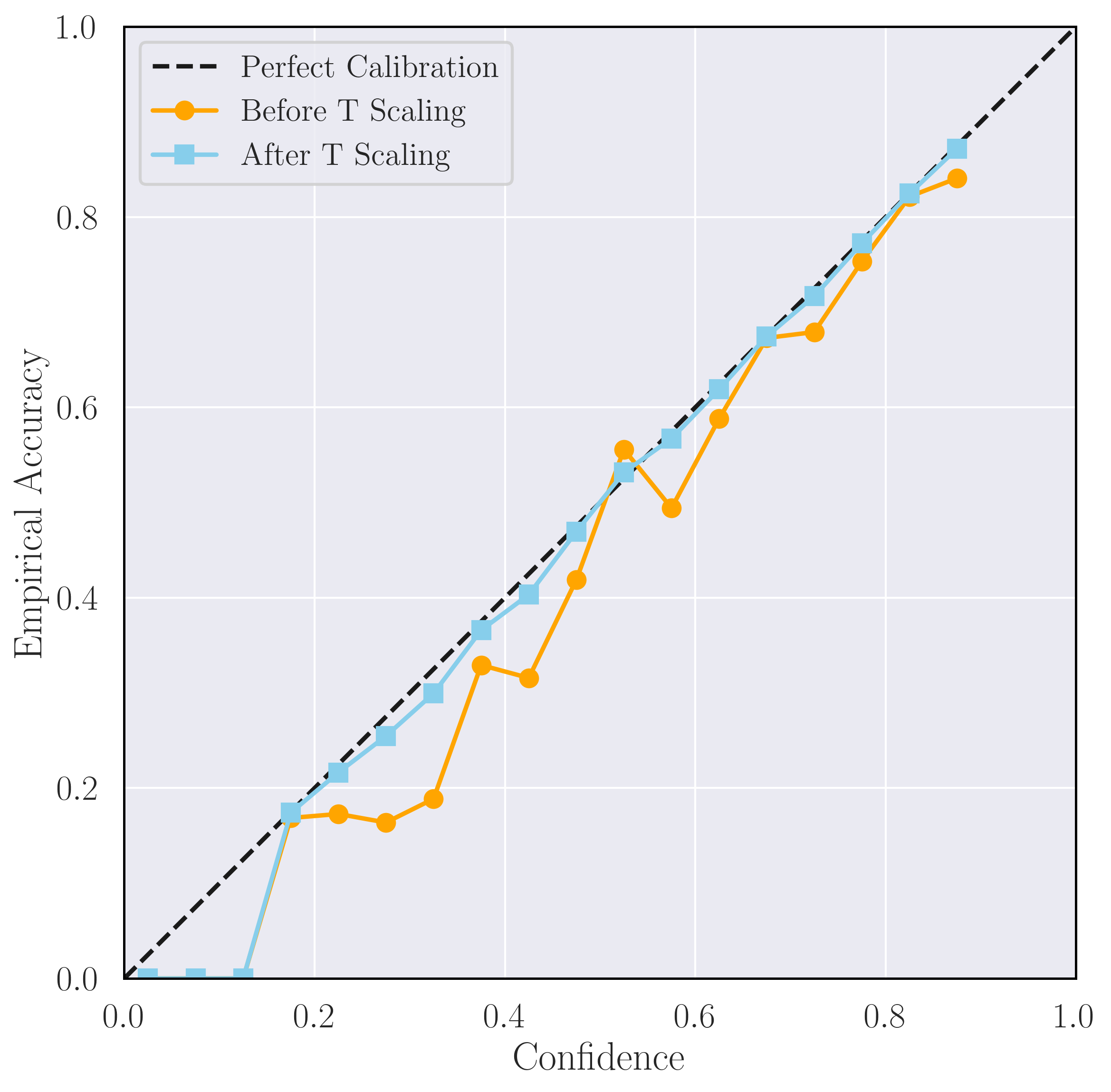}
  \caption{20NG}\label{fig_cal_20ng}
\end{subfigure}\hfill
\begin{subfigure}[t]{0.32\textwidth}
  \includegraphics[width=\linewidth]{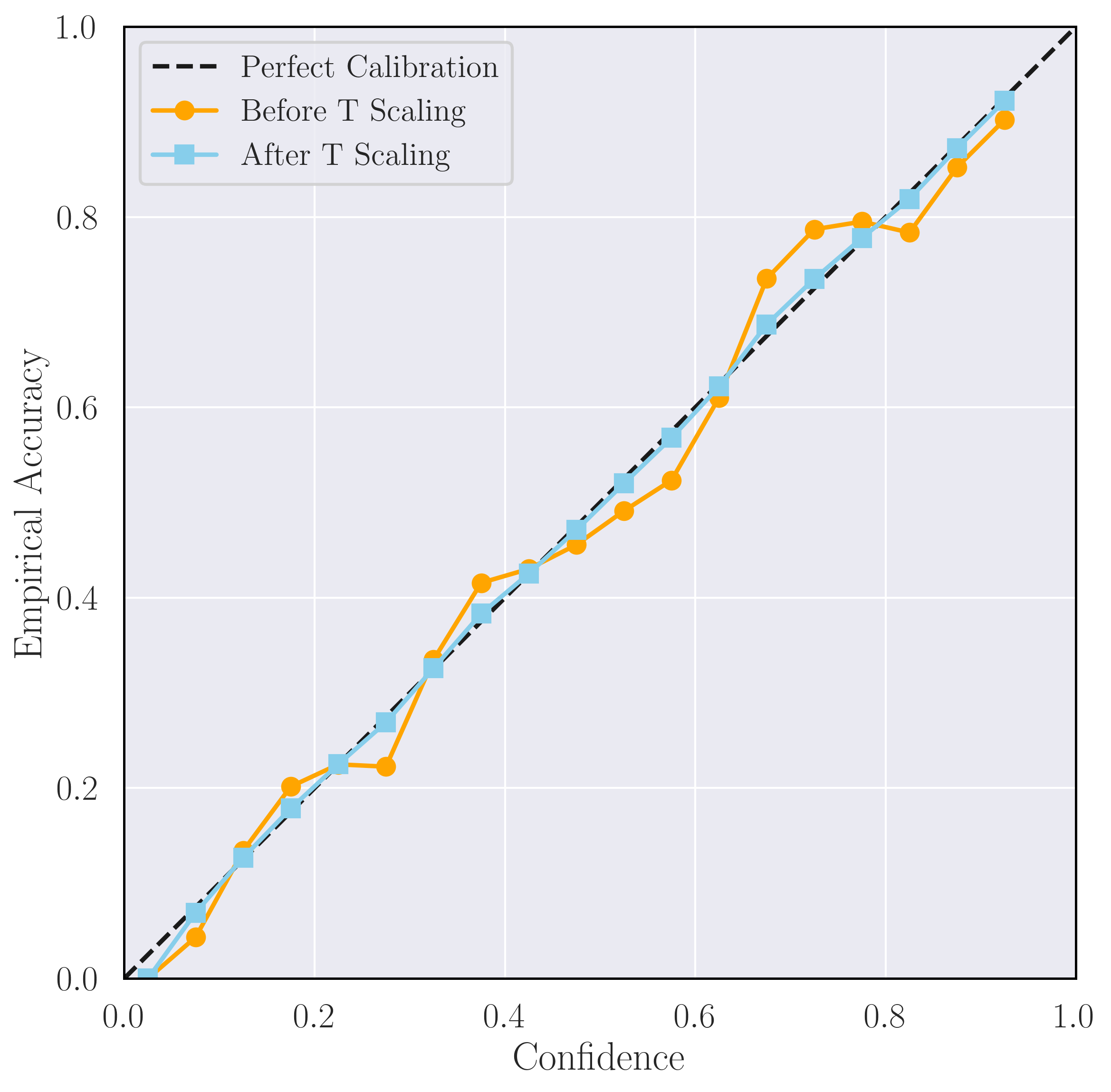}
  \caption{MNIST}\label{fig_cal_mnist}
\end{subfigure}

\par\vspace{0.9em} 

\hspace{\fill}
\begin{subfigure}[t]{0.36\textwidth}
  \includegraphics[width=\linewidth]{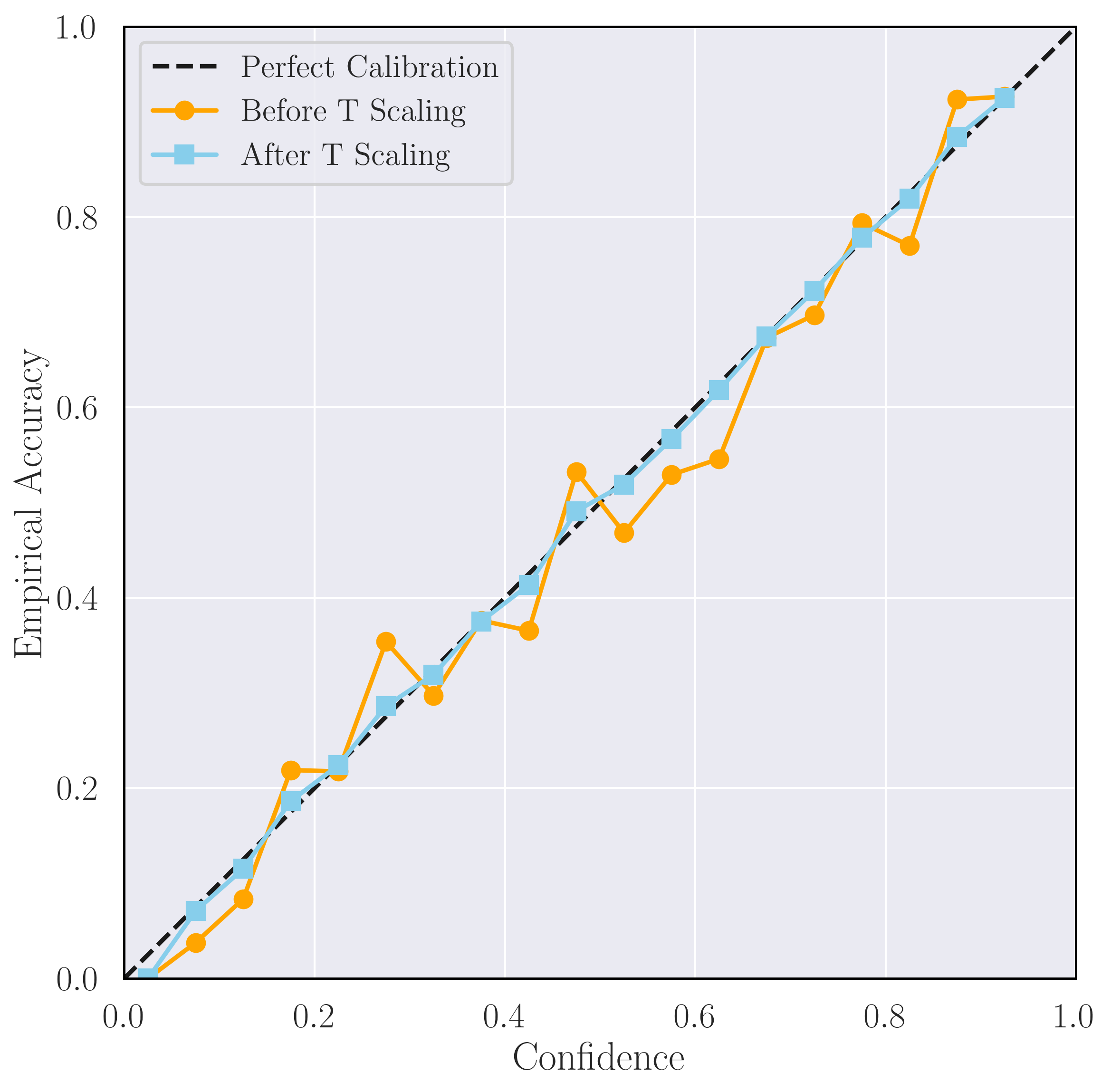}
  \caption{CIFAR10}\label{fig_cal_cifar10}
\end{subfigure}
\hspace{0.02\textwidth}
\begin{subfigure}[t]{0.36\textwidth}
  \includegraphics[width=\linewidth]{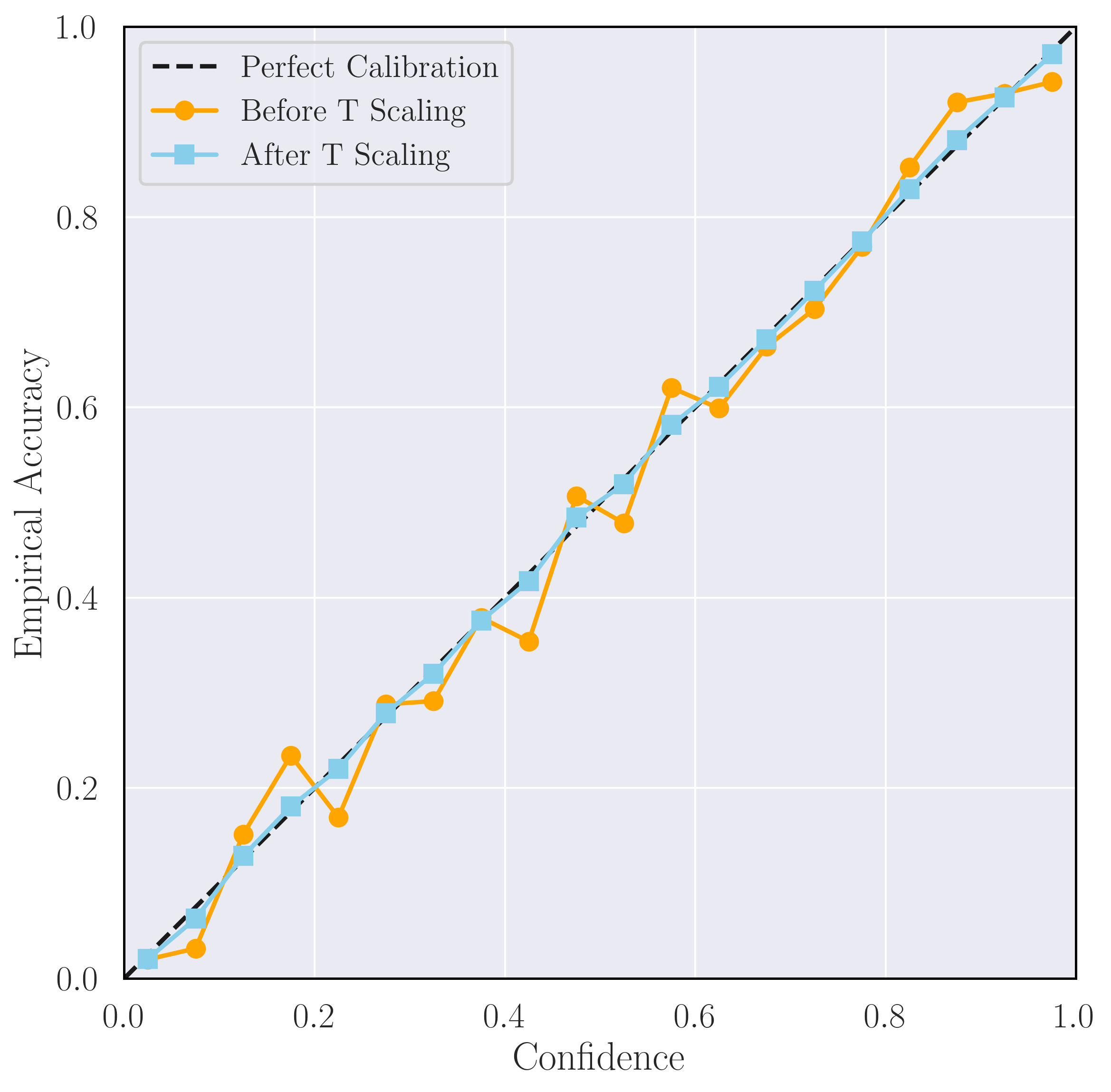}
  \caption{MSC17}\label{fig_cal_msc17}
\end{subfigure}
\hspace{\fill}

\caption{Reliability diagrams (test) before (\textcolor{orange}{orange}) and after (\textcolor{blue}{blue}) temperature scaling. Each panel shows empirical accuracy vs.\ predicted confidence using 20 equal-width bins; empty bins omitted; probabilities clipped to \([\,\varepsilon,1-\varepsilon\,]\). The dashed line denotes perfect calibration; \(T\) fitted on validation and applied to test.}
\label{fig_calibration}
\end{figure*}

Temperature scaling fitted on validation substantially improves the probability calibration of the canonical MEGP surrogates on test (Fig.~\ref{fig_calibration}, Table~\ref{tab_temp_scaling}). The learned temperatures are \(T>1\) in all cases (SST2G 1.10, 20NG 1.85, MNIST 1.20, CIFAR10 1.25, MSC17 1.15), indicating overconfident raw logits mitigated by dividing by \(T\) before the softmax. Post-scaling reliability curves align closely with the identity, with the largest corrections in mid-confidence bins \(p\in[0.4,0.7]\).

Quantitatively, expected calibration error (ECE) drops markedly:
SST2G \(0.019 \to 0.008\) \((-58\%)\),
20NG \(0.058 \to 0.014\) \((-76\%)\),
MNIST \(0.029 \to 0.010\) \((-66\%)\),
CIFAR10 \(0.024 \to 0.008\) \((-67\%)\),
MSC17 \(0.022 \to 0.009\) \((-59\%)\).
Brier scores also decrease (absolute \(\Delta\) of \(-0.005\), \(-0.007\), \(-0.004\), \(-0.001\), \(-0.001\), respectively), as do log-loss values (\(-0.002\), \(-0.032\), \(-0.005\), \(-0.008\), \(-0.002\)). Because temperature scaling is a monotone reparameterization of logits, discrimination is preserved: \(F_1\) and AUC in Table~\ref{tab_temp_scaling} match the pre-calibration values (see distributions in Fig.~\ref{fig_metrics}). Overall, temperature scaling yields well-calibrated posteriors without sacrificing accuracy, with the most pronounced gains on 20NG where class granularity and topic heterogeneity exacerbate miscalibration.

\subsection{Dimension usage and inter–logit overlap}\label{subsec:dim-overlap}
SPFP produces $V$ disjoint views per dataset (Table~\ref{tab_spfp}); MEGP subsequently induces sparse, reusable subsets of embedding dimensions within those views. The canonical models use only a small fraction of the available coordinates (Table~\ref{tab_symbol_stats}): SST2G $31/768\!\approx\!4.0\%$, 20NG $113/768\!\approx\!14.7\%$, MNIST $17/1024\!\approx\!1.7\%$, CIFAR10 $76/1024\!\approx\!7.4\%$, MSC17 $34/1152\!\approx\!3.0\%$. Median dimensions per logit are modest (SST2G 21, 20NG 19, MNIST 4, CIFAR10 12.5, MSC17 19), consistent with the parsimony reported earlier.

\begin{figure*}[ht]
\centering
\captionsetup[sub]{justification=centering}
\begin{minipage}[b]{0.32\textwidth}
  \includegraphics[width=\linewidth]{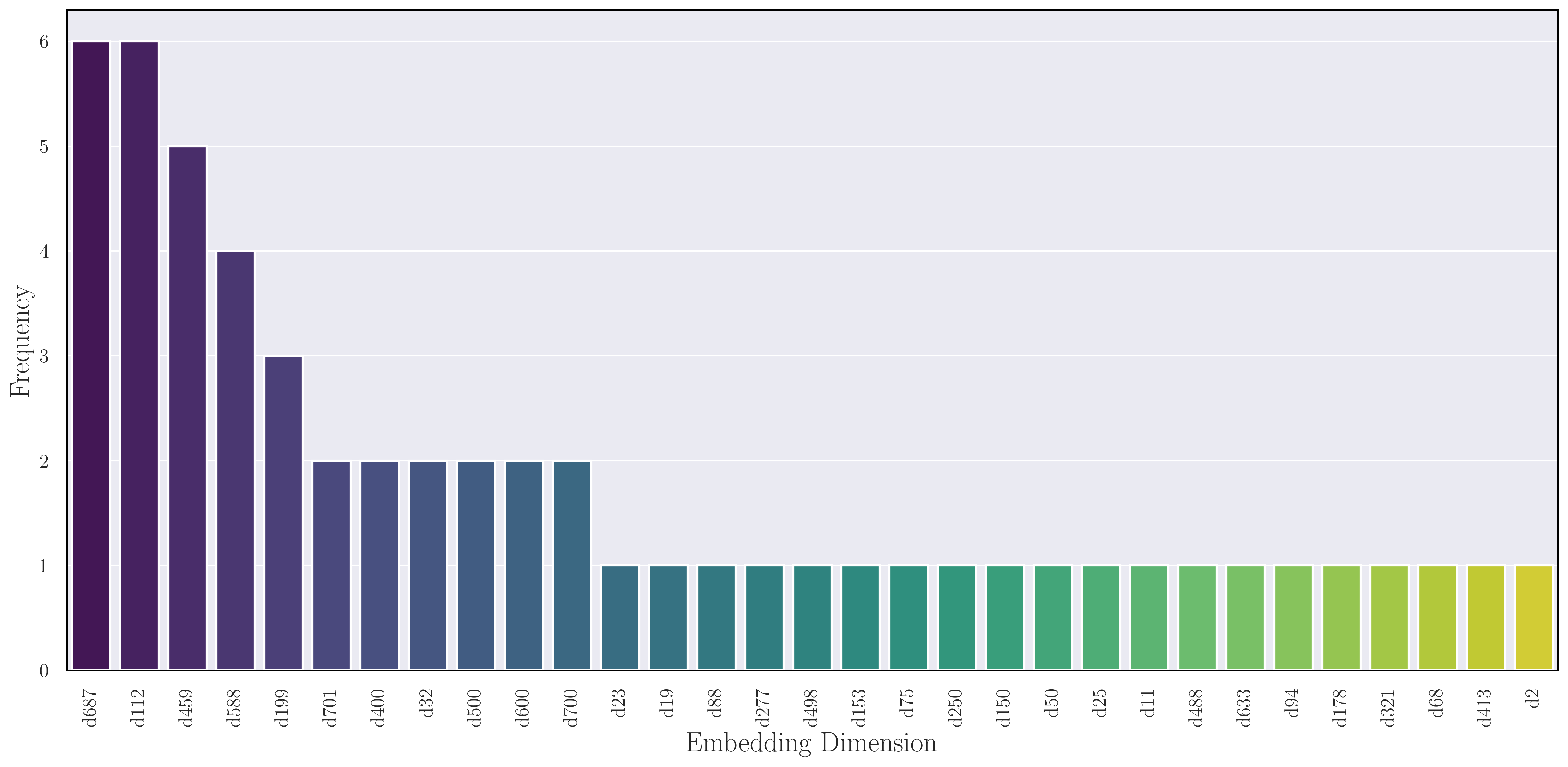}
  \subcaption{SST2G}
  \label{subfig:hist_sst2g}
\end{minipage}\hfill
\begin{minipage}[b]{0.32\textwidth}
  \includegraphics[width=\linewidth]{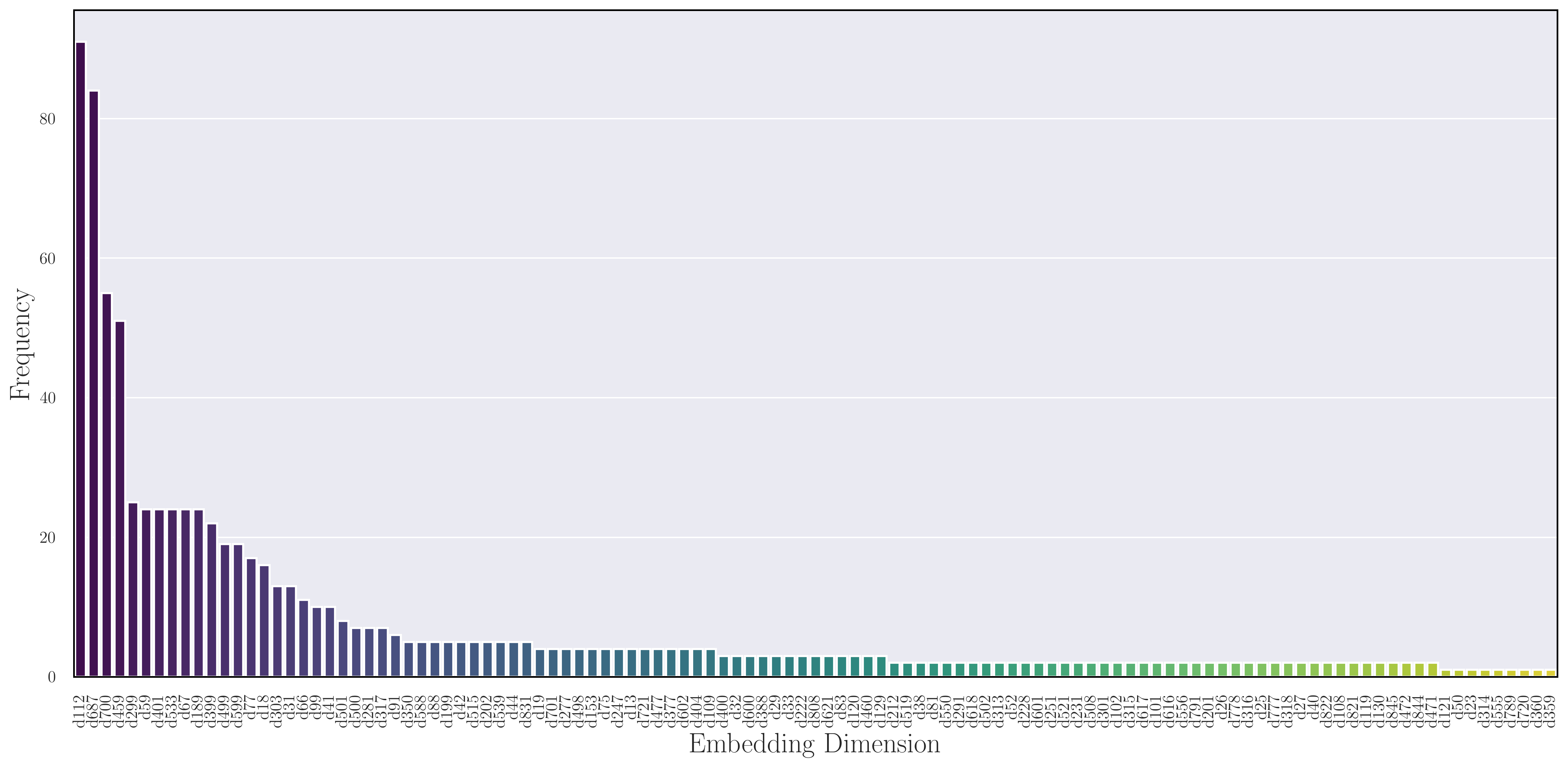}
  \subcaption{20NG}
  \label{subfig:hist_20ng}
\end{minipage}\hfill
\begin{minipage}[b]{0.32\textwidth}
  \includegraphics[width=\linewidth]{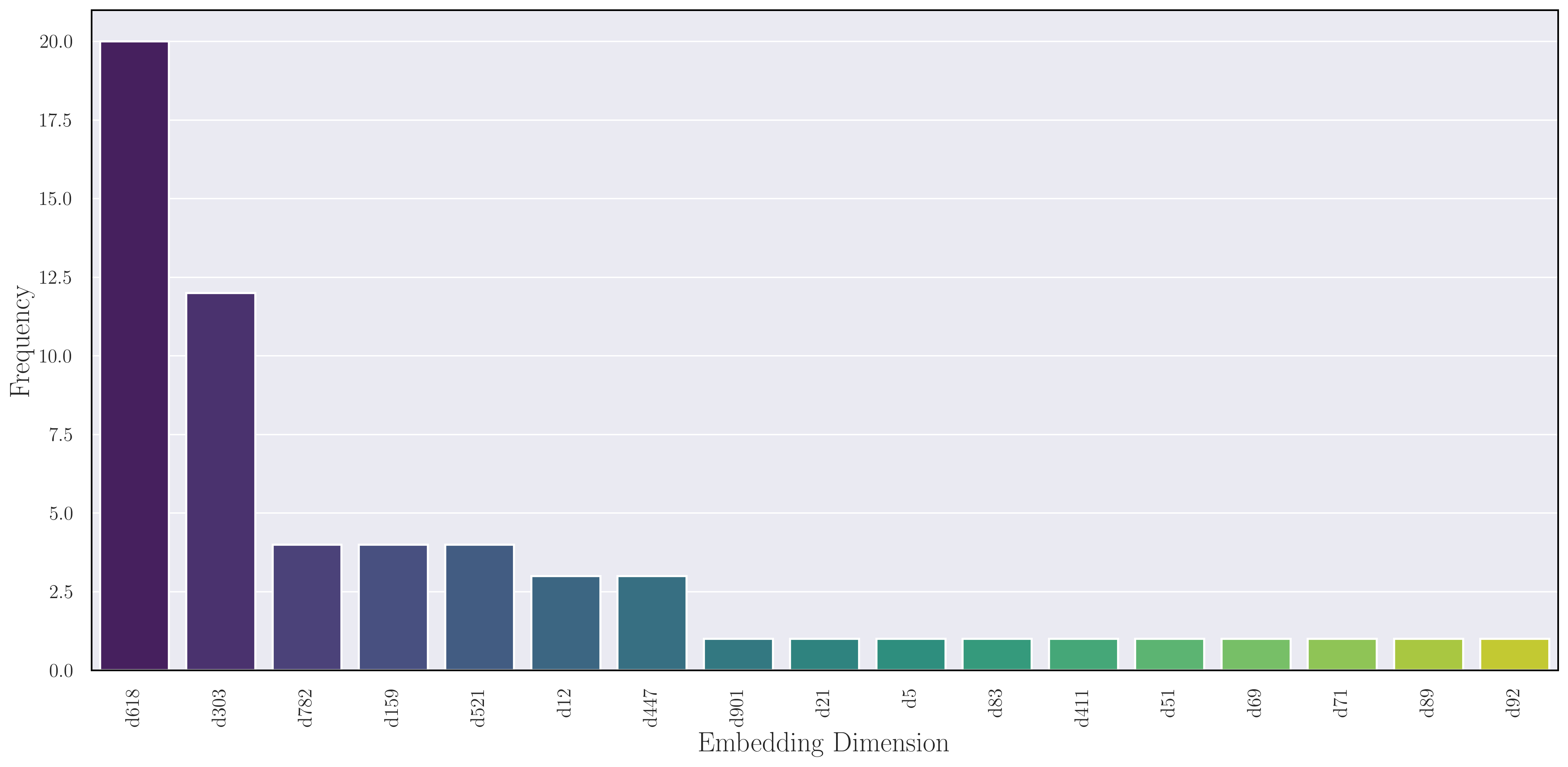}
  \subcaption{MNIST}
  \label{subfig:hist_mnist}
\end{minipage}

\vspace{0.8em} 

\begin{minipage}[b]{0.38\textwidth}
  \includegraphics[width=\linewidth]{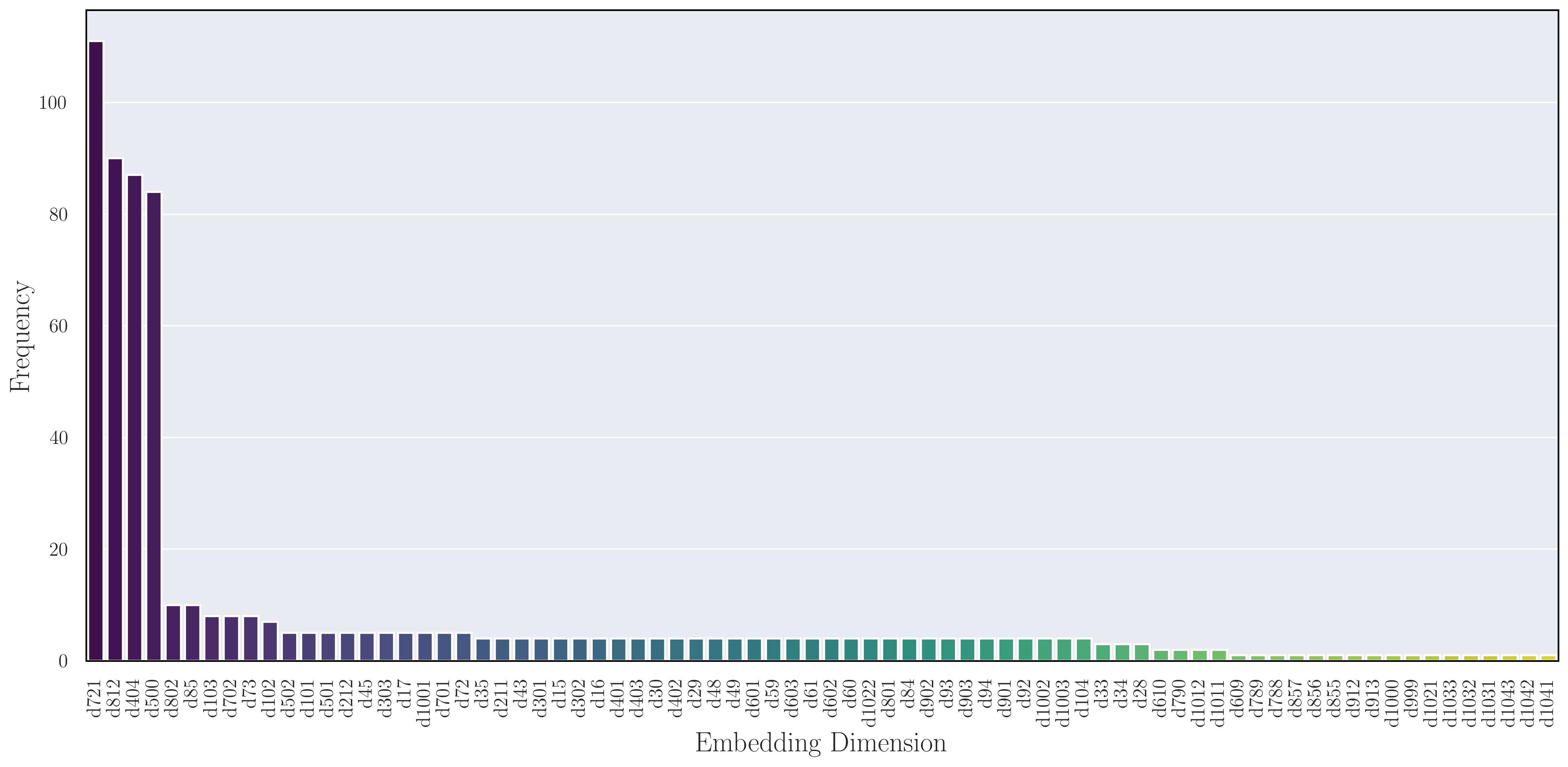}
  \subcaption{CIFAR10}
  \label{subfig:hist_cifar10}
\end{minipage}\hspace{0.02\textwidth}
\begin{minipage}[b]{0.38\textwidth}
  \includegraphics[width=\linewidth]{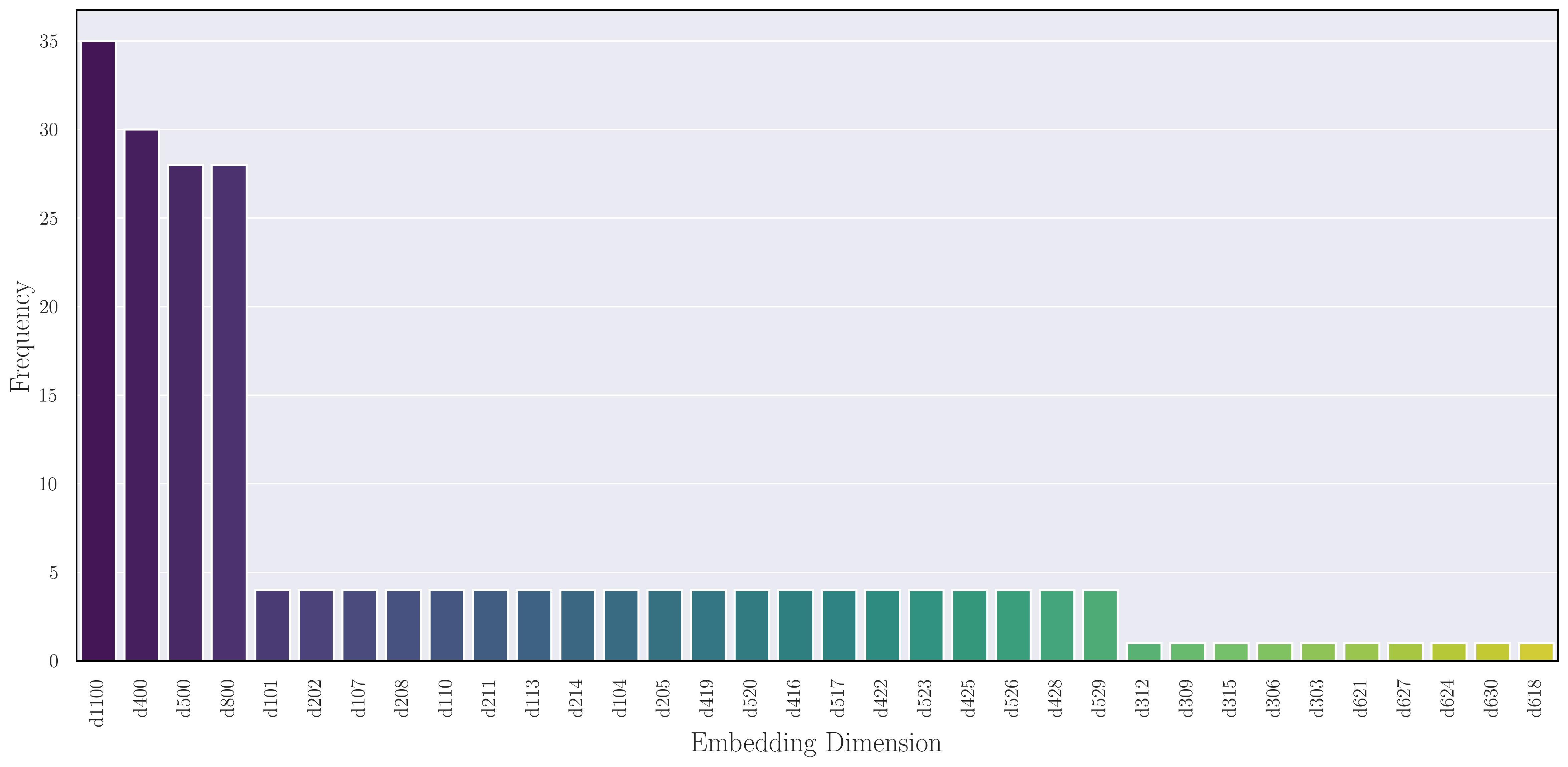}
  \subcaption{MSC17}
  \label{subfig:hist_msc17}
\end{minipage}

\caption{Embedding-dimension usage across canonical logits. Bars show the frequency with which each embedding dimension appears in the canonical GP expressions for each dataset (sorted in descending order).}
\label{fig_histogram}
\end{figure*}

Figure~\ref{fig_histogram} ranks dimensions by frequency of occurrence across the canonical logits. All datasets display a heavy–tailed reuse profile: a handful of dimensions are repeatedly selected (left side of each panel), followed by a long tail of one–off coordinates. The tail is shortest for MNIST, where most logits rely on disjoint, class–specific dimensions; it is longest for CIFAR10 and 20NG, whose top coordinates recur across many logits. This pattern reflects view–level diversity (from SPFP) coupled with cross–class reuse of especially informative coordinates.

\begin{figure*}[ht]
\centering

\captionsetup[subfigure]{justification=centering,singlelinecheck=false,font=small}

\begin{subfigure}[t]{0.32\textwidth}
  \centering
  \includegraphics[height=0.22\textheight,keepaspectratio]{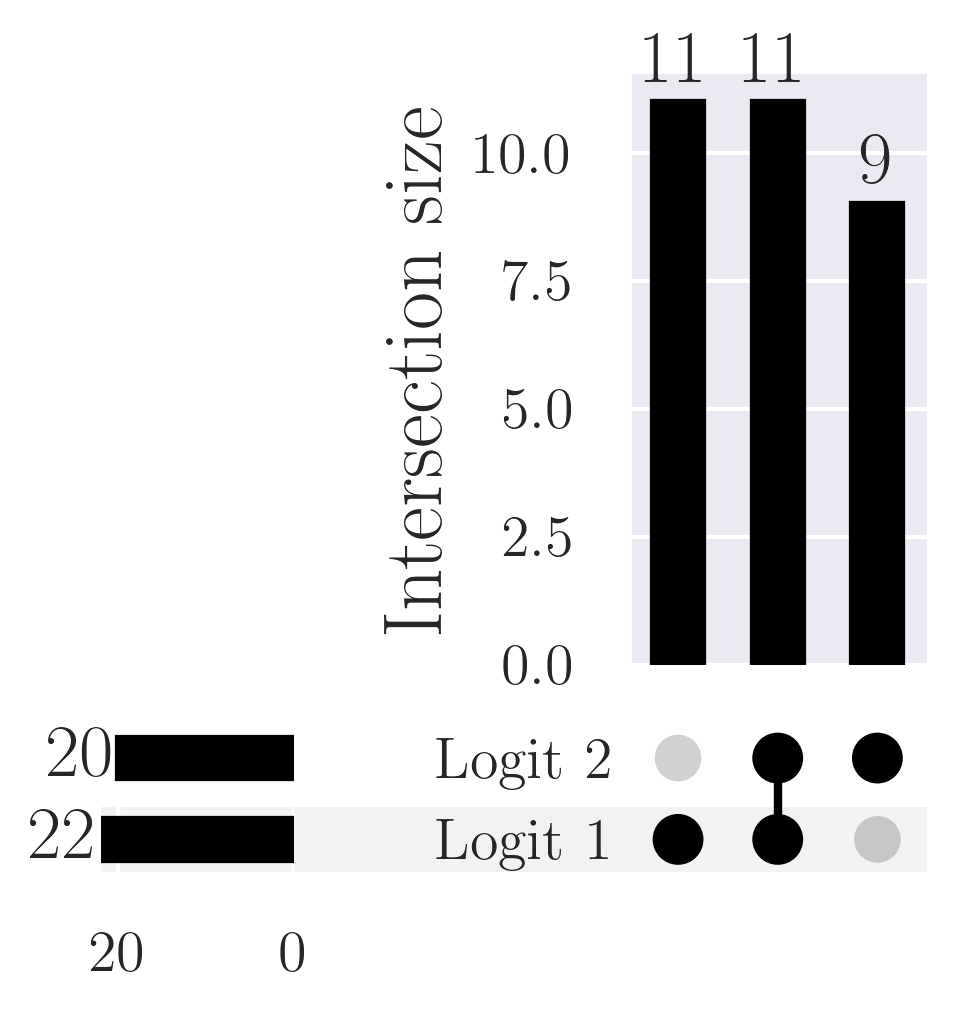}
  \caption{SST2G}\label{fig_upset_sst2g}
\end{subfigure}\hfill
\begin{subfigure}[t]{0.32\textwidth}
  \centering
  \includegraphics[height=0.22\textheight,keepaspectratio]{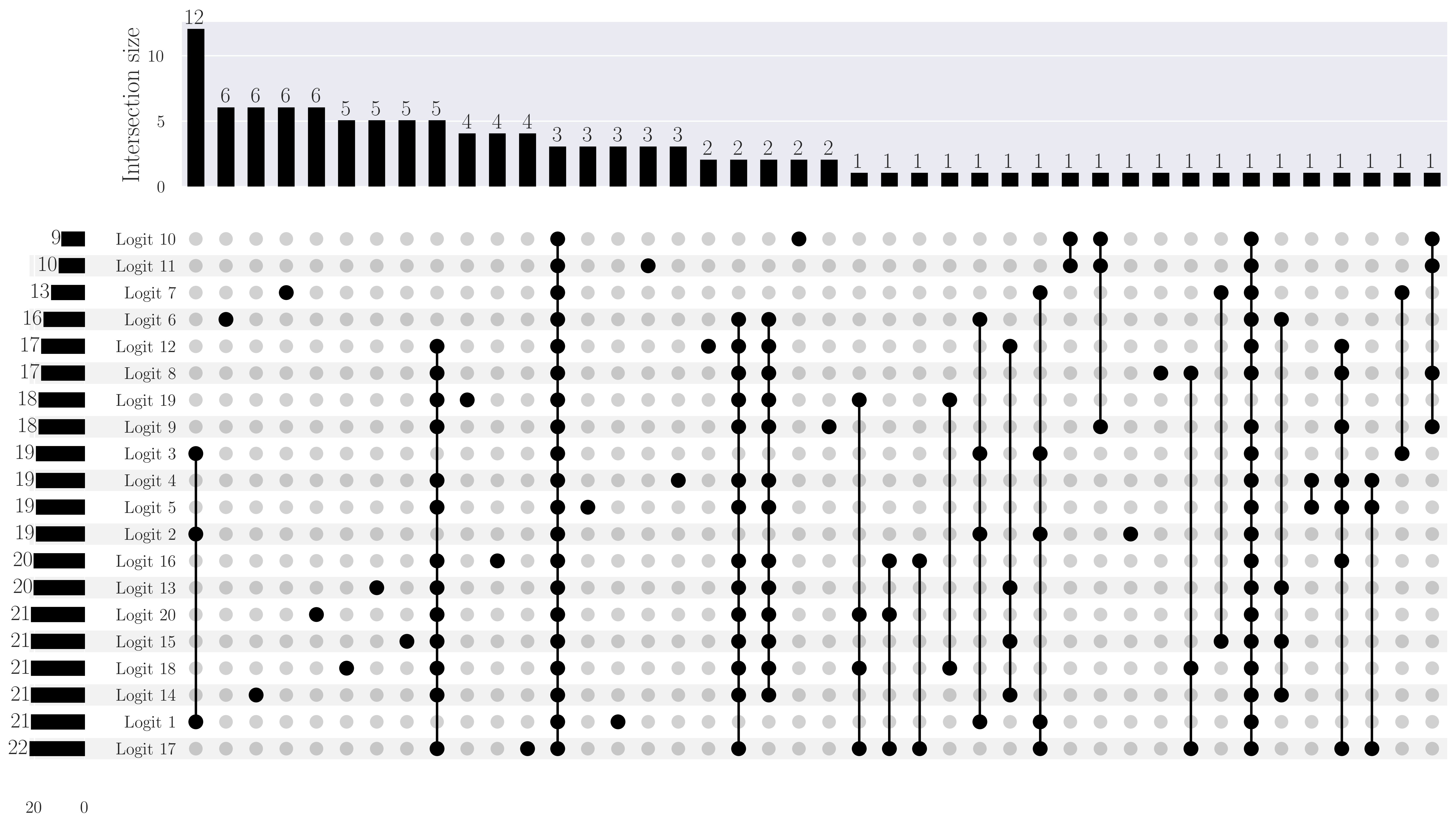}
  \caption{20NG}\label{fig_upset_20ng}
\end{subfigure}\hfill
\begin{subfigure}[t]{0.32\textwidth}
  \centering
  \includegraphics[height=0.22\textheight,keepaspectratio]{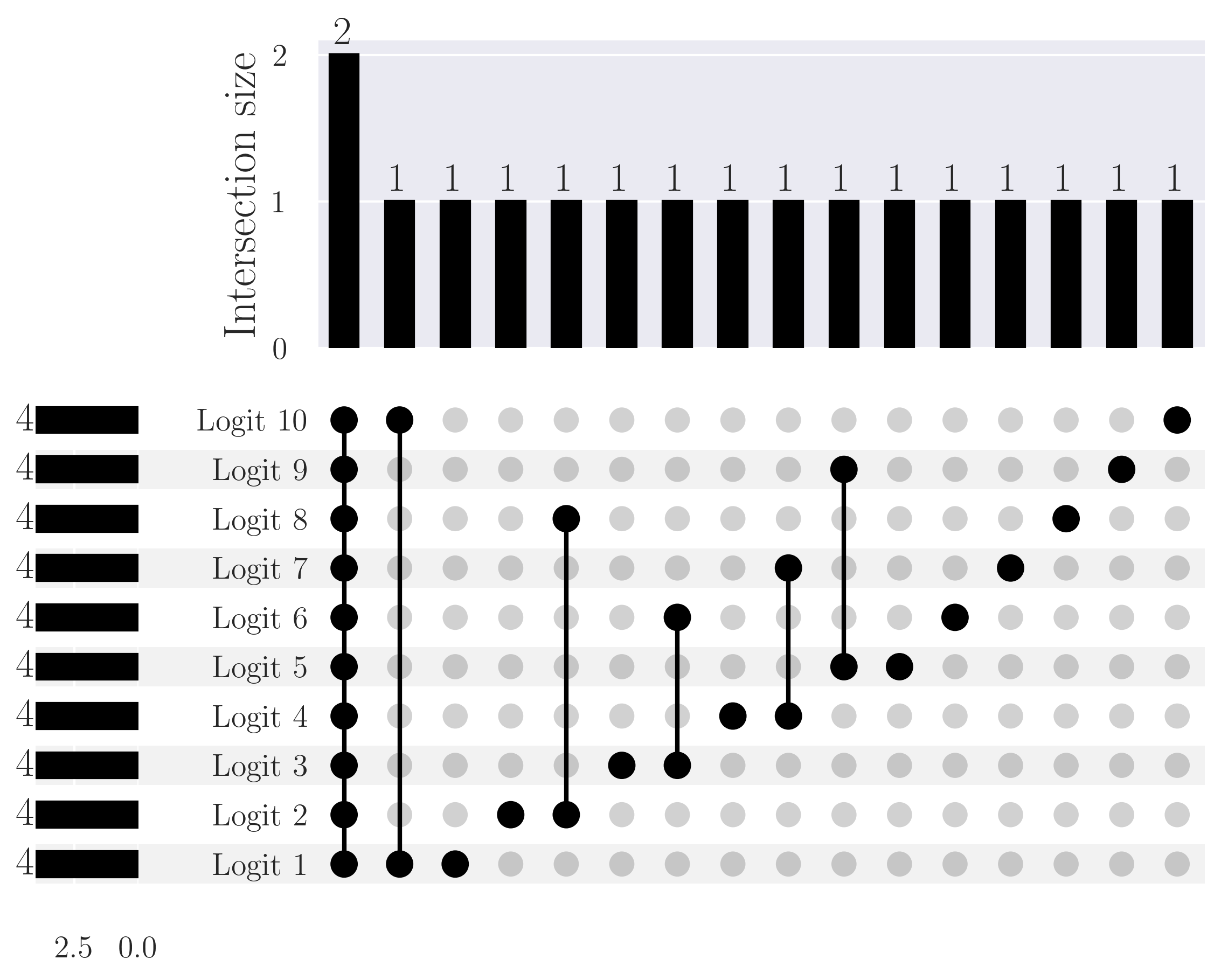}
  \caption{MNIST}\label{fig_upset_mnist}
\end{subfigure}

\vspace{0.6em}

\begin{subfigure}[t]{0.46\textwidth}
  \centering
  \includegraphics[height=0.24\textheight,keepaspectratio]{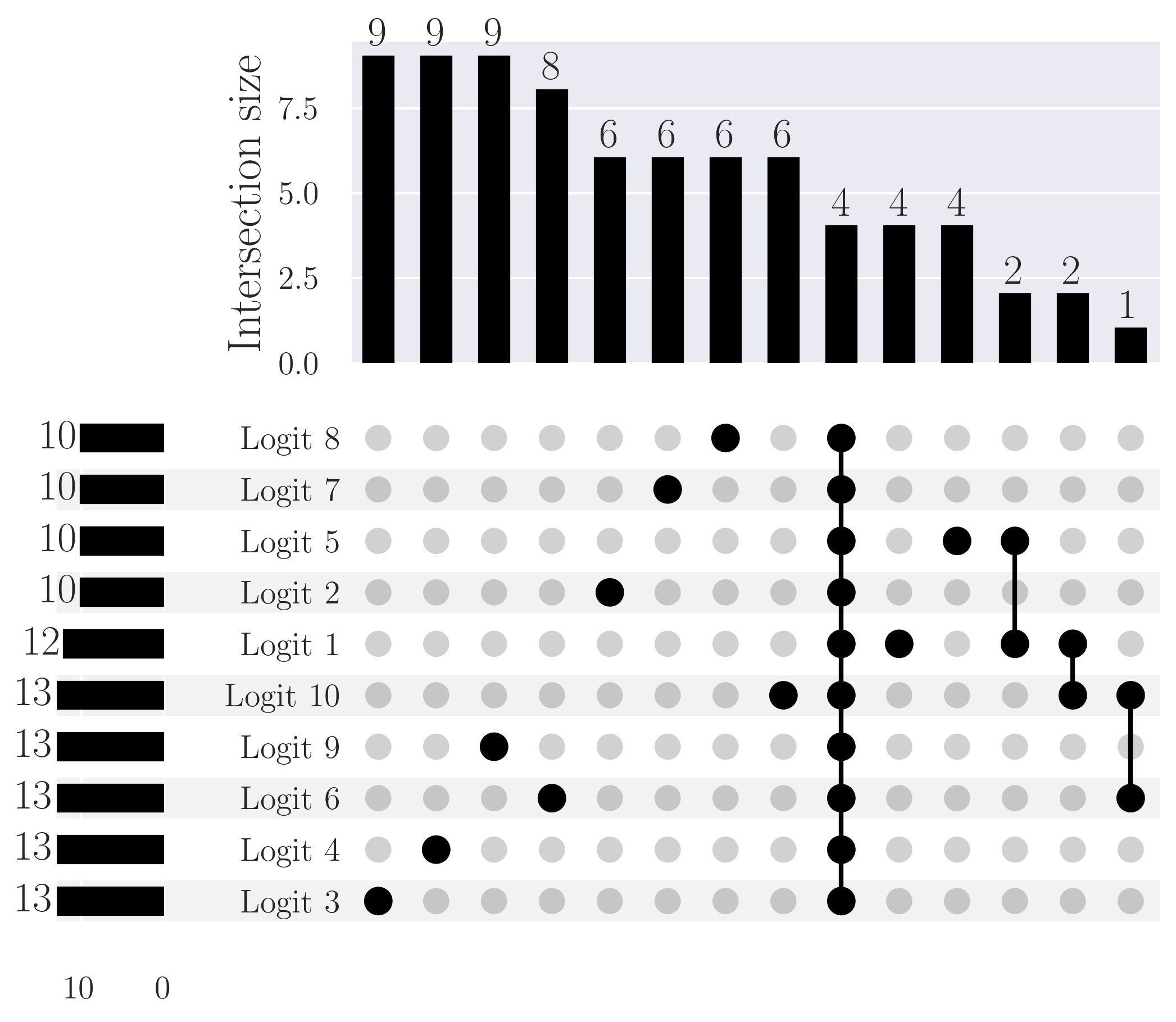}
  \caption{CIFAR10}\label{fig_upset_cifar10}
\end{subfigure}\hfill
\begin{subfigure}[t]{0.46\textwidth}
  \centering
  \includegraphics[height=0.24\textheight,keepaspectratio]{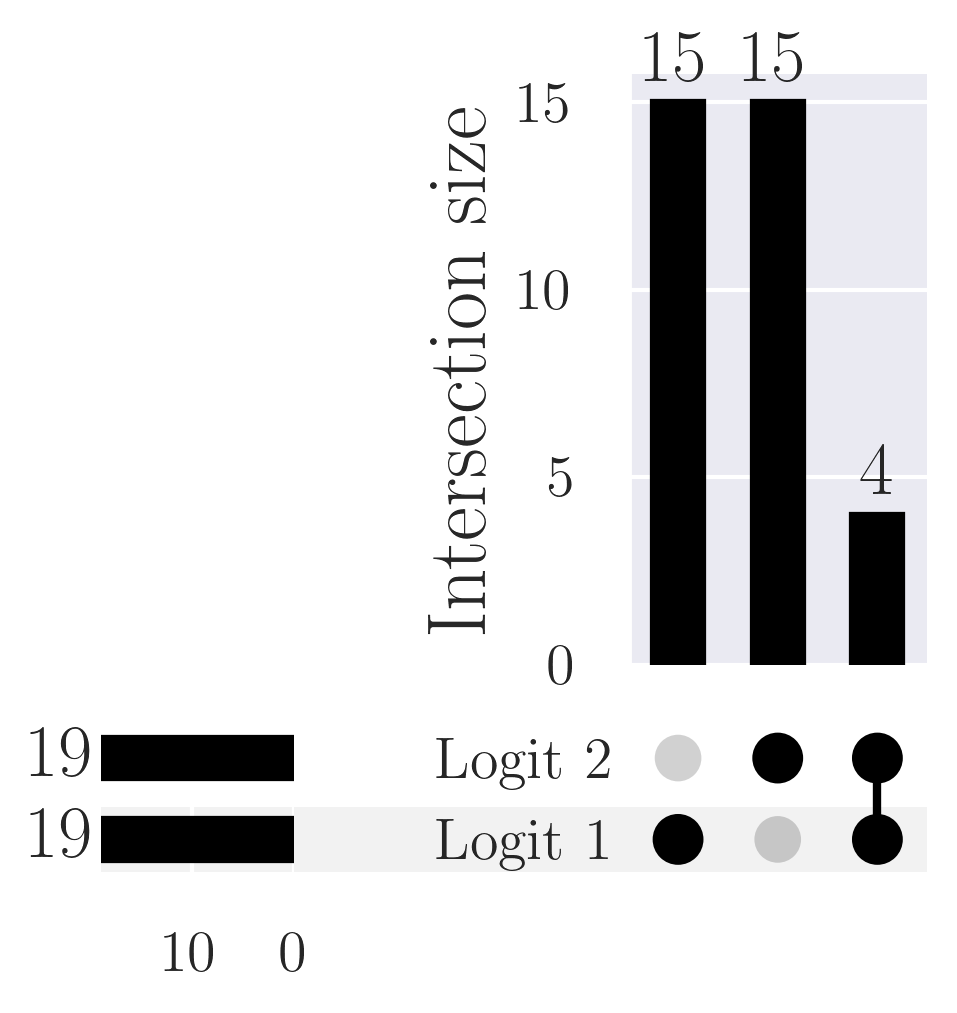}
  \caption{MSC17}\label{fig_upset_msc17}
\end{subfigure}

\caption{UpSet plots of embedding–dimension usage overlap for the canonical GP logits in each dataset. Bars (top) show intersection sizes; connected dots (bottom) indicate the participating logits for each intersection; left bars give total dimensions used by each logit.}
\label{fig_upset}
\end{figure*}

Inter–logit sharing is quantified by the UpSet plots in Fig.~\ref{fig_upset}. MNIST shows minimal intersections (dominant bars of size $1$–$2$), corroborating the near–disjoint usage noted above. CIFAR10 exhibits multiple large intersections (sizes $\gtrsim\!8$–$9$), indicating a common pool of highly predictive dimensions reused across many classes. Text datasets diverge: SST2G (two logits) shares roughly half of its dimensions between the two classes, while 20NG presents numerous medium–sized intersections scattered over many logits, consistent with topic heterogeneity and partially shared lexical signals. MSC17 (two logits) shows a large shared core with a smaller set of unique dimensions per logit, matching the bimodal image–text nature of the task.

Taken together, Table~\ref{tab_symbol_stats} with Figs.~\ref{fig_histogram} and~\ref{fig_upset} demonstrate that the canonical surrogates are both sparse and structured: they use few dimensions overall, reuse a small set of highly informative coordinates where beneficial (CIFAR10, 20NG, MSC17), and otherwise keep class–specific channels separated (MNIST). This organization emerges from SPFP’s view partitioning (Table~\ref{tab_spfp}) and MEGP’s cooperative search, and it underpins the interpretability analyses that follow.

\begin{table*}[t]
\centering
\footnotesize
\caption{Symbol usage, sparsity, and complexity of canonical GP logits.}
\label{tab_symbol_stats}
\begin{tabular}{l|rrrrrrrrrrrr}
\hline
\textbf{Dataset} & \textbf{\#Logits} & \textbf{Unique D} & \textbf{M D/L} & \textbf{M Nodes} & \textbf{M Depth} & \textbf{M Consts/L} & \textbf{M Complexity} & \textbf{ops ($+$)} & \textbf{ops ($-$)} & \textbf{(ops $\times$)} & \textbf{ops ($\div$)} \\
\hline
SST2G   & 2  & 31  & 21.00 & 70  & 27 & 1.00 & 1007 & 15  & 15  & 13  & 13  \\
20NG    & 20 & 113 & 19.00 & 109 & 24 & 2.00 & 1298 & 236 & 231 & 202 & 200 \\
MNIST   & 10 & 17  & 4.00  & 15  & 6  & 1.00 & 62   & 21  & 19  & 10  & 10  \\
CIFAR10 & 10 & 76  & 12.50 & 164 & 27 & 3.00 & 1954 & 182 & 170 & 151 & 150 \\
MSC17   & 2  & 34  & 19.00 & 270 & 33 & 5.00 & 3903 & 61  & 59  & 49  & 50  \\
\hline
\end{tabular}

\vspace{2pt}
\begin{minipage}{0.96\textwidth}\footnotesize
\textbf{Notes.} Abbreviations: \textbf{D} = embedding \emph{dimension}; \textbf{M} = \emph{median} across logits; \textbf{D/L} = \emph{dimensions per logit} (median). 
“\textbf{Unique D}” is the number of distinct embedding dimensions referenced by any canonical logit for that dataset. 
“\textbf{Total ops} ($\pm,\times,\div$)” are \emph{totals across all logits}. 
\textbf{M Complexity} follows our GP measure used throughout (sum of function/terminal nodes with subtree accounting).
\end{minipage}
\end{table*}


\subsection{Global importance and marginal effects of top embedding dimensions}
\label{subsec:results-importance}

\begin{figure*}[t]
  \centering

  \begin{subfigure}[t]{0.32\textwidth}
    \includegraphics[width=\linewidth]{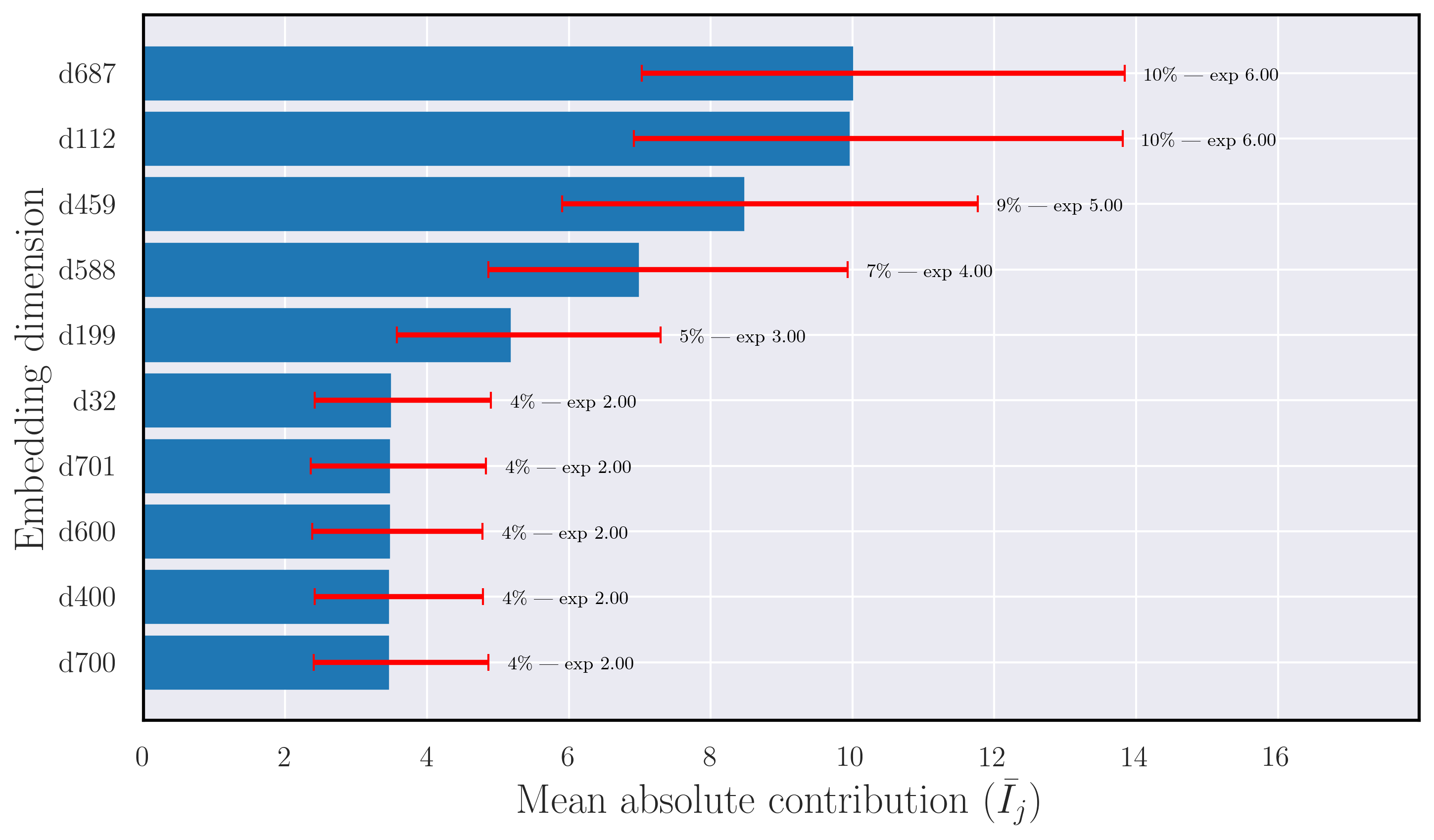}
    \caption{SST2G}\label{fig_importance_sst2g}
  \end{subfigure}\hfill
  \begin{subfigure}[t]{0.32\textwidth}
    \includegraphics[width=\linewidth]{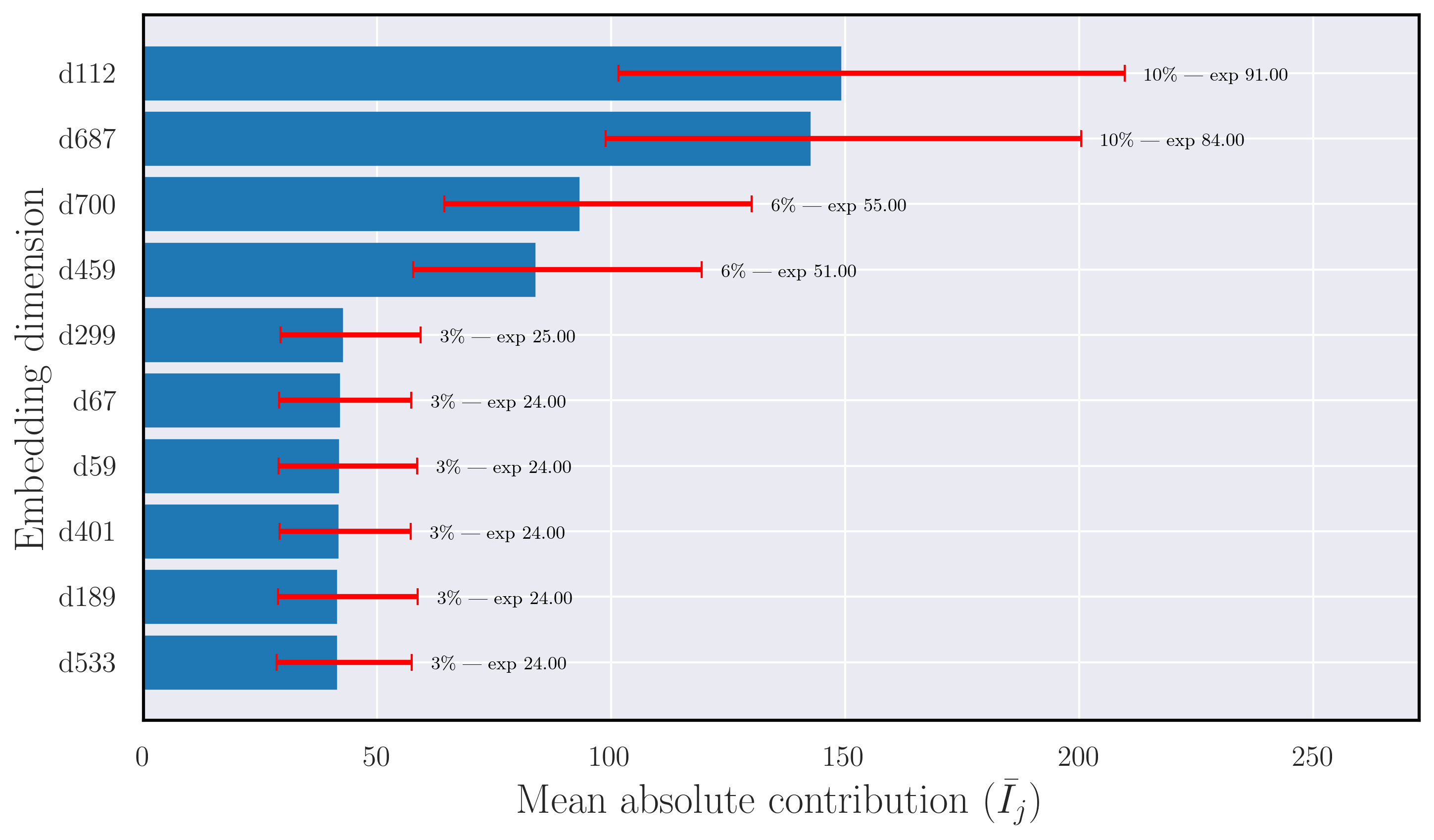}
    \caption{20NG}\label{fig_importance_20ng}
  \end{subfigure}\hfill
  \begin{subfigure}[t]{0.32\textwidth}
    \includegraphics[width=\linewidth]{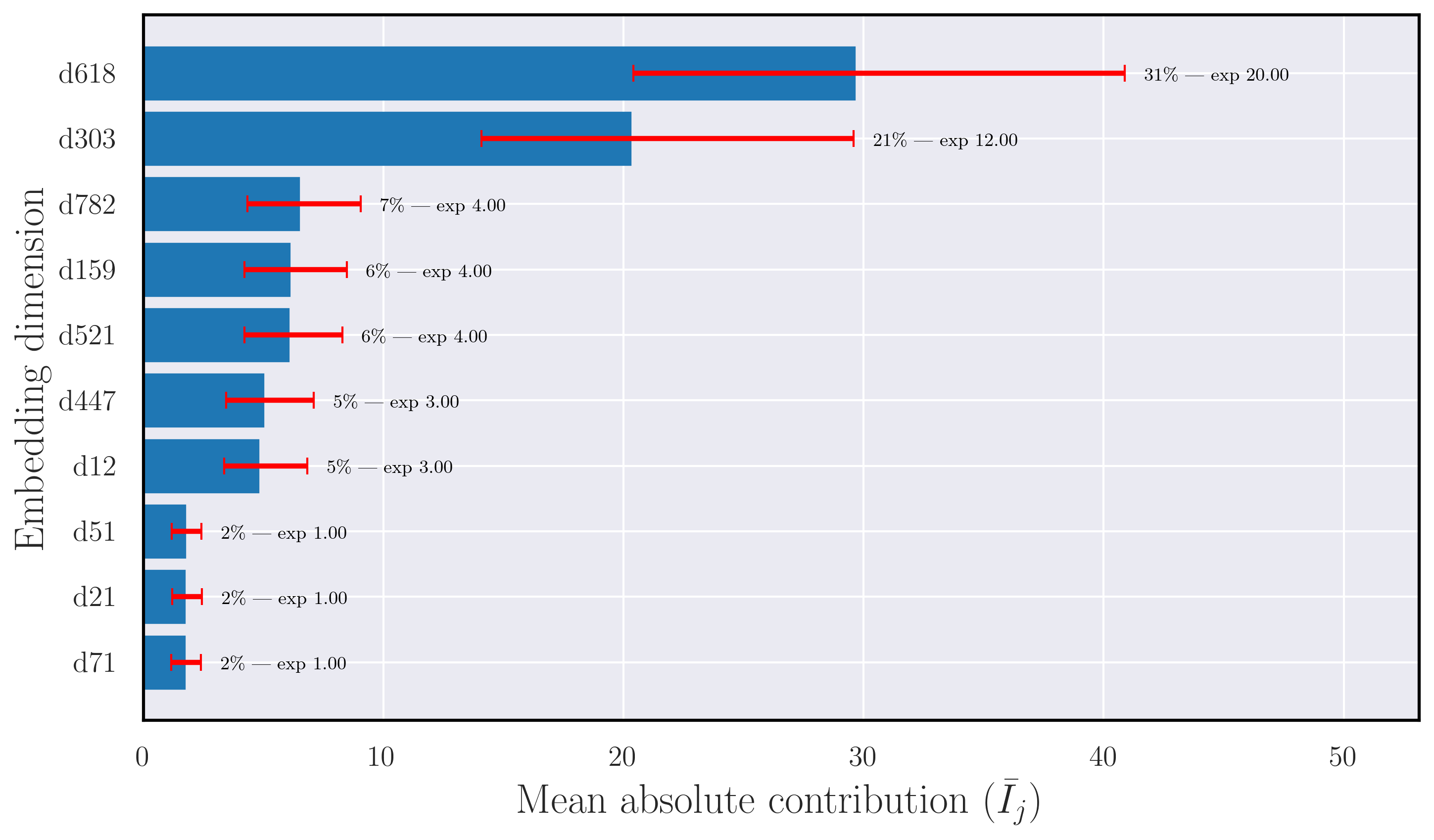}
    \caption{MNIST}\label{fig_importance_mnist}
  \end{subfigure}

  \vspace{0.35em}

  \begin{subfigure}[t]{0.48\textwidth}
    \centering
    \includegraphics[width=\linewidth]{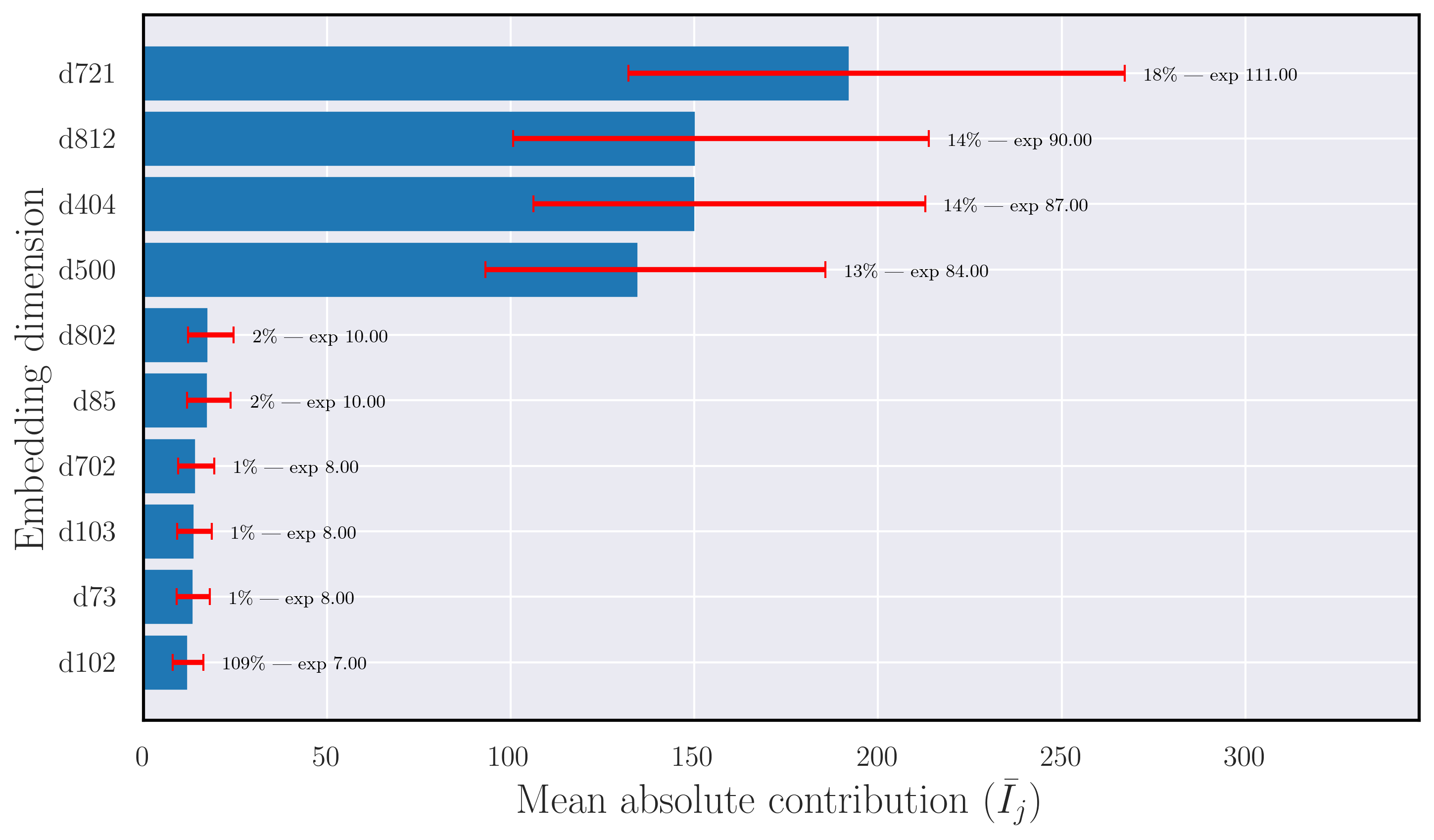}
    \caption{CIFAR10}\label{fig_importance_cifar10}
  \end{subfigure}\hfill
  \begin{subfigure}[t]{0.48\textwidth}
    \centering
    \includegraphics[width=\linewidth]{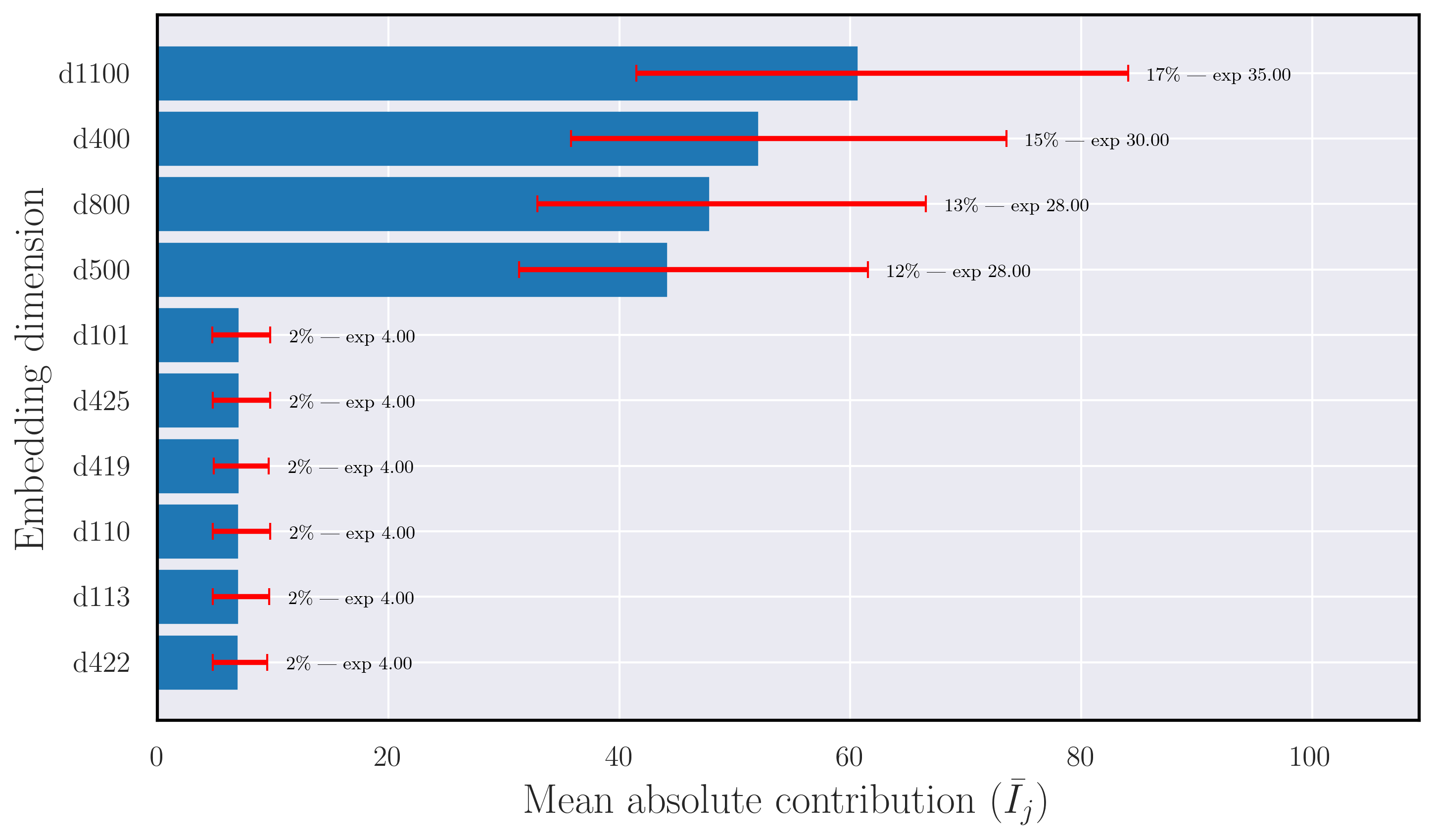}
    \caption{MSC17}\label{fig_importance_msc17}
  \end{subfigure}

  \caption{Top embedding dimensions by mean absolute contribution $\bar I_j$ for each dataset. Blue bars show $\bar I_j$; red whiskers annotate the share of runs in which the dimension appears (and the expected count).}
  \label{fig_importance}
\end{figure*}

\begin{figure*}[t]
\centering
\captionsetup[subfigure]{justification=centering,labelformat=parens,aboveskip=2pt,belowskip=0pt}
\begin{subfigure}[t]{0.19\textwidth}\centering
\includegraphics[width=\linewidth]{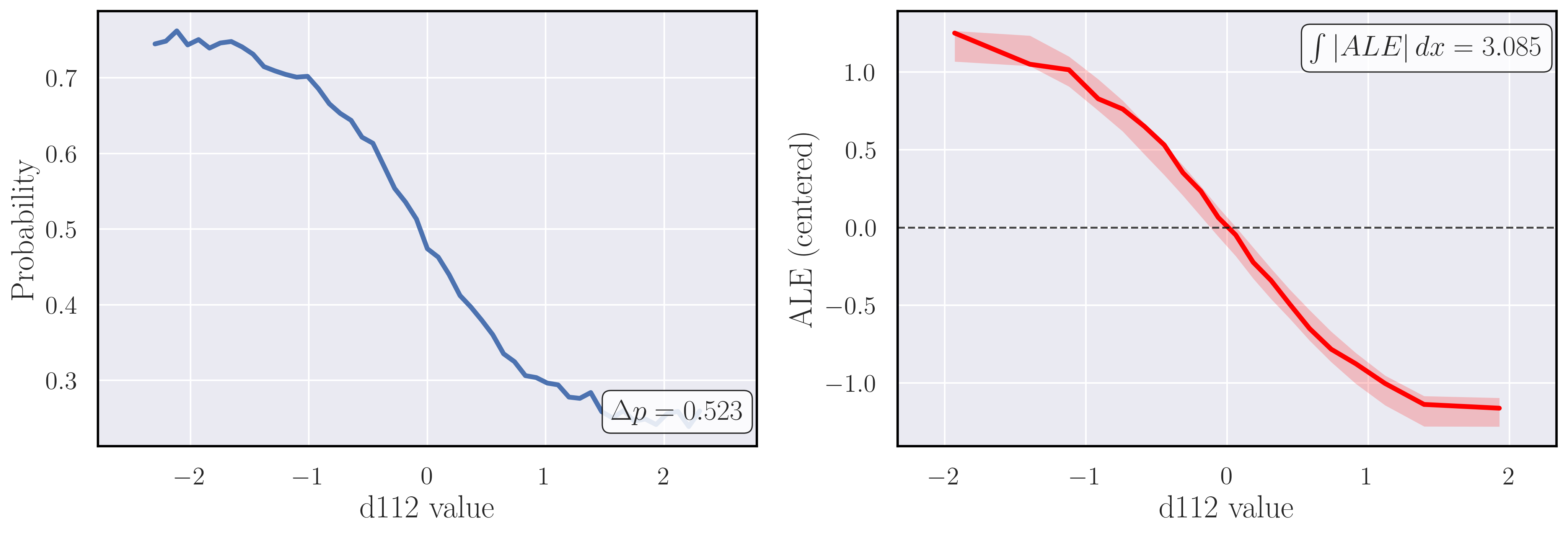}
\caption{SST2G, $d_{112}$}\label{subfig:pdpale_sst2g_d112}
\end{subfigure}\hfill
\begin{subfigure}[t]{0.19\textwidth}\centering
\includegraphics[width=\linewidth]{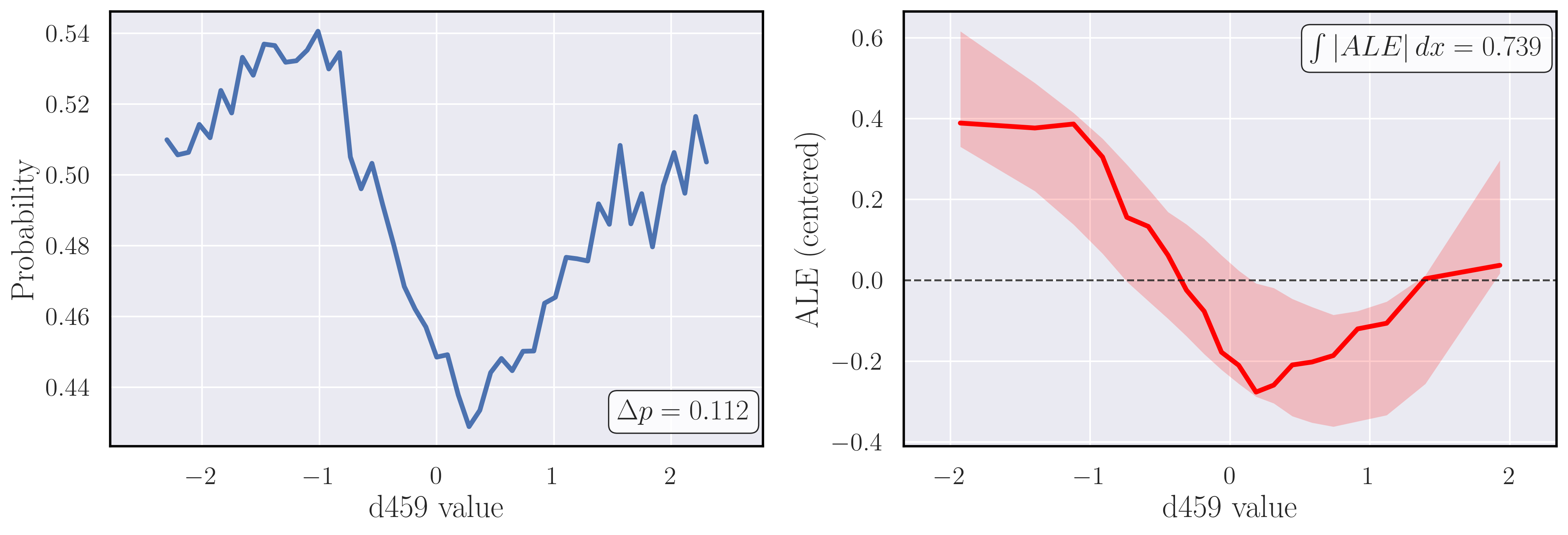}
\caption{SST2G, $d_{459}$}\label{subfig:pdpale_sst2g_d459}
\end{subfigure}\hfill
\begin{subfigure}[t]{0.19\textwidth}\centering
\includegraphics[width=\linewidth]{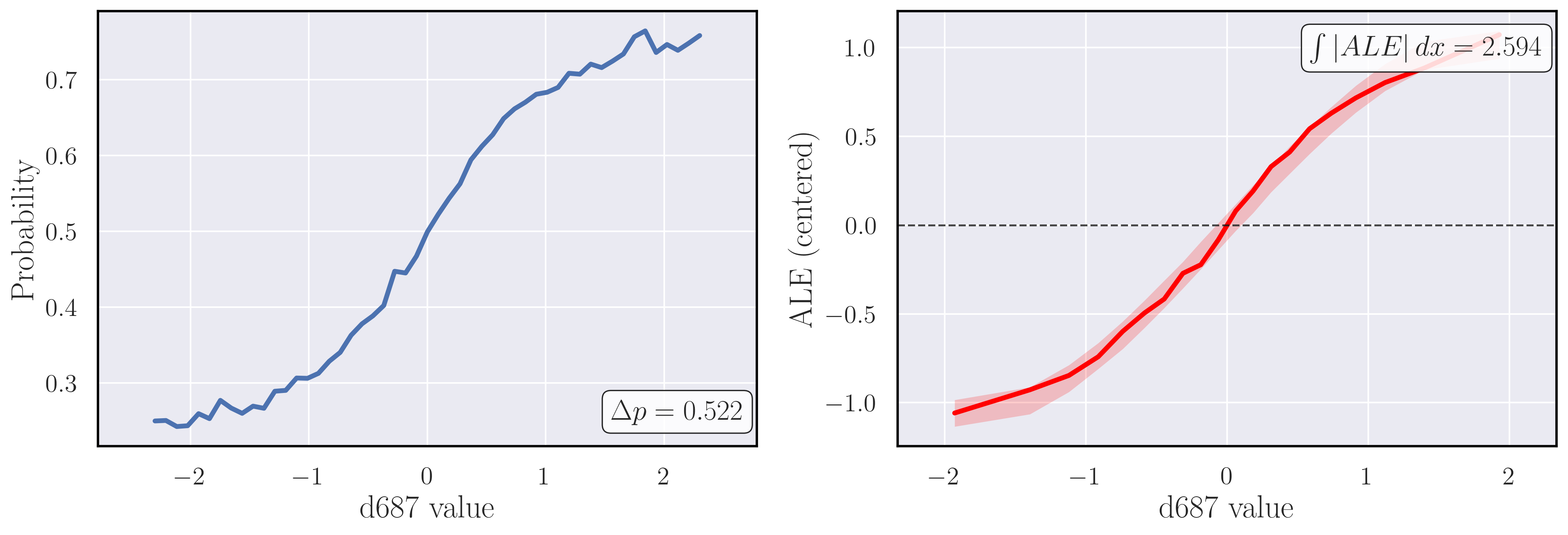}
\caption{SST2G, $d_{687}$}\label{subfig:pdpale_sst2g_d687}
\end{subfigure}\hfill
\begin{subfigure}[t]{0.19\textwidth}\centering
\includegraphics[width=\linewidth]{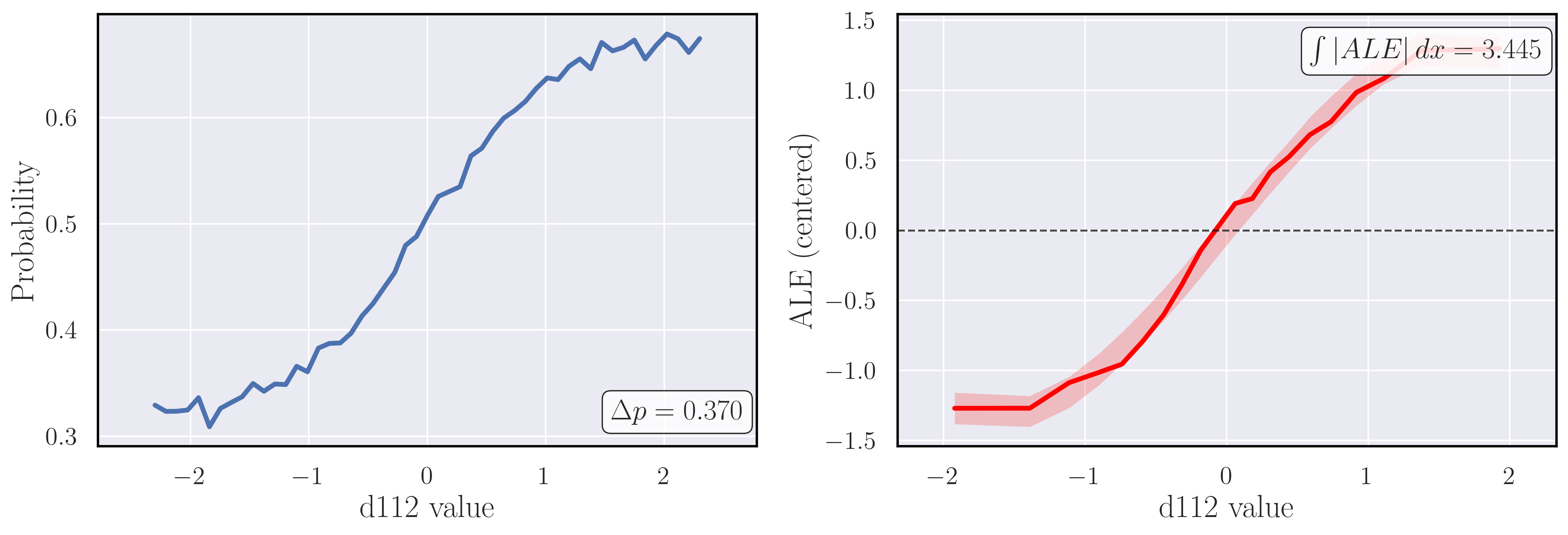}
\caption{20NG, $d_{112}$}\label{subfig:pdpale_20ng_d112}
\end{subfigure}\hfill
\begin{subfigure}[t]{0.19\textwidth}\centering
\includegraphics[width=\linewidth]{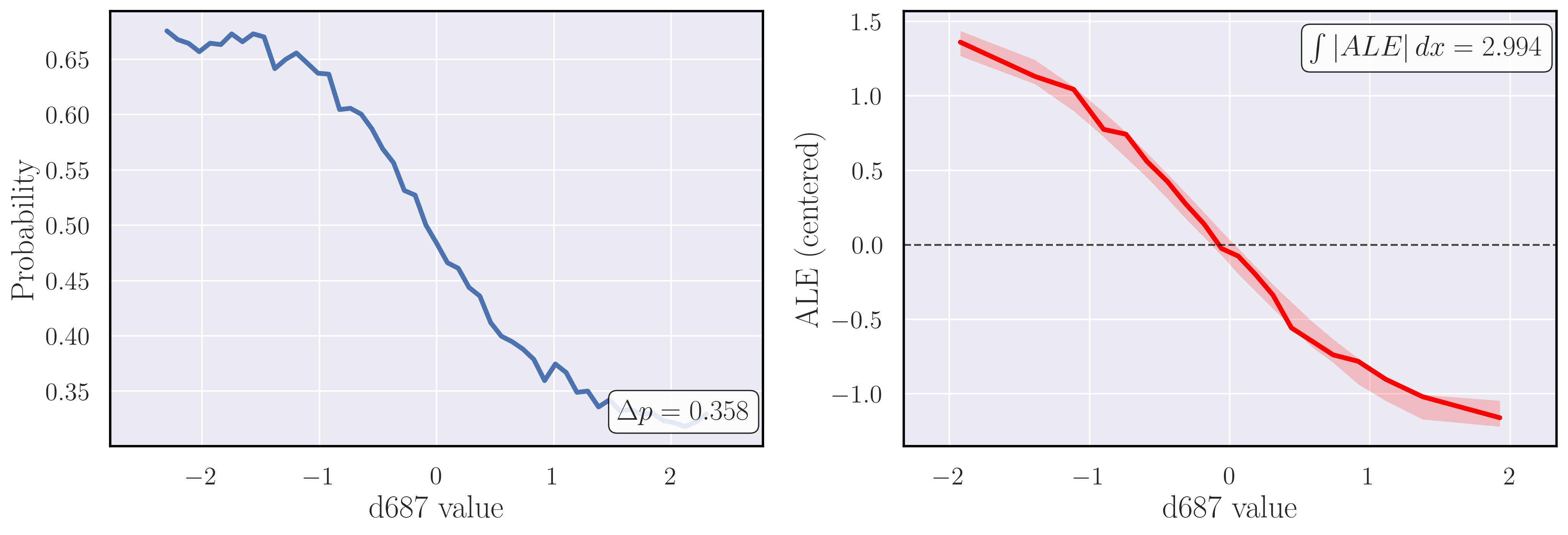}
\caption{20NG, $d_{687}$}\label{subfig:pdpale_20ng_d687}
\end{subfigure}

\vspace{0.35em}

\begin{subfigure}[t]{0.19\textwidth}\centering
\includegraphics[width=\linewidth]{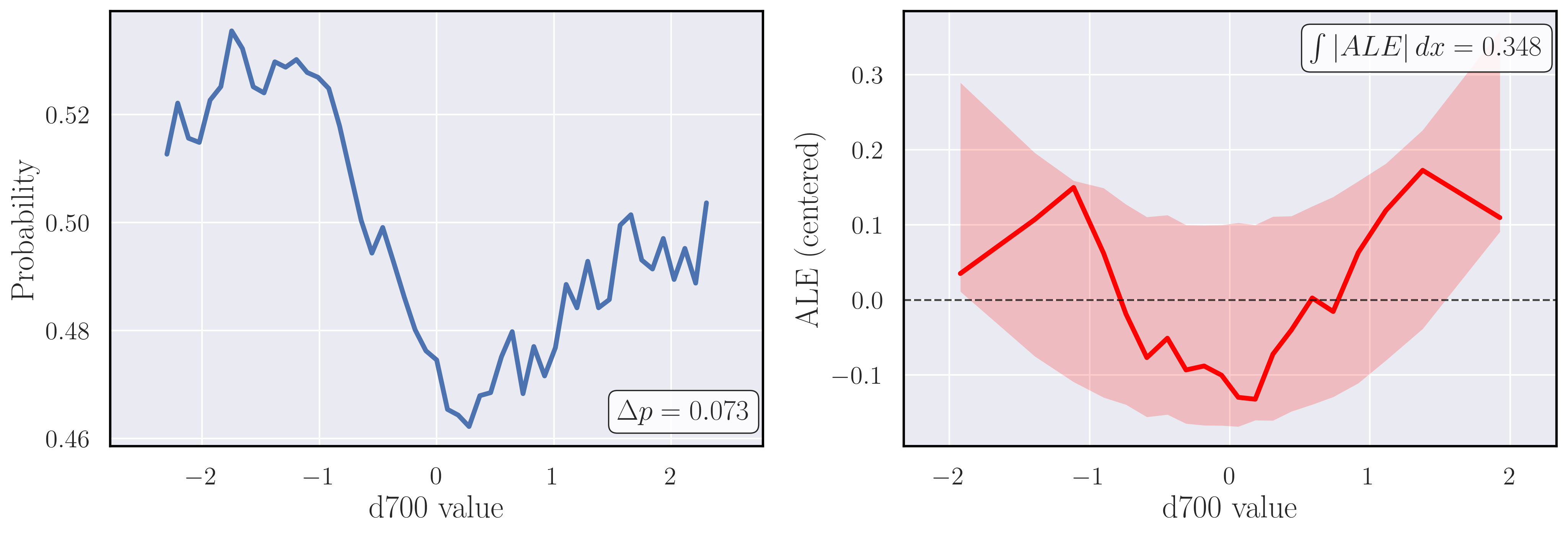}
\caption{20NG, $d_{700}$}\label{subfig:pdpale_20ng_d700}
\end{subfigure}\hfill
\begin{subfigure}[t]{0.19\textwidth}\centering
\includegraphics[width=\linewidth]{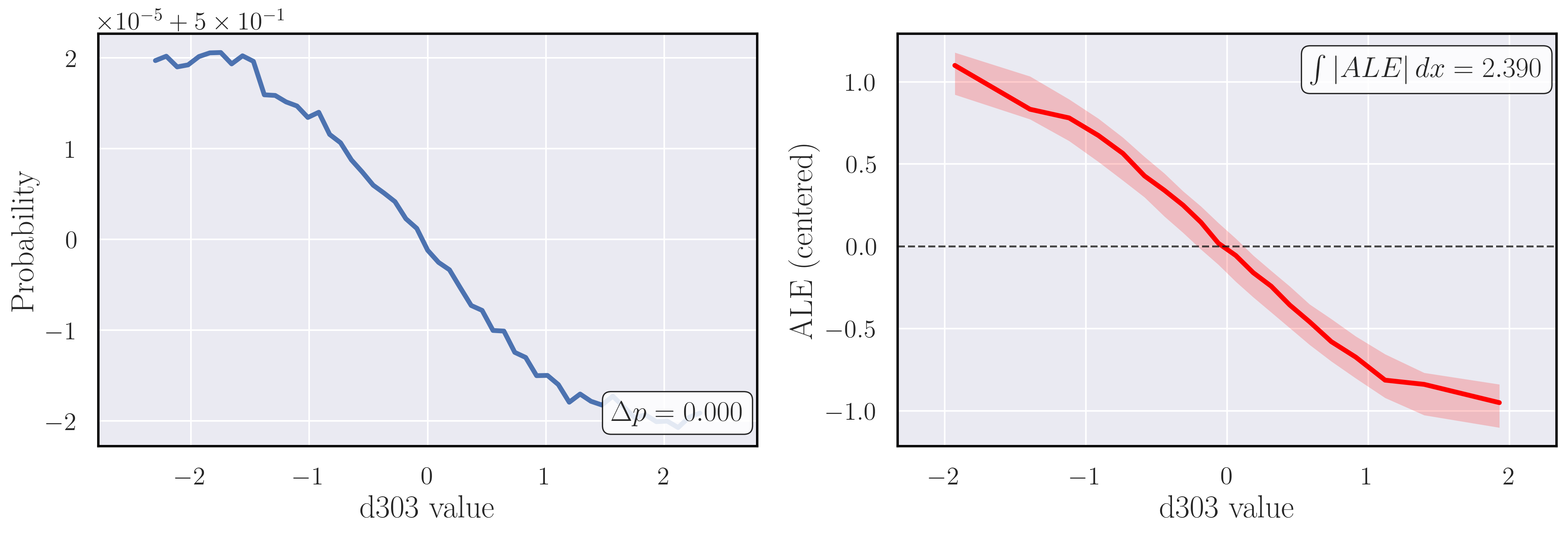}
\caption{MNIST, $d_{303}$}\label{subfig:pdpale_mnist_d303}
\end{subfigure}\hfill
\begin{subfigure}[t]{0.19\textwidth}\centering
\includegraphics[width=\linewidth]{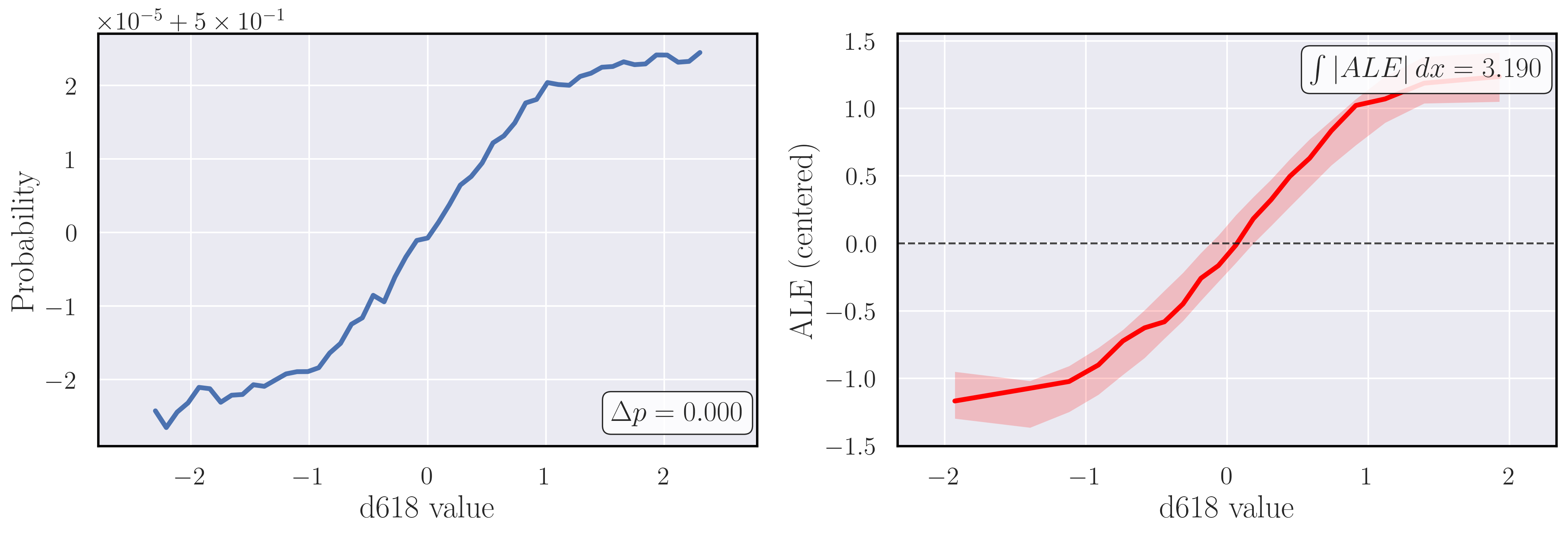}
\caption{MNIST, $d_{618}$}\label{subfig:pdpale_mnist_d618}
\end{subfigure}\hfill
\begin{subfigure}[t]{0.19\textwidth}\centering
\includegraphics[width=\linewidth]{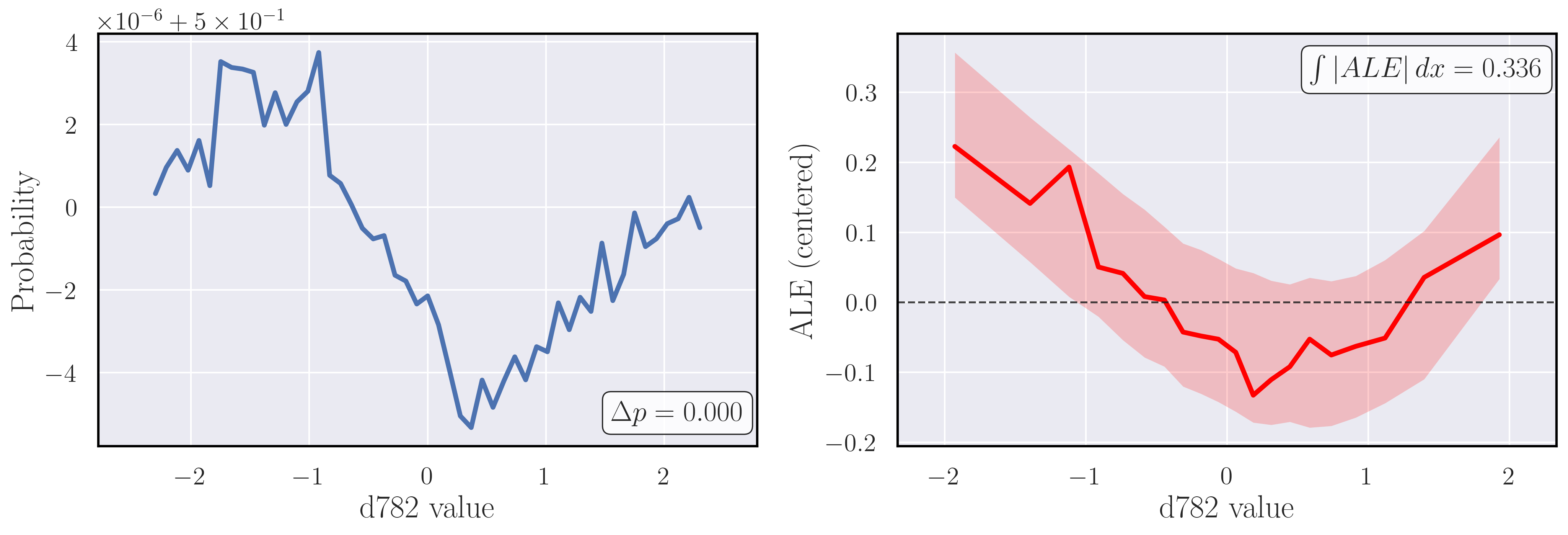}
\caption{MNIST, $d_{782}$}\label{subfig:pdpale_mnist_d782}
\end{subfigure}\hfill
\begin{subfigure}[t]{0.19\textwidth}\centering
\includegraphics[width=\linewidth]{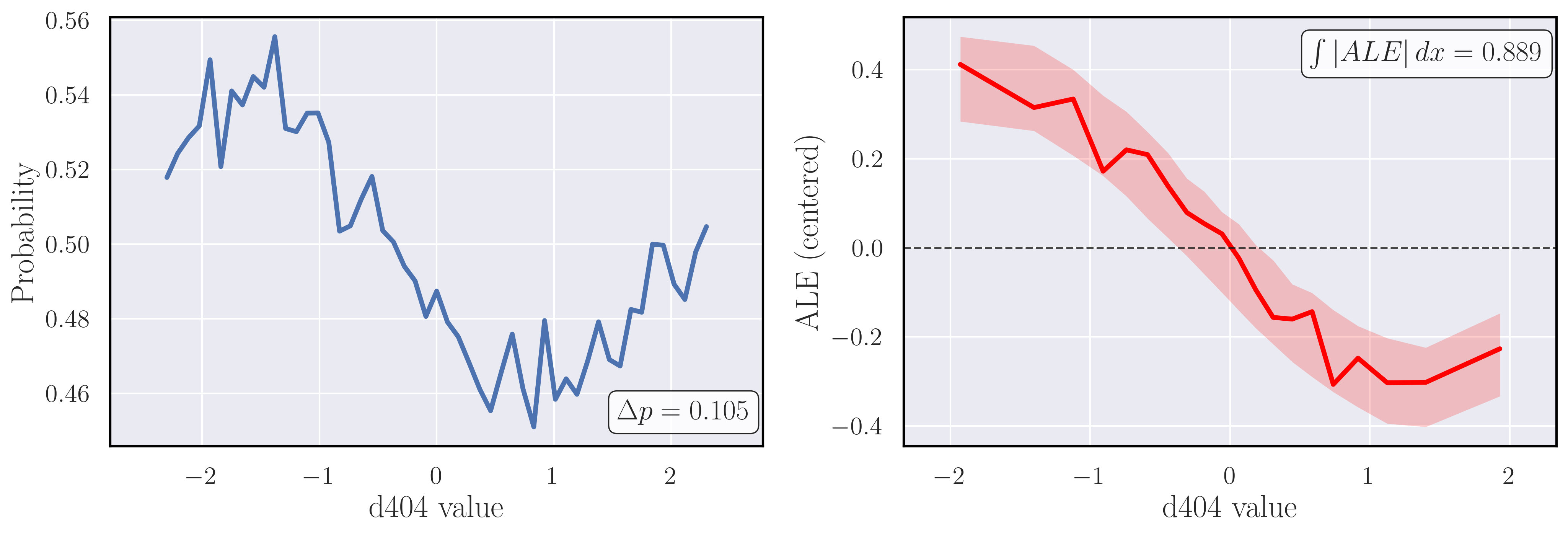}
\caption{CIFAR10, $d_{404}$}\label{subfig:pdpale_cifar10_d404}
\end{subfigure}

\vspace{0.35em}

\begin{subfigure}[t]{0.19\textwidth}\centering
\includegraphics[width=\linewidth]{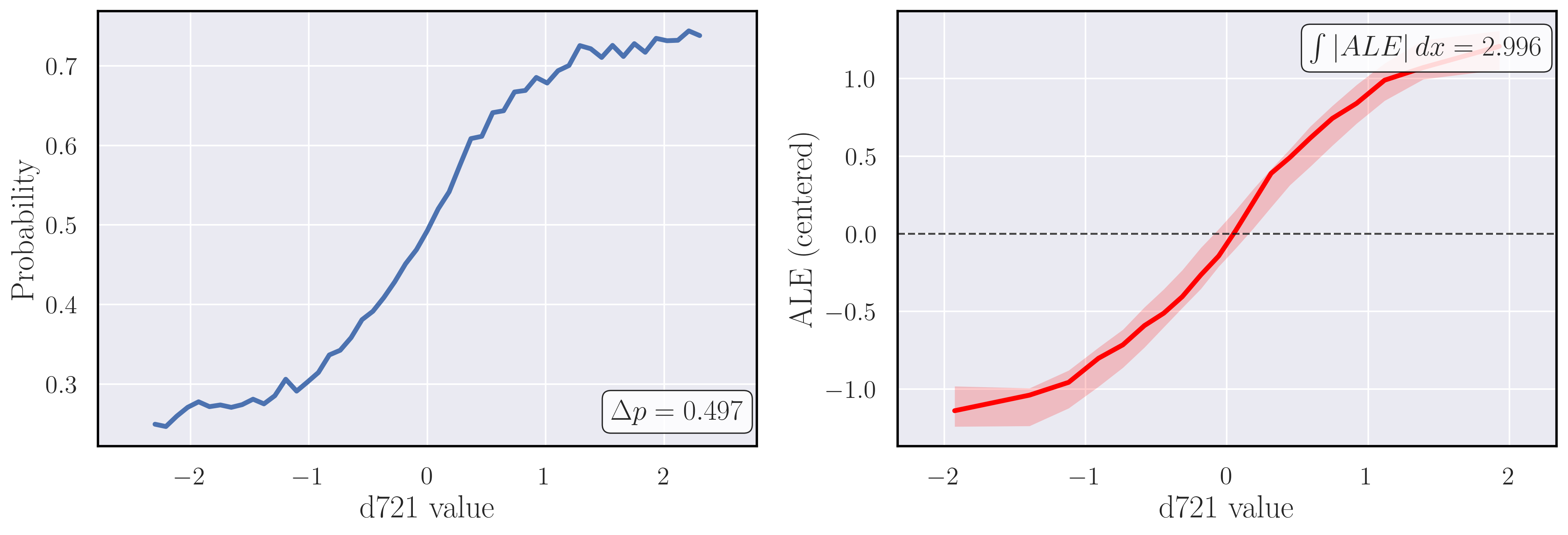}
\caption{CIFAR10, $d_{721}$}\label{subfig:pdpale_cifar10_d721}
\end{subfigure}\hfill
\begin{subfigure}[t]{0.19\textwidth}\centering
\includegraphics[width=\linewidth]{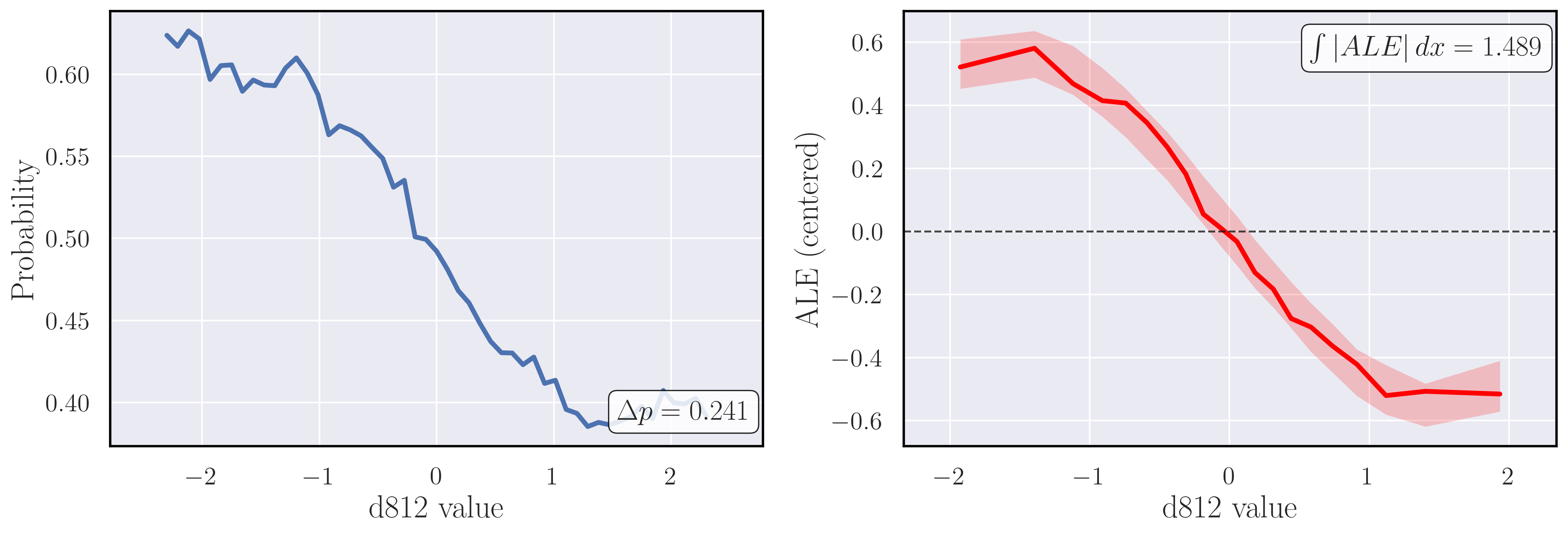}
\caption{CIFAR10, $d_{812}$}\label{subfig:pdpale_cifar10_d812}
\end{subfigure}\hfill
\begin{subfigure}[t]{0.19\textwidth}\centering
\includegraphics[width=\linewidth]{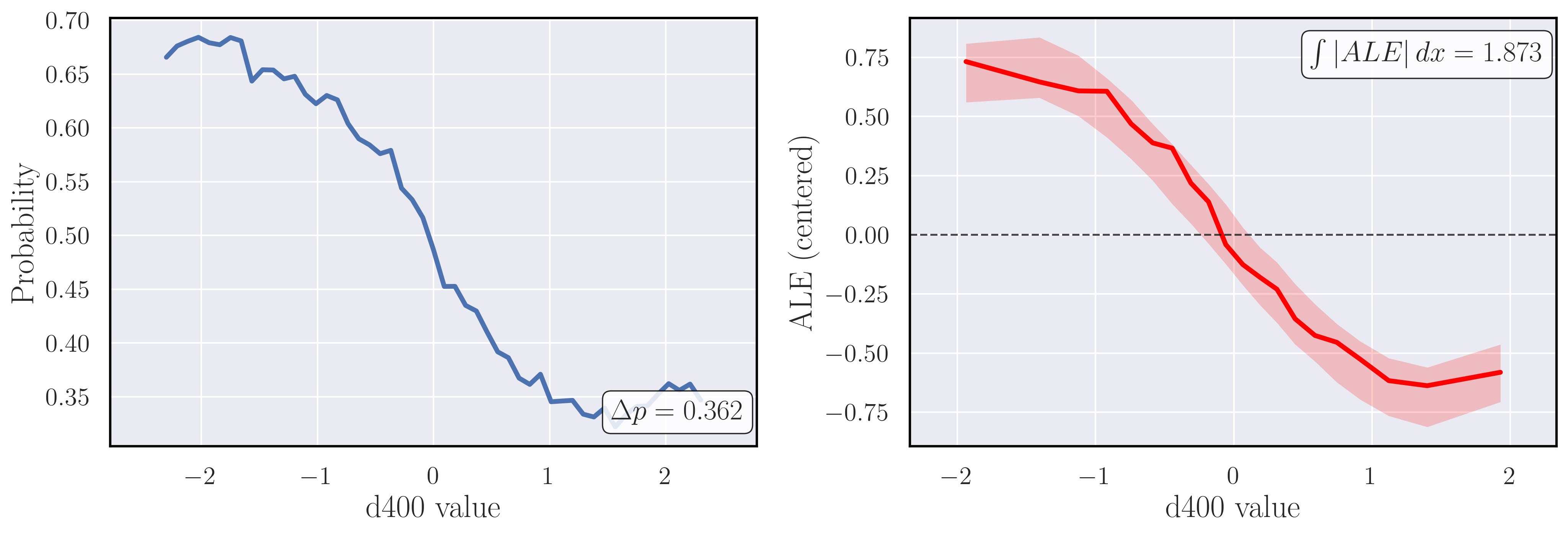}
\caption{MSC17, $d_{400}$}\label{subfig:pdpale_msc17_d400}
\end{subfigure}\hfill
\begin{subfigure}[t]{0.19\textwidth}\centering
\includegraphics[width=\linewidth]{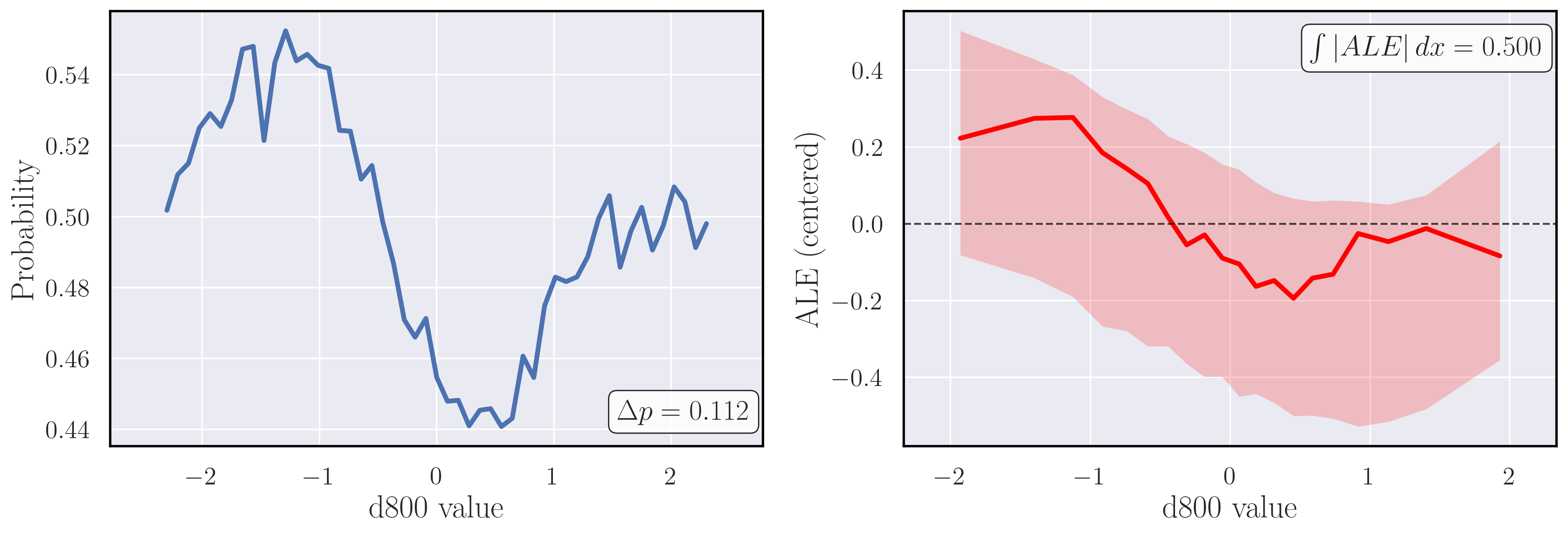}
\caption{MSC17, $d_{800}$}\label{subfig:pdpale_msc17_d800}
\end{subfigure}\hfill
\begin{subfigure}[t]{0.19\textwidth}\centering
\includegraphics[width=\linewidth]{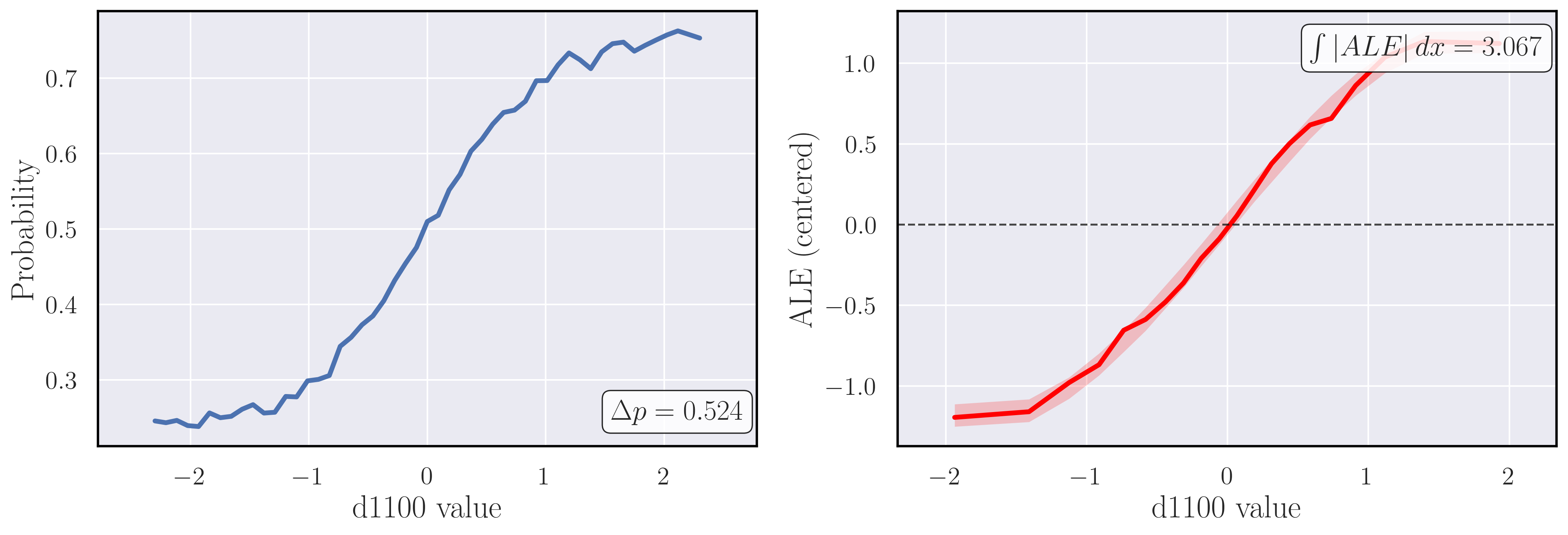}
\caption{MSC17, $d_{1100}$}\label{subfig:pdpale_msc17_d1100}
\end{subfigure}

\caption{PDP/ALE summaries for selected embedding dimensions across datasets. Each panel shows the partial dependence of the predicted class probability (left within image) and the centered ALE with a $95\%$ CI band (right). PDP annotations report $\Delta p$; ALE annotations give $\int |{\rm ALE}|\,dx$. Panels are grouped by dataset.}
\label{fig_pdp_ale}
\end{figure*}

\paragraph{Setup and scoring.}
For each canonical surrogate, we summarize the \emph{global} contribution of embedding coordinate $j$ by the mean absolute term contribution
\begin{equation}
\label{eq:global-importance}
\bar I_j \;=\; \frac{1}{N}\sum_{i=1}^{N}\sum_{k:\; j\in\mathcal S_k}\bigl|w_k\,t_k(\mathbf x_i)\bigr|,
\end{equation}
where $t_k$ is a program term with support $\mathcal S_k$ and weight $w_k$.
Figure~\ref{fig_importance} reports the top coordinates per dataset (blue bars).
For each coordinate, the red annotation indicates (i) the fraction of logits in which the coordinate appears at least once (``\% selected'') and (ii) the largest exponent learned for that coordinate in the symbolic program (``exp'').
To characterize \emph{marginal} behavior along a single coordinate, we compute classwise partial–dependence profiles (PDP) and centered one–dimensional accumulated local effects (ALE) on 20 quantile bins with 200 bootstrap replicates.
Representative panels and their effect magnitudes---PDP range $\Delta p$ and $\int |{\rm ALE}|\,dx$---are shown in Fig.~\ref{fig_pdp_ale}.

\paragraph{Heavy–tailed importance and cross–logit reuse.}
Across all benchmarks, $\bar I_j$ is heavy–tailed: only a small number of coordinates accumulate most of the contribution mass (Fig.~\ref{fig_importance}).
The same coordinates are reused across multiple logits, consistent with the sparsity/overlap analysis of Sec.~\ref{subsec:dim-overlap}.
On 20NG, $d112$ and $d687$ dominate (each selected in $\approx\!10\%$ of logits with large exponents $\approx\!91$ and $\approx\!84$), followed by $d700$, $d459$, and $d299$ (3–6\%, modest exponents).
On SST2G, the leading set is $\{d687,d112,d459,d588,d199\}$ with selection rates $\approx\!4$–10\% and exponents $\approx\!2$–6.
On CIFAR10, the largest bars are $\{d721,d812,d404,d500\}$, selected in $\approx\!13$–18\% of logits with exponents $\approx\!84$–111.
MSC17 concentrates mass on $\{d1100,d400,d800,d500\}$ (12–17\%, exponents $\approx\!28$–35), while MNIST is dominated by $\{d618,d303\}$ (21–31\%, smaller exponents), followed by a long tail of weak coordinates (mostly 2–7\% selection).

\paragraph{Monotone, high–impact directions.}
Table~\ref{tap_top3_monotonicity} summarizes, for each dataset, the three most influential embedding coordinates together with a quantitative assessment of their marginal monotonicity (absolute Spearman $\rho$) computed both on the PDP and on the centered ALE grid. Two consistent patterns emerge. 
First, in \emph{every} dataset the top two coordinates show near–perfect monotone effects in both summaries ($|\rho_{\text{PDP}}|,|\rho_{\text{ALE}}|\ge 0.95$), and they move the one–vs–rest probability in opposite directions: a strongly \emph{increasing} axis (CIFAR10: $d721$; 20NG: $d112$; SST2G: $d687$; MSC17: $d1100$; MNIST: $d618$) paired with a strongly \emph{decreasing} axis (CIFAR10: $d812$; 20NG: $d687$; SST2G: $d112$; MSC17: $d400$; MNIST: $d303$). These “antagonistic pairs’’ carry the largest global contributions (see the \textsc{Importance} column) and are tagged as \emph{dominant} or \emph{clean effect}, indicating that the same direction is reused across logits without requiring additional nonlinear corrections.

Second, the third-ranked coordinate in each dataset systematically exhibits weaker monotonicity and signs of curvature or correlation with other features. The PDP–ALE discrepancy is particularly diagnostic: for 20NG $d700$, $|\rho_{\text{PDP}}|=0.58$ but $|\rho_{\text{ALE}}|=0.16$, suggesting that the naive average (PDP) is partially confounded by feature dependence whereas the local adjustment (ALE) reveals little net monotone trend; for CIFAR10 $d404$, the effect is almost monotone in ALE ($0.96$) but only moderately so in PDP ($0.71$), again consistent with correlation–induced bias in PDP. Analogous behavior appears for SST2G $d459$, MNIST $d782$, and MSC17 $d800$, all marked \emph{nonlinear}. 

This indicates that the canonical MEGP surrogates route most predictive signal through a small pair of nearly linear axes with opposite signs, while a third coordinate provides dataset-specific nonlinear or interaction corrections. This decomposition aligns with the heavy–tailed global importance pattern and with the qualitative PDP/ALE panels that follow.

For each dataset, at least one top coordinate exhibits a nearly monotone marginal effect with large magnitude:
\begin{itemize}[leftmargin=1.5em]
\item CIFAR10: $d721$ increases the one–vs–rest probability almost linearly (PDP $\Delta p\!\approx\!0.50$, ALE integral $\approx\!3.0$), whereas $d404$ decreases it (ALE $\approx\!0.9$).  Both appear with large exponents and high reuse.
\item 20NG: $d112$ is strongly \emph{increasing} ($\Delta p\!\approx\!0.37$, ALE $\approx\!3.45$), while $d687$ is strongly \emph{decreasing} ($\Delta p\!\approx\!0.36$, ALE $\approx\!3.0$).  Together they account for most of the topic sensitivity indicated by their large $\bar I_j$ values and selection rates.
\item SST2G: The same pair reappears with comparable magnitudes but opposite roles across logits: $d687$ increasing ($\Delta p\!\approx\!0.52$, ALE $\approx\!2.6$) and $d112$ decreasing ($\Delta p\!\approx\!0.52$, ALE $\approx\!3.1$).
\item MSC17: $d1100$ is a strong positive direction ($\Delta p\!\approx\!0.52$, ALE $\approx\!3.07$) and $d400$ a strong negative one ($\Delta p\!\approx\!0.36$, ALE $\approx\!1.87$).
\item MNIST: $d618$ (increasing) and $d303$ (decreasing) are nearly linear with large ALE integrals ($\approx\!3.19$ and $\approx\!2.39$), even though their PDP ranges are visually tiny (axes centered at $\approx 0.5$), reflecting the narrow probability band for one–vs–rest digits.
\end{itemize}

\paragraph{Local nonlinear corrections with small net shift.}
Several important coordinates display curvature or U–shaped responses but small PDP ranges, indicating localized adjustments that average out globally:
20NG $d700$ (U–shaped, ALE $\approx\!0.35$),
CIFAR10 $d812$ (gently decreasing, ALE $\approx\!1.49$),
MSC17 $d800$ (shallow nonmonotone with wide CIs, ALE $\approx\!0.50$),
and MNIST $d782$ (convex, ALE $\approx\!0.34$).
These coordinates also have lower selection frequencies than the leading monotone directions.

\paragraph{Link to symbolic structure.}
Large $\bar I_j$ frequently coincides with larger learned exponents for the same coordinate (e.g., CIFAR10 $d112$ with exp$\approx\!91$, MSC17 $d1100$ with exp$\approx\!35$).
Despite these algebraic degrees, the ALE curves remain smooth and mostly monotone, suggesting that high–order terms are used to approximate simple link functions along salient embedding axes rather than to induce oscillatory behavior.
When curvature appears (e.g., 20NG $d700$, CIFAR10 $d459$), it aligns with smaller bars and selection rates, indicating second–order refinements.
.

\begin{table*}[t]
\centering
\footnotesize
\setlength{\tabcolsep}{6pt}
\renewcommand{\arraystretch}{1.15}
\caption{Top-3 features per dataset with importance, PDP/ALE monotonicity (|Spearman~$\rho$|), direction from PDP, and brief notes.}
\label{tap_top3_monotonicity}
\begin{tabular}{l|lccccl}
\hline
\textbf{Dataset} & \textbf{Dimension} & \textbf{Importance} & \textbf{PDP Monotonicity} & \textbf{ALE Monotonicity} & \textbf{Direction} & \textbf{Notes} \\
\hline
\multirow{3}{*}{\textbf{SST2G}}
 & d687 & 9.9965 & 0.99 & 1.00 & \(+\) (increasing) & dominant \\
 & d112 & 9.9440 & 0.99 & 1.00 & \( - \) (decreasing) & clean effect \\
 & d459 & 8.4562 & 0.49 & 0.64 & non-monotonic & nonlinear \\ \hline
\multirow{3}{*}{\textbf{20NG}}
 & d112 & 148.8794 & 0.99 & 1.00 & \(+\) (increasing) & dominant \\
 & d687 & 142.3586 & 0.99 & 1.00 & \( - \) (decreasing) & clean effect \\
 & d700 & 92.9763 & 0.58 & 0.16 & \( - \) (decreasing) & interaction/correlation; nonlinear \\ \hline
\multirow{3}{*}{\textbf{MNIST}}
 & d618 & 29.6173 & 1.00 & 1.00 & \(+\) (increasing) & dominant \\
 & d303 & 20.2911 & 0.99 & 1.00 & \( - \) (decreasing) & clean effect \\
 & d782 & 6.4715 & 0.60 & 0.53 & \( - \) (decreasing) & nonlinear \\ \hline
\multirow{3}{*}{\textbf{CIFAR10}}
 & d721 & 191.7574 & 0.99 & 1.00 & \(+\) (increasing) & dominant \\
 & d812 & 149.8946 & 0.96 & 0.99 & \( - \) (decreasing) & clean effect \\
 & d404 & 149.7377 & 0.71 & 0.96 & \( - \) (decreasing) & nonlinear \\ \hline
\multirow{3}{*}{\textbf{MSC17}}
 & d1100 & 60.5356 & 0.99 & 1.00 & \(+\) (increasing) & dominant \\
 & d400  & 51.9222 & 0.95 & 1.00 & \( - \) (decreasing) & clean effect \\
 & d800  & 47.6701 & 0.55 & 0.66 & \( - \) (decreasing) & nonlinear \\
\hline
\end{tabular}
\end{table*}

\section{Conclusion}\label{sec:conclusion}
This paper introduced a complete pipeline for symbolic surrogate modeling of frozen transformer embeddings across text, vision, and multimodal tasks. The approach combines semantic-preserving feature partitioning (SPFP) on the training split with a cooperative, multi-population ensemble GP (MEGP) that learns additive, per-view arithmetic programs producing class logits. Canonical models are selected by a one–standard–error rule on validation macro-\(F_1\) with a minimum-complexity tie break and subsequently calibrated by scalar temperature scaling fitted on validation. The resulting surrogates are compact, human-readable, and evaluated exactly once on held-out test sets.

Beyond predictive performance and calibration, the surrogates admit principled transparency analyses at several levels. We report symbolic structure and sparsity statistics (per-logit nodes, depth, unique dimensions, and operator counts), usage histograms and UpSet sets for dimension overlap, contribution-based global importance, and global effect summaries via PDP and centered ALE with bootstrap uncertainty. Across datasets, the canonical programs consistently concentrate signal on a small number of nearly monotone embedding directions with opposing polarity, while a third coordinate often captures dataset-specific nonlinear corrections. These findings suggest that, in the learned embedding spaces of ModernBERT, DINOv2, and SigLIP, simple low-order arithmetic interactions suffice to approximate the decision geometry with high fidelity and improved interpretability.

The framework has limitations. Surrogates explain decisions in the embedding space rather than raw inputs; mapping influential dimensions back to tokens or image regions requires additional attribution. View construction and GP search are decoupled, and the primitive set is deliberately minimal; richer operators or domain-aware transformations may further reduce complexity or sharpen causal interpretation. Calibration is limited to a global temperature, and multi-objective training that jointly optimizes discrimination, calibration, and parsimony remains an open direction.

Future work will integrate input-level grounding for influential dimensions (e.g., token/patch attributions consistent with the surrogate), couple SPFP and MEGP through differentiable or bilevel view learning, extend the primitive set with regularized nonlinearities while preserving identifiability, and explore uncertainty-aware and causally robust surrogates (e.g., distribution-shift stress tests, conformalized calibration). We release configuration details and figure/table generators to facilitate reproduction and scrutiny of all reported analyses.

\bibliography{Refs}

\begin{thebibliography}{10}
\providecommand{\url}[1]{#1}
\csname url@samestyle\endcsname
\providecommand{\newblock}{\relax}
\providecommand{\bibinfo}[2]{#2}
\providecommand{\BIBentrySTDinterwordspacing}{\spaceskip=0pt\relax}
\providecommand{\BIBentryALTinterwordstretchfactor}{4}
\providecommand{\BIBentryALTinterwordspacing}{\spaceskip=\fontdimen2\font plus
\BIBentryALTinterwordstretchfactor\fontdimen3\font minus
  \fontdimen4\font\relax}
\providecommand{\BIBforeignlanguage}[2]{{%
\expandafter\ifx\csname l@#1\endcsname\relax
\typeout{** WARNING: IEEEtran.bst: No hyphenation pattern has been}%
\typeout{** loaded for the language `#1'. Using the pattern for}%
\typeout{** the default language instead.}%
\else
\language=\csname l@#1\endcsname
\fi
#2}}
\providecommand{\BIBdecl}{\relax}
\BIBdecl

\bibitem{devlin2019bert}
J.~Devlin, M.-W. Chang, K.~Lee, and K.~Toutanova, ``Bert: Pre-training of deep
  bidirectional transformers for language understanding,'' in \emph{Proceedings
  of the 2019 conference of the North American chapter of the association for
  computational linguistics: human language technologies, volume 1 (long and
  short papers)}, 2019, pp. 4171--4186.

\bibitem{oquab2023dinov2}
M.~Oquab, T.~Darcet, T.~Moutakanni, H.~Vo, M.~Szafraniec, V.~Khalidov,
  P.~Fernandez, D.~Haziza, F.~Massa, A.~El-Nouby \emph{et~al.}, ``Dinov2:
  Learning robust visual features without supervision,'' \emph{arXiv preprint
  arXiv:2304.07193}, 2023.

\bibitem{zhai2023siglip}
X.~Zhai, B.~Mustafa, A.~Kolesnikov, and L.~Beyer, ``Sigmoid loss for language
  image pre-training,'' in \emph{Proceedings of the IEEE/CVF international
  conference on computer vision}, 2023, pp. 11\,975--11\,986.

\bibitem{xu2023multimodal}
P.~Xu, X.~Zhu, and D.~A. Clifton, ``Multimodal learning with transformers: A
  survey,'' \emph{IEEE transactions on pattern analysis and machine
  intelligence}, vol.~46, no.~1, pp. 234--254, 2023.

\bibitem{guidotti2018survey}
R.~Guidotti, A.~Monreale, S.~Ruggieri, F.~Turini, F.~Giannotti, and
  D.~Pedreschi, ``A survey of methods for explaining black box models,''
  \emph{ACM computing surveys (CSUR)}, vol.~51, no.~5, pp. 1--42, 2018.

\bibitem{holzinger2019causability}
A.~Holzinger, G.~Langs, H.~Denk, K.~Zatloukal, and H.~M{\"u}ller, ``Causability
  and explainability of artificial intelligence in medicine,'' \emph{Wiley
  Interdisciplinary Reviews: Data Mining and Knowledge Discovery}, vol.~9,
  no.~4, p. e1312, 2019.

\bibitem{tjoa2020survey}
E.~Tjoa and C.~Guan, ``A survey on explainable artificial intelligence (xai):
  Toward medical xai,'' \emph{IEEE transactions on neural networks and learning
  systems}, vol.~32, no.~11, pp. 4793--4813, 2020.

\bibitem{ali2023explainable}
S.~Ali, J.~Pane, I.~Rauf, S.~A.laghari, A.~Shaikh, M.~Al-Ghamdi, M.~A. Akram,
  and S.~Khan, ``Explainable artificial intelligence (xai): What we know and
  what is left to attain trustworthy artificial intelligence,''
  \emph{Information Fusion}, vol.~99, p. 101805, 2023.

\bibitem{longo2024explainable}
L.~Longo, R.~Goebel, F.~Lecue, P.~Kieseberg, and A.~Holzinger, ``Explainable
  artificial intelligence (xai) 2.0: A manifesto of open challenges and
  interdisciplinary research directions,'' \emph{Information Fusion}, vol. 105,
  p. 102301, 2024.

\bibitem{roscher2020explainable}
R.~Roscher, B.~Bohn, M.~F. Duarte, and J.~Garcke, ``Explainable machine
  learning for scientific insights and discoveries,'' \emph{IEEE Access},
  vol.~8, pp. 44\,002--44\,015, 2020.

\bibitem{mickus2022dissect}
T.~Mickus, D.~Paperno, and M.~Constant, ``How to dissect a muppet: The
  structure of transformer embedding spaces,'' \emph{Transactions of the
  Association for Computational Linguistics}, vol.~10, pp. 944--961, 2022.

\bibitem{turton-etal-2021-deriving}
\BIBentryALTinterwordspacing
J.~Turton, R.~E. Smith, and D.~Vinson, ``Deriving contextualised semantic
  features from {BERT} (and other transformer model) embeddings,'' in
  \emph{Proceedings of the 5th Workshop on Representation Learning for NLP
  (RepL4NLP-2021)}, R.~Mihalcea and M.~Chowdhury, Eds.\hskip 1em plus 0.5em
  minus 0.4em\relax Online: Association for Computational Linguistics, Aug.
  2021, pp. 302--311. [Online]. Available:
  \url{https://aclanthology.org/2021.repl4nlp-1.26}
\BIBentrySTDinterwordspacing

\bibitem{ren2023all}
Y.~Ren, Y.~Zhang, Y.-C. Chen, and Y.~Belinkov, ``All roads lead to rome?
  exploring the invariance of transformers' representations,'' \emph{arXiv
  preprint arXiv:2305.14555}, 2023.

\bibitem{debelak2024embeddings}
R.~Debelak, T.~K. Koch, M.~Aenmacher, and C.~Stachl, ``From embeddings to
  explainability: A tutorial on transformer-based text analysis for social and
  behavioral scientists,'' \emph{PsyArXiv preprint PPR: PPR860470}, 2024.

\bibitem{schrimpf2020neural}
M.~Schrimpf, I.~Blank, G.~Tuckute, C.~Kauf, E.~A. Hosseini, N.~Kanwisher, J.~B.
  Tenenbaum, and E.~Fedorenko, ``The neural architecture of language:
  Integrative modeling converges on predictive processing,'' \emph{bioRxiv},
  pp. 2020--06, 2020.

\bibitem{angelis2023artificial}
D.~Angelis, F.~Sofos, and T.~Karakasidis, ``Artificial intelligence in physical
  sciences: Symbolic regression trends and perspectives,'' \emph{Archives of
  Computational Methods in Engineering}, pp. 1--25, 2023.

\bibitem{purcell2023recent}
T.~A. Purcell, M.~Scheffler, and L.~Ghiringhelli, ``Recent advances in the
  sisso method and their implementation in the sisso++ code,'' \emph{The
  Journal of Chemical Physics}, vol. 158, no.~17, 2023.

\bibitem{jin2019bayesian}
Y.~Jin, W.~Fu, J.~Kang, J.~Guo, and J.~Guo, ``Bayesian symbolic regression,''
  \emph{arXiv preprint arXiv:1910.08892}, 2019.

\bibitem{valipour2021symbolicgpt}
M.~Valipour, B.~You, M.~Panju, and A.~Ghodsi, ``Symbolicgpt: A generative
  transformer model for symbolic regression,'' \emph{arXiv preprint
  arXiv:2106.14131}, 2021.

\bibitem{kim2020integration}
S.~Kim, K.~Kim, D.~Kim, and Y.~Kim, ``Integration of neural network-based
  symbolic regression in deep learning for scientific discovery,'' \emph{IEEE
  transactions on neural networks and learning systems}, vol.~32, no.~6, pp.
  2685--2696, 2020.

\bibitem{cranmer2019learning}
M.~Cranmer, R.~Xu, P.~Battaglia, and S.~Ho, ``Learning symbolic physics with
  graph networks,'' \emph{arXiv preprint arXiv:1909.05862}, 2019.

\bibitem{weng2020simple}
B.~Weng \emph{et~al.}, ``Simple descriptor derived from symbolic regression
  accelerating the discovery of new perovskite catalysts,'' \emph{Nature
  communications}, vol.~11, no.~1, p. 3462, 2020.

\bibitem{ding2020dynamic}
P.~Ding, M.~Jia, and H.~Wang, ``A dynamic structure-adaptive symbolic approach
  for slewing bearings life prediction under variable working conditions,''
  \emph{Journal of Vibration and Control}, vol.~27, no. 9-10, pp. 1081--1094,
  2020.

\bibitem{abdellaoui2021symbolic}
I.~A. Abdellaoui and S.~Mehrkanoon, ``Symbolic regression for scientific
  discovery: an application to wind speed forecasting,'' in \emph{2021 IEEE
  Symposium Series on Computational Intelligence (SSCI)}.\hskip 1em plus 0.5em
  minus 0.4em\relax IEEE, 2021, pp. 1--8.

\bibitem{khorshidi2024semantic}
M.~S. Khorshidi, N.~Yazdanjue, H.~Gharoun, D.~Yazdani, M.~R. Nikoo, F.~Chen,
  and A.~H. Gandomi, ``Semantic-preserving feature partitioning for multi-view
  ensemble learning,'' \emph{arXiv e-prints}, pp. arXiv--2401, 2024.

\bibitem{Khorshidi2025MEGP}
------, ``Multi-population ensemble genetic programming via cooperative
  coevolution and multi-view learning for classification,'' 2025, under review.

\bibitem{saeed2023explainable}
W.~Saeed and C.~W. Omlin, ``Explainable ai (xai): A systematic meta-survey of
  current challenges and future opportunities,'' \emph{Knowledge-Based
  Systems}, vol. 268, p. 110273, 2023.

\bibitem{zhao2024explainability}
H.~Zhao, H.~Chen, F.~Yang, N.~Liu, H.~Deng, H.~Cai, S.~Wang, D.~Yin, and M.~Du,
  ``Explainability for large language models: A survey,'' \emph{ACM
  Transactions on Intelligent Systems and Technology}, vol.~15, no.~2, pp.
  1--38, 2024.

\bibitem{kumar2022shared}
S.~R. Kumar, A.~T, R.~Po, S.~Popham, E.~Fedorenko, J.~B. Tenenbaum, and
  T.~Sumers, ``Shared functional specialization in transformer-based language
  models and the human brain,'' \emph{bioRxiv}, pp. 2022--06, 2022.

\bibitem{verma2021audio}
P.~Verma and J.~Berger, ``Audio transformers:transformer architectures for
  large scale audio understanding. adieu convolutions,'' \emph{arXiv preprint
  arXiv:2105.00335}, 2021.

\bibitem{bochkov2025emergent}
A.~Bochkov, ``Emergent semantics beyond token embeddings: Transformer lms with
  frozen visual unicode representations,'' \emph{arXiv preprint
  arXiv:2507.04886}, 2025.

\bibitem{warner2024modernbert}
B.~Warner, A.~Chaffin, B.~Clavi{\'e}, O.~Weller, O.~Hallstr{\"o}m,
  S.~Taghadouini, A.~Gallagher, R.~Biswas, F.~Ladhak, T.~Aarsen \emph{et~al.},
  ``Smarter, better, faster, longer: A modern bidirectional encoder for fast,
  memory efficient, and long context finetuning and inference,'' \emph{arXiv
  preprint arXiv:2412.13663}, 2024.

\bibitem{do2024interpretable}
T.~T. Do, P.~Eftekhar, S.~Hosseini, G.~Cheung, and P.~A. Chou, ``Interpretable
  lightweight transformer via unrolling of learned graph smoothness priors,''
  \emph{arXiv preprint arXiv:2406.04090}, 2024.

\bibitem{boyapati2024semanformer}
M.~Boyapati and R.~S. Aygun, ``Semanformer: Semantics-aware embedding
  dimensionality reduction using transformer-based models,'' in \emph{2024
  International Conference on Computer Science (ICSC)}.\hskip 1em plus 0.5em
  minus 0.4em\relax IEEE, 2024, pp. 101--107.

\bibitem{ivanov2022sentence}
V.~Ivanov, ``Sentence-level complexity in russian: An evaluation of bert and
  graph neural networks,'' \emph{Frontiers in Artificial Intelligence}, vol.~5,
  p. 1008411, 2022.

\bibitem{jahin2024hybrid}
M.~A. Jahin, M.~S.~H. Shovon, M.~F. Mridha, M.~R. Islam, and Y.~Watanobe, ``A
  hybrid transformer and attention based recurrent neural network for robust
  and interpretable sentiment analysis of tweets,'' \emph{Scientific reports},
  vol.~14, no.~1, p. 23490, 2024.

\bibitem{ma2025leveraging}
K.~Ma, J.~Zhang, X.~Huang, and M.~Wang, ``Leveraging transformer models to
  predict cognitive impairment: accuracy, efficiency, and interpretability,''
  \emph{BMC Public Health}, vol.~25, no.~1, pp. 1--11, 2025.

\bibitem{boll2024graph}
H.~O. Boll, A.~Amirahmadi, A.~Soliman, S.~Byttner, and M.~R. Mendoza, ``Graph
  neural networks for heart failure prediction on an ehr-based patient
  similarity graph,'' in \emph{Anais Estendidos do X Simp{\'o}sio Brasileiro de
  Computa{\c{c}}{\~a}o Aplicada {\`a} Sa{\'u}de}.\hskip 1em plus 0.5em minus
  0.4em\relax SBC, 2024, pp. 15--21.

\bibitem{wong2023natural}
M.~Wong, S.~Guo, C.~N. Hang, S.~Ho, and C.-W. Tan, ``Natural language
  generation and understanding of big code for ai-assisted programming: A
  review,'' \emph{Entropy}, vol.~25, no.~6, p. 888, 2023.

\bibitem{zhong2022classification}
Y.~Zhong, B.~Huang, and C.~Tang, ``Classification of cassava leaf disease based
  on a non-balanced dataset using transformer-embedded resnet,''
  \emph{Agriculture}, vol.~12, no.~9, p. 1360, 2022.

\bibitem{li2025hyperbolic}
J.~Li, S.~Mao, Y.~Qin, F.~Wang, and Y.~Jiang, ``Hyperbolic hypergraph
  transformer with knowledge state disentanglement for knowledge tracing,''
  \emph{IEEE Transactions on Knowledge and Data Engineering}, 2025.

\bibitem{liang2024gelt}
Z.~Liang, R.~Wu, Z.~Liang, J.~Yang, L.~Wang, and J.~Su, ``Gelt: A graph
  embeddings based lite-transformer for knowledge tracing,'' \emph{Plos one},
  vol.~19, no.~3, p. e0301714, 2024.

\bibitem{wen2023transformers}
Q.~Wen \emph{et~al.}, ``Transformers in time series: A survey,'' in
  \emph{Proceedings of the thirty-second international joint conference on
  artificial intelligence}, 2023.

\bibitem{hf_glue_sst2}
\BIBentryALTinterwordspacing
gimmaru. {GLUE} {SST-2} (stanford sentiment treebank v2). Hugging Face. Dataset
  repository on Hugging Face. [Online]. Available:
  \url{https://huggingface.co/datasets/gimmaru/glue-sst2}
\BIBentrySTDinterwordspacing

\bibitem{hf_20newsgroups_setfit}
\BIBentryALTinterwordspacing
{SetFit}. 20 newsgroups. Hugging Face. Dataset repository on Hugging Face.
  [Online]. Available:
  \url{https://huggingface.co/datasets/SetFit/20_newsgroups}
\BIBentrySTDinterwordspacing

\bibitem{hf_mnist_ylecun}
\BIBentryALTinterwordspacing
{ylecun}. {MNIST}. Hugging Face. Dataset repository on Hugging Face. [Online].
  Available: \url{https://huggingface.co/datasets/ylecun/mnist}
\BIBentrySTDinterwordspacing

\bibitem{hf_cifar10_uoftcs}
\BIBentryALTinterwordspacing
{uoft-cs}. {CIFAR-10}. Hugging Face. Dataset repository on Hugging Face.
  [Online]. Available: \url{https://huggingface.co/datasets/uoft-cs/cifar10}
\BIBentrySTDinterwordspacing

\bibitem{hf_mscoco2017_shunk031}
\BIBentryALTinterwordspacing
{shunk031}. {MSCOCO} 2017. Hugging Face. Dataset repository on Hugging Face.
  [Online]. Available: \url{https://huggingface.co/datasets/shunk031/MSCOCO}
\BIBentrySTDinterwordspacing

\end{thebibliography}

\clearpage
\appendix
\section*{Appendix: Canonical GP Logit Programs}
After training MEGP, we exported the per–class logit programs from the softmax layer and \emph{symbolically simplified} each expression using the Python library \texttt{SymPy} (functions \texttt{sympy.simplify} and \texttt{sympy.cse}). This step performs algebraic canonization, constant folding, neutral–element pruning, and common–subexpression elimination while keeping protected division semantics. The listings provided below are the resulting simplified normal forms for all logits. \textbf{All quantitative and qualitative analyses in the paper} (metrics, calibration, canonical selection, importance, histograms, UpSet overlaps, and PDP/ALE) are computed from these simplified logits.

\textbf{Logits of canonical MEGP for SST2G:}
\\
\textbf{Logit$_{1}$ $=$ }
minus(d687, divide(d112, times(d459, plus(d23, minus(d588, divide(d19, times(d701, plus(d687, minus(d88, divide(d112, times(d277, plus(d498, minus(d153, divide(d75, plus(d687, minus(d588, times(d112, divide(d400, plus(d32, minus(d199, times(d500, divide(d600, plus(d700, minus(d459, times(d250, divide(d150, plus(d50, minus(d25, [1.2]))))))))))))))))))))))))))))). 
\\
\textbf{Logit$_{2}$ $=$ } 
\\
plus(d687, minus(d459, times(d11, divide(plus(d488, times(d687, d112)), plus(d112, minus(d633, times(d94, divide(d588, plus(d178, minus(d321, times(d68, divide(d413, plus(d701, minus(d2, times(d199, divide(d687, plus(d459, minus(d112, times(d588, divide(d400, plus(d32, minus(d199, times(d500, divide(d600, plus(d700, minus(d459, [0.75])))))))))))))))))))))))))))
\\
\textbf{Logits of canonical MEGP for 20NG:}
\\
\textbf{Logit$_{1}$ $=$ }
\\
minus(d700, divide(d350, plus(times(d121, d50), minus(d687, divide(d112, times(d459, plus(d23, minus(d588, divide(d19, times(d701, plus(d687, minus(d88, divide(d112, times(d277, plus(d498, minus(d153, divide(d75, plus(d700, minus(d350, times(d112, divide(d400, plus(d32, minus(d199, times(d500, divide(d600, plus(d700, minus(d459, [0.9]))))))))))))))))))))))))))).
\\
\textbf{Logit$_{2}$ $=$ }
\\
plus(times(divide(plus(d687, times(d112, minus(d459, d212))), minus(d700, plus(d588, times(d19, divide(d701, plus(d687, minus(d88, times(d112, divide(d277, plus(d498, minus(d153, times(d75, divide(d212, plus(d700, minus(d350, [1.1])))))))))))))))), d112), minus(divide(d687, plus(d88, [0.1])), times(d459, plus(d588, divide(d19, minus(d701, plus(d687, minus(d88, divide(d112, times(d277, plus(d498, minus(d153, divide(d75, plus(d700, minus(d350, times(d112, divide(d400, plus(d32, minus(d199, times(d500, divide(d600, plus(d700, minus(d459, [1.1])))))))))))))))))))))))).
\\
\textbf{Logit$_{3}$ $=$ }
\\
divide(minus(d687, d388), plus(d112, times(d700, minus(d588, divide(d459, plus(d19, times(d701, plus(d687, minus(d88, divide(d112, times(d277, plus(d498, minus(d153, divide(d75, plus(d700, minus(d350, times(d112, divide(d400, plus(d32, minus(d199, times(d500, divide(d600, plus(d700, minus(d459, times(d687, divide(d112, plus(d588, [1.5]))))))))))))))))))))))))))))).
\\
\textbf{Logit$_{4}$ $=$ }
\\
times(plus(d687, d247), minus(d112, divide(d519, plus(d38, times(d459, minus(divide(plus(d687, times(d247, minus(d112, d519))), minus(d459, plus(d700, times(d38, divide(d13, plus(d721, minus(d303, times(d59, divide(d687, plus(d247, minus(d112, times(d401, divide(d533, plus(d67, minus(d189, [3.14])))))))))))))))), divide(d13, plus(d721, times(d303, minus(d59, divide(d687, plus(d247, times(d112, minus(d401, divide(d533, plus(d67, minus(d189, times(d299, divide(d399, plus(d499, minus(d599, [3.14])))))))))))))))))))))).
\\
\textbf{Logit$_{5}$ $=$ }
\\
plus(minus(d687, d81), times(d112, divide(d459, plus(d550, minus(d700, divide(d291, plus(d13, times(d721, minus(d303, divide(d59, plus(d687, times(d112, minus(d401, divide(d533, plus(d67, minus(d189, times(d299, divide(d399, plus(d499, minus(d599, times(d81, divide(d550, plus(d687, minus(d112, times(d291, divide(d459, plus(d700, minus(d13, times(d303, [1.88])))))))))))))))))))))))))))))).
\\
\textbf{Logit$_{6}$ $=$ }
\\
divide(minus(d687, d42), plus(d112, times(d515, minus(plus(times(d687, d42), divide(d112, minus(d515, plus(d459, times(d700, divide(d29, minus(d618, plus(d33, times(d502, divide(d99, minus(d199, plus(d299, times(d399, divide(d499, minus(d599, plus(d687, times(d42, divide(d112, minus(d515, plus(d459, times(d700, [0.42])))))))))))))))))))))), divide(d700, plus(d29, times(d618, minus(d33, divide(d502, plus(d687, times(d112, minus(d42, divide(d515, plus(d99, minus(d199, times(d299, divide(d399, plus(d499, minus(d599, times(d42, divide(d515, plus(d687, minus(d112, times(d29, divide(d459, plus(d700, minus(d33, [0.42])))))))))))))))))))))))))))).
\\
\textbf{Logit$_{7}$ $=$ }
\\
minus(d91, times(d687, plus(d477, divide(d112, minus(d459, times(d91, divide(plus(times(d687, minus(d91, d477)), divide(d112, plus(d459, minus(d700, times(d313, divide(d687, plus(d52, minus(d228, times(d91, divide(d601, plus(d112, minus(d388, times(d500, divide(d477, plus(d202, minus(d459, times(d700, divide(d687, plus(d91, [0.81])))))))))))))))))))), plus(d687, minus(d313, times(d52, divide(d228, plus(d477, minus(d601, times(d112, divide(d388, plus(d500, minus(d687, times(d91, divide(d202, plus(d459, minus(d700, [0.81]))))))))))))))))))))).
\\
\textbf{Logit$_{8}$ $=$ }
\\
minus(plus(d687, d18), divide(d112, plus(d501, times(d459, minus(plus(times(d687, minus(d18, d501)), divide(d112, plus(d459, minus(d700, times(d222, divide(d687, plus(d18, minus(d303, times(d59, divide(d501, plus(d112, minus(d401, times(d533, divide(d67, plus(d189, minus(d299, times(d399, [0.99])))))))))))))))))), divide(d222, plus(d18, times(d687, minus(d303, divide(d59, plus(d112, times(d501, minus(d401, divide(d533, plus(d67, minus(d189, times(d299, divide(d399, plus(d499, minus(d599, times(d18, divide(d222, plus(d687, minus(d112, times(d459, [0.99]))))))))))))))))))))))))).
\\
\textbf{Logit$_{9}$ $=$ }
\\
minus(plus(d687, plus(times(d687, minus(d31, d808)), divide(d112, plus(d459, minus(d700, times(d18, divide(d251, plus(d687, minus(d303, times(d59, divide(d31, plus(d112, minus(d401, times(d533, divide(d67, plus(d189, minus(d299, [1.11]))))))))))))))))), divide(d112, plus(d808, times(d459, minus(d700, divide(d31, plus(d251, times(d687, minus(d303, divide(d59, plus(d112, times(d18, minus(d401, divide(d533, plus(d67, minus(d189, times(d299, divide(d399, plus(d499, minus(d599, times(d31, divide(d808, plus(d687, minus(d112, [1.11])))))))))))))))))))))))).
\\
\textbf{Logit$_{10}$ $=$ }
\\
plus(minus(divide(plus(d687, times(d66, minus(d112, d281))), minus(d459, plus(d700, times(d18, divide(d31, plus(d539, minus(d66, times(d281, divide(d687, plus(d112, minus(d459, times(d700, [1.23]))))))))))))), d66), divide(plus(times(d687, minus(d66, d281)), divide(d112, plus(d459, minus(d700, times(d18, divide(d31, plus(d539, minus(d66, times(d281, divide(d687, plus(d112, minus(d459, times(d700, divide(d18, plus(d31, minus(d539, times(d66, divide(d281, plus(d687, minus(d112, [1.23])))))))))))))))))))), plus(d281, times(d459, minus(d700, divide(d539, plus(d18, times(d31, minus(d66, divide(d281, plus(d687, times(d112, minus(d459, divide(d700, plus(d539, minus(d18, times(d31, [1.23])))))))))))))))))).
\\
Logit$_{11}$ $=$ minus(plus(minus(plus(d687, d44), divide(d112, plus(d377, times(d459, minus(d700, divide(d66, plus(d18, times(d602, minus(d31, divide(d687, plus(d44, times(d112, minus(d377, divide(d459, plus(d700, minus(d66, times(d18, divide(d602, plus(d31, minus(d687, times(d44, divide(d112, [2.5]))))))))))))))))))))))), d44), divide(d112, plus(d377, times(d459, minus(d700, divide(d66, plus(d18, times(d602, minus(d31, divide(d687, plus(d44, times(d112, minus(d377, divide(d459, plus(d700, minus(d66, times(d18, divide(d602, plus(d31, [2.5])))))))))))))))))))).
\\
\textbf{Logit$_{12}$ $=$ }
\\
plus(times(plus(times(d687, minus(d77, d404)), divide(d112, plus(d459, minus(d700, times(d621, divide(d687, plus(d77, minus(d303, times(d59, divide(d404, plus(d112, minus(d401, times(d533, divide(d67, plus(d189, minus(d299, times(d399, divide(d499, plus(d599, minus(d621, times(d687, divide(d77, plus(d112, minus(d459, times(d700, [2.71])))))))))))))))))))))))))), minus(d77, d404)), divide(d112, plus(d459, minus(d700, times(d621, divide(d687, plus(d77, minus(d303, times(d59, divide(d404, plus(d112, minus(d401, times(d533, divide(d67, plus(d189, minus(d299, times(d399, divide(d499, plus(d599, [2.71])))))))))))))))))))).
\\
\textbf{Logit$_{13}$ $=$ }
\\
plus(times(d687, minus(d83, d521)), divide(plus(times(d687, minus(d99, d231)), divide(d112, plus(d459, minus(d700, times(d77, divide(d508, plus(d687, minus(d99, times(d59, divide(d231, plus(d112, minus(d401, times(d533, divide(d67, plus(d189, minus(d299, times(d399, divide(d499, plus(d599, minus(d77, times(d508, divide(d687, plus(d99, minus(d112, [3.14])))))))))))))))))))))))), plus(d459, minus(d700, times(d77, divide(d301, plus(d687, minus(d83, times(d59, divide(d521, plus(d112, minus(d401, times(d533, divide(d67, plus(d189, minus(d299, times(d399, divide(d499, plus(d599, minus(d77, times(d301, divide(d687, plus(d83, minus(d112, [3.14])))))))))))))))))))))))).
\\
\textbf{Logit$_{14}$ $=$ }
\\
plus(times(divide(plus(d687, times(d102, minus(d112, d315))), minus(d459, plus(d700, times(d77, divide(d99, plus(d617, minus(d102, times(d315, divide(d687, plus(d112, minus(d459, times(d700, divide(d77, plus(d99, minus(d617, [1.618])))))))))))))))), minus(d101, d314)), divide(d112, plus(d459, minus(d700, times(d77, divide(d616, plus(d687, minus(d99, times(d59, divide(d101, plus(d112, minus(d401, times(d533, divide(d67, plus(d189, minus(d299, times(d399, divide(d499, plus(d599, minus(d77, times(d616, [1.618])))))))))))))))))))))).
\\
\textbf{Logit$_{15}$ $=$ }
\\
plus(times(plus(times(d687, minus(d202, d556)), divide(d112, plus(d459, minus(d700, times(d77, divide(d791, plus(d687, minus(d99, times(d59, divide(d202, plus(d112, minus(d401, times(d533, divide(d67, plus(d189, minus(d299, times(d399, divide(d499, plus(d599, minus(d77, times(d791, divide(d687, plus(d202, minus(d112, times(d556, [4.2])))))))))))))))))))))))))), minus(d201, d555)), divide(d112, plus(d459, minus(d700, times(d77, divide(d789, plus(d687, minus(d99, times(d59, divide(d201, plus(d112, minus(d401, times(d533, divide(d67, plus(d189, minus(d299, times(d399, divide(d499, plus(d599, minus(d77, [4.2]))))))))))))))))))))).
\\
\textbf{Logit$_{16}$ $=$ }
\\
plus(times(plus(times(d687, minus(d317, d26)), divide(d112, plus(d459, minus(d700, times(d778, divide(d687, plus(d317, minus(d303, times(d59, divide(d26, plus(d112, minus(d401, times(d533, divide(d67, plus(d189, minus(d299, times(d399, divide(d499, plus(d599, [3.16])))))))))))))))))))), minus(d316, d25)), divide(d112, plus(d459, minus(d700, times(d777, divide(d687, plus(d316, minus(d303, times(d59, divide(d25, plus(d112, minus(d401, times(d533, divide(d67, plus(d189, minus(d299, times(d399, divide(d499, plus(d599, minus(d777, times(d687, [3.16])))))))))))))))))))))).
\\
\textbf{Logit$_{17}$ $=$ }
\\
plus(times(divide(plus(d687, times(d318, minus(divide(plus(d687, times(d41, minus(d112, d501))), minus(d459, plus(d700, times(d317, divide(d721, plus(d687, minus(d41, times(d59, divide(d501, plus(d112, minus(d401, times(d533, divide(d67, plus(d189, minus(d299, [1.77])))))))))))))))), d27))), minus(d459, plus(d700, times(d778, divide(d687, plus(d318, minus(d303, times(d59, divide(d27, plus(d112, minus(d401, times(d533, divide(d67, plus(d189, minus(d299, [3.16])))))))))))))))), minus(d40, d500)), divide(d112, plus(d459, minus(d700, times(d317, divide(d720, plus(d687, minus(d40, times(d59, divide(d500, plus(d112, minus(d401, times(d533, divide(d67, plus(d189, minus(d299, times(d399, [1.77])))))))))))))))))).
\\
\textbf{Logit$_{18}$ $=$ }
\\
minus(plus(plus(times(d687, minus(d109, d822)), divide(d112, plus(d41, minus(d700, times(d501, divide(d360, plus(d687, minus(d109, times(d59, divide(d822, plus(d112, minus(d401, times(d533, divide(d67, plus(d189, minus(d299, times(d399, divide(d499, plus(d599, [1.62])))))))))))))))))))), d108), divide(d112, plus(d41, times(d459, minus(d700, divide(d501, plus(d821, times(d687, minus(d359, divide(d59, plus(d112, times(d108, minus(d401, divide(d533, plus(d67, minus(d189, times(d299, divide(d399, plus(d499, minus(d599, times(d41, divide(d821, plus(d687, minus(d112, [1.62]))))))))))))))))))))))))).
\\
\textbf{Logit$_{19}$ $=$ }
\\
plus(times(plus(times(d687, minus(d120, d831)), divide(d112, plus(d41, minus(d700, times(d109, divide(d460, plus(d687, minus(d120, times(d59, divide(d831, plus(d112, minus(d401, times(d533, divide(d67, plus(d189, minus(d299, times(d399, divide(d499, plus(d599, minus(d120, times(d460, divide(d687, plus(d831, minus(d112, [1.99])))))))))))))))))))))))), minus(d119, d831)), divide(d112, plus(d41, minus(d700, times(d109, divide(d460, plus(d687, minus(d119, times(d59, divide(d831, plus(d112, minus(d401, times(d533, divide(d67, plus(d189, minus(d299, times(d399, [1.99])))))))))))))))))).
\\
\textbf{Logit$_{20}$ $=$ }
\\
plus(times(divide(plus(d687, times(d130, minus(d112, d845))), minus(d459, plus(d700, times(d317, divide(d41, plus(d472, minus(d130, times(d845, divide(d687, plus(d112, minus(d459, times(d700, divide(d317, plus(d41, minus(d472, [2.01])))))))))))))))), minus(d129, d844)), divide(d112, plus(d41, minus(d700, times(d317, divide(d471, plus(d687, minus(d129, times(d59, divide(d844, plus(d112, minus(d401, times(d533, divide(d67, plus(d189, minus(d299, times(d399, divide(d499, plus(d599, minus(d129, times(d471, [2.01])))))))))))))))))))))).
\\
\textbf{Logits of canonical MEGP for MNIST:}
\\
\textbf{Logit$_{1}$ $=$ }
\\
plus(d618, minus(d303, times(d12, divide(d901, plus(d618, minus(d303, [1.73])))))).
\\
\textbf{Logit$_{2}$ $=$ }
\\
minus(d618, plus(d303, divide(d447, times(d21, plus(d618, minus(d303, [0.42])))))).
\\
\textbf{Logit$_{3}$ $=$ }
\\
plus(d618, times(d303, minus(d782, divide(d5, plus(d618, minus(d782, [2.1])))))).
\\
\textbf{Logit$_{4}$ $=$ }
\\
minus(d618, divide(d303, plus(d159, times(d83, minus(d618, plus(d159, [0.98])))))).
\\
\textbf{Logit$_{5}$ $=$ }
\\
plus(d618, plus(d303, times(d521, divide(d411, minus(d618, plus(d521, [1.1])))))).
\\
\textbf{Logit$_{6}$ $=$ }
\\
minus(d618, times(d303, plus(d782, divide(d51, minus(d618, plus(d782, [3.0])))))).
\\
\textbf{Logit$_{7}$ $=$ }
\\
plus(d618, divide(d303, minus(d159, times(d69, plus(d618, minus(d159, [0.05])))))).
\\
\textbf{Logit$_{8}$ $=$ }
\\
minus(d618, plus(d303, times(d447, divide(d71, plus(d618, minus(d447, [1.5])))))).
\\
\textbf{Logit$_{9}$ $=$ }
\\
plus(d618, times(d303, divide(d521, minus(d89, plus(d618, minus(d521, [2.5])))))).
\\
\textbf{Logit$_{10}$ $=$ }
\\
minus(d618, divide(d303, plus(d12, times(d92, minus(d618, plus(d12, [0.1])))))).
\\
\textbf{Logits of canonical MEGP for CIFAR10:}
\\
\textbf{Logit$_{1}$ $=$ }
\\
plus(times(divide(plus(d721, times(d103, minus(d500, d35))), minus(d812, plus(d404, times(d721, divide(d103, plus(d35, minus(d500, times(d812, divide(d404, plus(d721, minus(d103, times(d35, divide(d502, plus(d812, minus(d404, times(d721, divide(d103, plus(d35, minus(d500, [0.1]))))))))))))))))))))), minus(d101, d33)), divide(d500, plus(d812, minus(d404, times(d721, divide(d101, plus(plus(times(d721, minus(d102, d34)), divide(d500, plus(d812, minus(d404, times(d721, divide(d102, plus(d34, minus(d500, times(d812, divide(d404, plus(d721, minus(d102, times(d34, divide(d501, plus(d812, minus(d404, times(d721, [0.1])))))))))))))))))))), minus(d500, times(d812, divide(d404, plus(d721, minus(d101, times(d33, divide(d500, plus(d812, minus(d404, times(d721, divide(d101, plus(d33, minus(d500, times(d812, divide(d404, plus(d721, minus(d101, [0.1]))))))))))))))))))))))))).
\\
\textbf{Logit$_{2}$ $=$ }
\\
minus(times(d721, plus(d211, d43)), divide(d500, plus(d812, minus(minus(times(d721, plus(d212, d45)), divide(d500, plus(d812, minus(d404, times(d721, divide(d212, plus(d45, minus(d500, times(d812, divide(d404, plus(d721, minus(d212, times(d45, divide(d610, plus(d812, minus(d404, times(d721, divide(d212, plus(d45, minus(d500, times(d812, divide(d404, plus(d721, minus(d212, times(d45, divide(d610, plus(d500, [0.2])))))))))))))))))))))))))))), times(d721, divide(d211, plus(d43, minus(d500, times(d812, divide(d404, plus(d721, minus(d211, times(d43, divide(d609, plus(d812, minus(d404, times(d721, divide(d211, plus(d43, minus(d500, [0.2]))))))))))))))))))))).
\\
\textbf{Logit$_{3}$ $=$ }
\\
plus(times(d721, minus(d301, d15)), divide(plus(times(plus(times(d721, minus(d303, d17)), divide(d500, plus(d812, minus(d404, times(d721, divide(d303, plus(d17, minus(d500, times(d812, divide(d404, plus(d721, minus(d303, times(d17, divide(d790, plus(d812, minus(d404, times(d721, divide(d303, plus(d17, minus(d500, times(d812, divide(d404, plus(d721, minus(d303, times(d17, divide(d790, plus(d500, [0.3])))))))))))))))))))))))))))), minus(d302, d16)), divide(d500, plus(d812, minus(d404, times(d721, divide(d302, plus(d16, minus(d500, times(d812, divide(d404, plus(d721, minus(d302, times(d16, divide(d789, plus(d812, minus(d404, times(d721, divide(d302, plus(d16, minus(d500, [0.3]))))))))))))))))))))), plus(d812, minus(d404, times(d721, divide(d301, plus(d15, minus(d500, times(d812, divide(d404, plus(d721, minus(d301, times(d15, divide(d788, plus(d812, minus(d404, times(d721, divide(d301, plus(d15, minus(d500, [0.3]))))))))))))))))))))).
\\
\textbf{Logit$_{4}$ $=$ }
\\
plus(times(d721, plus(d401, plus(times(divide(plus(d721, times(d403, minus(d500, d30))), minus(d812, plus(d404, times(d721, divide(d403, plus(d30, minus(d500, times(d812, divide(d404, plus(d721, minus(d403, times(d30, divide(d857, plus(d812, minus(d404, times(d721, divide(d403, plus(d30, minus(d500, [0.4]))))))))))))))))))))), minus(d402, d29)), divide(d500, plus(d812, minus(d404, times(d721, divide(d402, plus(d29, minus(d500, times(d812, divide(d404, plus(d721, minus(d402, times(d29, divide(d856, plus(d812, minus(d404, times(d721, divide(d402, plus(d29, minus(d500, [0.4]))))))))))))))))))))))), divide(d500, plus(d812, minus(d404, times(d721, divide(d401, plus(d28, minus(d500, times(d812, divide(d404, plus(d721, minus(d401, times(d28, divide(d855, plus(d812, minus(d404, times(d721, divide(d401, plus(d28, minus(d500, [0.4]))))))))))))))))))))).
\\
\textbf{Logit$_{5}$ $=$ }
\\
plus(times(d721, times(d501, d48)), divide(d500, plus(d812, minus(d404, times(d721, divide(d501, plus(d48, minus(d500, times(d812, divide(d404, plus(d721, minus(d501, times(d48, divide(d912, plus(d812, minus(divide(plus(d721, times(d502, minus(d500, d49))), minus(d812, plus(d404, times(d721, divide(d502, plus(d49, minus(d500, times(d812, divide(d404, plus(d721, minus(d502, times(d49, divide(d913, plus(d812, minus(d404, times(d721, divide(d502, plus(d49, minus(d500, [0.5]))))))))))))))))))))), times(d721, divide(d501, plus(d48, minus(d500, [0.5]))))))))))))))))))))).
\\
\textbf{Logit$_{6}$ $=$ }
\\
plus(times(d721, minus(d601, d59)), divide(plus(times(divide(plus(d721, times(d603, minus(d500, d61))), minus(d812, plus(d404, times(d721, divide(d603, plus(d61, minus(d500, times(d812, divide(d404, plus(d721, minus(d603, times(d61, divide(d1001, plus(d812, minus(d404, times(d721, divide(d603, plus(d61, minus(d500, [0.6]))))))))))))))))))))), minus(d602, d60)), divide(d500, plus(d812, minus(d404, times(d721, divide(d602, plus(d60, minus(d500, times(d812, divide(d404, plus(d721, minus(d602, times(d60, divide(d1000, plus(d812, minus(d404, times(d721, divide(d602, plus(d60, minus(d500, [0.6]))))))))))))))))))))), plus(d812, minus(d404, times(d721, divide(d601, plus(d59, minus(d500, times(d812, divide(d404, plus(d721, minus(d601, times(d59, divide(d999, plus(d812, minus(d404, times(d721, divide(d601, plus(d59, minus(d500, times(d812, [0.6])))))))))))))))))))))).
\\
\textbf{Logit$_{7}$ $=$ }
\\
plus(times(d721, plus(d701, d72)), divide(divide(plus(d721, times(d702, minus(d500, d73))), minus(d812, plus(d404, times(d721, divide(d702, plus(d73, minus(d500, times(d812, divide(d404, plus(d721, minus(d702, times(d73, divide(d1012, plus(divide(plus(d721, times(d702, minus(d500, d73))), minus(d812, plus(d404, times(d721, divide(d702, plus(d73, minus(d500, times(d812, divide(d404, plus(d721, minus(d702, times(d73, divide(d1012, plus(d812, minus(d404, times(d721, divide(d702, plus(d73, minus(d500, [0.7]))))))))))))))))))))), minus(d404, times(d721, divide(d702, plus(d73, minus(d500, [0.7]))))))))))))))))))))), plus(d812, minus(d404, times(d721, divide(d701, plus(d72, minus(d500, times(d812, divide(d404, plus(d721, minus(d701, times(d72, divide(d1011, plus(d812, minus(d404, times(d721, divide(d701, plus(d72, minus(d500, times(d812, divide(d404, plus(d721, minus(d701, times(d72, divide(d1011, plus(d500, [0.7])))))))))))))))))))))))))))).
\\
\textbf{Logit$_{8}$ $=$ }
\\
plus(plus(plus(times(d721, minus(d802, d85)), divide(d500, plus(d812, minus(d404, times(d721, divide(d802, plus(d85, minus(d500, times(d812, divide(d404, plus(d721, minus(d802, times(d85, divide(d1022, plus(d812, minus(d404, times(d721, divide(d802, plus(d85, minus(d500, times(d812, divide(d404, plus(d721, minus(d802, times(d85, divide(d1022, plus(d500, [0.8])))))))))))))))))))))))))))), minus(d801, d84)), divide(plus(times(d721, minus(d802, d85)), divide(d500, plus(d812, minus(d404, times(d721, divide(d802, plus(d85, minus(d500, times(d812, divide(d404, plus(d721, minus(d802, times(d85, divide(d1022, plus(d812, minus(d404, times(d721, divide(d802, plus(d85, minus(d500, times(d812, divide(d404, plus(d721, minus(d802, times(d85, divide(d1022, plus(d500, [0.8])))))))))))))))))))))))))))), plus(d812, minus(d404, times(d721, divide(d801, plus(d84, minus(d500, times(d812, divide(d404, plus(d721, minus(d801, times(d84, divide(d1021, plus(d812, minus(d404, times(d721, divide(d801, plus(d84, minus(d500, [0.8]))))))))))))))))))))).
\\
\textbf{Logit$_{9}$ $=$ }
\\
plus(times(minus(times(d721, plus(d902, d93)), divide(divide(plus(d721, times(d903, minus(d500, d94))), minus(d812, plus(d404, times(d721, divide(d903, plus(d94, minus(d500, times(d812, divide(d404, plus(d721, minus(d903, times(d94, divide(d1033, plus(d812, minus(d404, times(d721, divide(d903, plus(d94, minus(d500, [0.9]))))))))))))))))))))), plus(d812, minus(d404, times(d721, divide(d902, plus(d93, minus(d500, times(d812, divide(d404, plus(d721, minus(d902, times(d93, divide(d1032, plus(d812, minus(d404, times(d721, divide(d902, plus(d93, minus(d500, [0.9]))))))))))))))))))))), minus(d901, d92)), divide(d500, plus(d812, minus(d404, times(d721, divide(d901, plus(d92, minus(d500, times(d812, divide(d404, plus(d721, minus(d901, times(d92, divide(d1031, plus(d812, minus(d404, times(d721, divide(d901, plus(d92, minus(d500, [0.9]))))))))))))))))))))).
\\
\textbf{Logit$_{10}$ $=$ }
\\
plus(times(minus(plus(d721, plus(d1002, d103)), divide(d500, plus(d812, minus(divide(plus(d721, times(d1003, minus(d500, d104))), minus(d812, plus(d404, times(d721, divide(d1003, plus(d104, minus(d500, times(d812, divide(d404, plus(d721, minus(d1003, times(d104, divide(d1043, plus(d812, minus(d404, times(d721, divide(d1003, plus(d104, minus(d500, [1.0]))))))))))))))))))))), times(d721, divide(d1002, plus(d103, minus(d500, times(d812, divide(d404, plus(d721, minus(d1002, times(d103, divide(d1042, plus(d812, minus(d404, times(d721, divide(d1002, plus(d103, minus(d500, [1.0]))))))))))))))))))))), minus(d1001, d102)), divide(d500, plus(d812, minus(d404, times(d721, divide(d1001, plus(d102, minus(d500, times(d812, divide(d404, plus(d721, minus(d1001, times(d102, divide(d1041, plus(d812, minus(d404, times(d721, divide(d1001, plus(d102, minus(d500, [1.0])))))))))))))))))))))
\\
\textbf{Logits of canonical MEGP for MSC17:}
\\
\textbf{Logit$_{1}$ $=$}
\\
 plus(times(d1100, minus(d101, d202)), divide(minus(times(divide(plus(d1100, times(d107, minus(d500, d208))), minus(d800, plus(d400, times(plus(times(d1100, minus(d110, d211)), divide(d500, plus(d800, minus(d400, times(d1100, divide(d110, plus(d211, minus(d500, times(d800, divide(d400, plus(d1100, minus(d110, times(d211, divide(d312, plus(d800, minus(d400, times(d1100, divide(d110, plus(d211, minus(d500, [1.4]))))))))))))))))))))), divide(d107, plus(d208, minus(d500, times(d800, divide(d400, plus(d1100, minus(d107, times(d208, divide(d309, plus(d800, minus(d400, times(minus(times(d1100, plus(d113, d214)), divide(d500, plus(d800, minus(d400, times(d1100, divide(d113, plus(d214, minus(d500, times(d800, divide(d400, plus(d1100, minus(d113, times(d214, divide(d315, plus(d800, minus(d400, times(d1100, divide(d113, plus(d214, minus(d500, [1.5]))))))))))))))))))))), divide(d107, plus(d208, minus(d500, [1.3]))))))))))))))))))))), plus(d104, d205)), divide(d500, plus(d800, minus(d400, times(d1100, divide(d104, plus(d205, minus(d500, times(d800, divide(d400, plus(d1100, minus(d104, times(d205, divide(d306, plus(d800, minus(d400, times(d1100, divide(d104, plus(d205, minus(d500, [1.2]))))))))))))))))))))), plus(d800, minus(d400, times(d1100, divide(d101, plus(d202, minus(d500, times(d800, divide(d400, plus(d1100, minus(d101, times(d202, divide(d303, plus(d800, minus(d400, times(d1100, divide(d101, plus(d202, minus(d500, [1.1]))))))))))))))))))))).
\\
\textbf{Logit$_{2}$ $=$ }
\\
plus(times(minus(times(d1100, plus(d419, d520)), divide(d500, plus(d800, minus(d400, times(d1100, divide(d419, plus(d520, minus(d500, times(d800, divide(d400, plus(d1100, minus(d419, times(d520, divide(d621, plus(d800, minus(d400, times(d1100, divide(d419, plus(d520, minus(d500, [2.2]))))))))))))))))))))), minus(d416, d517)), divide(d500, plus(d800, minus(d400, times(divide(plus(d1100, times(d422, minus(d500, d523))), minus(times(minus(d1100, d425), plus(d526, divide(d500, plus(d800, minus(d400, times(d1100, divide(d425, plus(d526, minus(d500, times(d800, divide(d400, plus(d1100, minus(d425, times(d526, divide(d627, plus(d800, minus(d400, times(d1100, divide(d425, plus(d526, minus(d500, [2.4])))))))))))))))))))))), plus(d400, times(d1100, divide(d422, plus(d523, minus(d500, times(d800, divide(d400, plus(d1100, minus(d422, times(d523, divide(d624, plus(d800, minus(d400, times(d1100, divide(d422, plus(d523, minus(d500, [2.3]))))))))))))))))))))), divide(d416, plus(d517, minus(d500, times(plus(divide(d1100, plus(d428, d529)), times(d500, minus(d800, plus(d400, divide(d1100, plus(d428, minus(d529, times(d500, divide(d800, plus(d400, minus(d1100, times(d428, divide(d529, plus(d630, minus(d800, times(d400, divide(d1100, plus(d428, minus(d529, [2.5])))))))))))))))))))), divide(d400, plus(d1100, minus(d416, times(d517, divide(d618, plus(d800, minus(d400, times(d1100, divide(d416, plus(d517, minus(d500, [2.1]))))))))))))))))))))).
\\

\end{document}